\RequirePackage{rotating} 
\documentclass[nonacm, format=acmsmall, review=false, screen=true, prologue,table,xcdraw]{acmart}
\usepackage[utf8]{inputenc}
\usepackage[T1]{fontenc}    
\usepackage{hyperref}       
\usepackage{url}            
\usepackage{booktabs}       
\usepackage{amsfonts}       
\usepackage{nicefrac}       
\usepackage{microtype}      
\usepackage{graphicx}
\usepackage{booktabs}
\usepackage{enumitem}
\usepackage{adjustbox}
\usepackage{array}
\usepackage{longtable}
\usepackage{pdflscape}
\usepackage{makecell}
\usepackage{rotating}
\usepackage{afterpage}
\usepackage{multirow}

\newcolumntype{L}[1]{>{\raggedright\let\newline\\\arraybackslash\hspace{0pt}}m{#1}}
\newcolumntype{C}[1]{>{\centering\let\newline\\\arraybackslash\hspace{0pt}}m{#1}}
\newcolumntype{R}[1]{>{\raggedleft\let\newline\\\arraybackslash\hspace{0pt}}m{#1}}

\makeatletter
\def\subsubsection{\@startsection{subsubsection}{3}{10pt}%
	{-.5\baselineskip \@plus -2\p@ \@minus -.2\p@}%
	{3.5\p@}{\@subsubsecfont}}
\makeatother

\begin{document}

\title{The Roles and Modes of Human Interactions with Automated Machine Learning Systems}
\subtitle{A Critical Review and Perspectives}

\author{Thanh Tung Khuat}
\affiliation{
	\institution{Complex Adaptive Systems Lab, University of Technology Sydney}
	\city{Sydney}
	\state{New South Wales}
	\postcode{2007}
	\country{Australia}
}
\email{thanhtung.khuat@uts.edu.au}

\author{David Jacob Kedziora}
\affiliation{
	\institution{Complex Adaptive Systems Lab, University of Technology Sydney}
	\city{Sydney}
	\state{New South Wales}
	\postcode{2007}
	\country{Australia}
}
\email{david.kedziora@uts.edu.au}

\author{Bogdan Gabrys}
\affiliation{
	\institution{Complex Adaptive Systems Lab, University of Technology Sydney}
	\city{Sydney}
	\state{New South Wales}
	\postcode{2007}
	\country{Australia}
}
\email{bogdan.gabrys@uts.edu.au}

\renewcommand{\shortauthors}{T.T. Khuat, D.J. Kedziora, B. Gabrys}


\begin{abstract}
	As automated machine learning (AutoML) systems continue to progress in both sophistication and performance, it becomes important to understand the `how' and `why' of human-computer interaction (HCI) within these frameworks, both current and expected. Such a discussion is necessary for optimal system design, leveraging advanced data-processing capabilities to support decision-making involving humans, but it is also key to identifying the opportunities and risks presented by ever-increasing levels of machine autonomy. Within this context, we focus on the following questions: (i) How does HCI currently look like for state-of-the-art AutoML algorithms, especially during the stages of development, deployment, and maintenance? (ii) Do the expectations of HCI within AutoML frameworks vary for different types of users and stakeholders? (iii) How can HCI be managed so that AutoML solutions acquire human trust and broad acceptance? (iv) As AutoML systems become more autonomous and capable of learning from complex open-ended environments, will the fundamental nature of HCI evolve? To consider these questions, we project existing literature in HCI into the space of AutoML; this connection has, to date, largely been unexplored. In so doing, we review topics including user-interface design, human-bias mitigation, and trust in artificial intelligence (AI). Additionally, to rigorously gauge the future of HCI, we contemplate how AutoML may manifest in effectively open-ended environments. This discussion necessarily reviews projected developmental pathways for AutoML, such as the incorporation of reasoning, although the focus remains on how and why HCI may occur in such a framework rather than on any implementational details. Ultimately, this review serves to identify key research directions aimed at better facilitating the roles and modes of human interactions with both current and future AutoML systems.
	
\end{abstract}

\maketitle

\section{Introduction}\label{intro}

Broad interest in machine learning (ML) has ebbed and flowed ever since the 1950s, but recent years have arguably witnessed a new phase in the history of the field: an unprecedented level of technological uptake and engagement by the mainstream. From deepfakes for memes to recommendation systems for commerce, ML has become a regular fixture in broader society.
Unsurprisingly though, this ongoing transition from purely academic confines is not smooth; the general public does not have the extensive expertise in data science required to fully exploit the capabilities of ML.

The ideal solution for democratisation is to make the application of ML optionally independent of human involvement. This is the primary goal of automated/autonomous machine learning (AutoML/AutonoML), an endeavour that, despite a rich multi-faceted history~\citep{kemu20}, has only truly taken off within the last decade. Popularised by significant optimisation advances applied narrowly to model selection~\citep{thhu13, swdr17, sabu18}, the scope of AutoML has since expanded to automating all aspects of an ML application. Indeed, as long as there is a will and a way, it seems inevitable that ML systems will move ever closer towards autonomy.

As of the early 2020s, much has been written specifically around mechanisms and integrated systems for automating operations in both general ML~\citep{kemu20} and the fashionable subclass known as deep learning (DL)~\citep{doke21}; the topic of mechanising the latter is abbreviated as AutoDL. These discussions have mostly taken the notion of `automation' to heart, wrestling with the challenges of how computers can make high-level decisions on their own.
However, one important topic has been left underexplored: how do humans fit into the picture? This is crucial to consider, as, no matter how far its capacity for autonomous function evolves, the purpose of an AutoML system is to support human decision-making.
Thus, perhaps counterintuitively, interactions cannot be an afterthought~\citep{amwe19}.

Even with an academic focus on model accuracy and algorithmic efficiency, systems cannot be considered optimal if they do not welcome and make use of optional human input. Moreover, beyond academia, the concept of `performant' ML becomes much more complex and user-centred~\citep{scke21}; the most promising algorithms and architectures will likely be ones that flexibly tailor outputs to satisfy a very broad set of requirements. Then there is debate on just how much autonomy ML systems should be given. While the nature of the human-system relationship may eventually become one of collaboration~\citep{wawe19}, it is unlikely that humans will ever relinquish supervisory oversight~\citep{en17}. Many researchers have echoed similar opinions, stating that human experience is indispensable and AI cannot be expected to autonomously operate in a socially responsible way~\citep{xiwu21}. For this multitude of reasons, a holistic appreciation of AutoML requires an associated study of human-computer interaction (HCI).

This is a rich topic; the nature of human interactions with AutoML, both in terms of roles and modes, has evolved and will continue to evolve alongside developments in the field. Consider, as an analogy to those developments, the history of artificial intelligence (AI) with respect to the game of chess. In the late 1960s, Mac Hack became the first chess program to play in human tournaments and even score a victory in doing so~\citep{grea67}. Automated but heavily reliant on domain knowledge -- it incorporated approximately 50 expertise-based heuristics -- and hardly a threat to human dominance in chess, Mac Hack can be likened to proto-AutoML model-recommendation systems that were developed prior to the 2010s \citep{va11, seva13}: novel and impressive for the time, but severely limited. Eventually though, by 1997, the swell of computational resources and advances in algorithmic techniques enabled a chess-playing computer known as Deep Blue to defeat a reigning world champion \citep{caho02}. As with the new wave of hyperparameter-optimising AutoML systems in the 2010s~\citep{thhu13, swdr17, sabu18}, themselves becoming more and more capable of scaling competition leaderboards~\citep{ermu20}, Deep Blue heralded an era where computers would be far more competent than humans at performing a specific task.

Notably, even during the famed contest of 1997, Deep Blue was far from autonomous, leveraging a human-prescribed database of openings and endgame meta-knowledge, while also being manually adapted by grandmasters between games. Only in 2017, with the initial release of AlphaZero~\citep{zhyu20}, has human input almost completely been removed, with an AI system autonomously learning to dominate humans in chess via self-play. In fact, the newest generation of chess-based AI has started to shift the roles of humans from mentors to students, with, for instance, an AI propensity for `h-pawn thrusts' giving high-level players pause for thought~\citep{miya20}. The field of AutoML has not yet reached the same level of autonomy\footnote{The AlphaZero approach is shared by AutoML systems such as AlphaD3M~\citep{drkr18}, but, for the purpose of the chess analogy, the two are not equivalent; automation in chess is significantly more advanced than automation across an entire ML application.}, but it is nonetheless worth asking: is this the state of interactions to plan for in the future? Will AutoML eventually produce more insight on how to solve an ML task than it currently receives?

\begin{figure}[!ht]
	\centering
	\includegraphics[width=\textwidth]{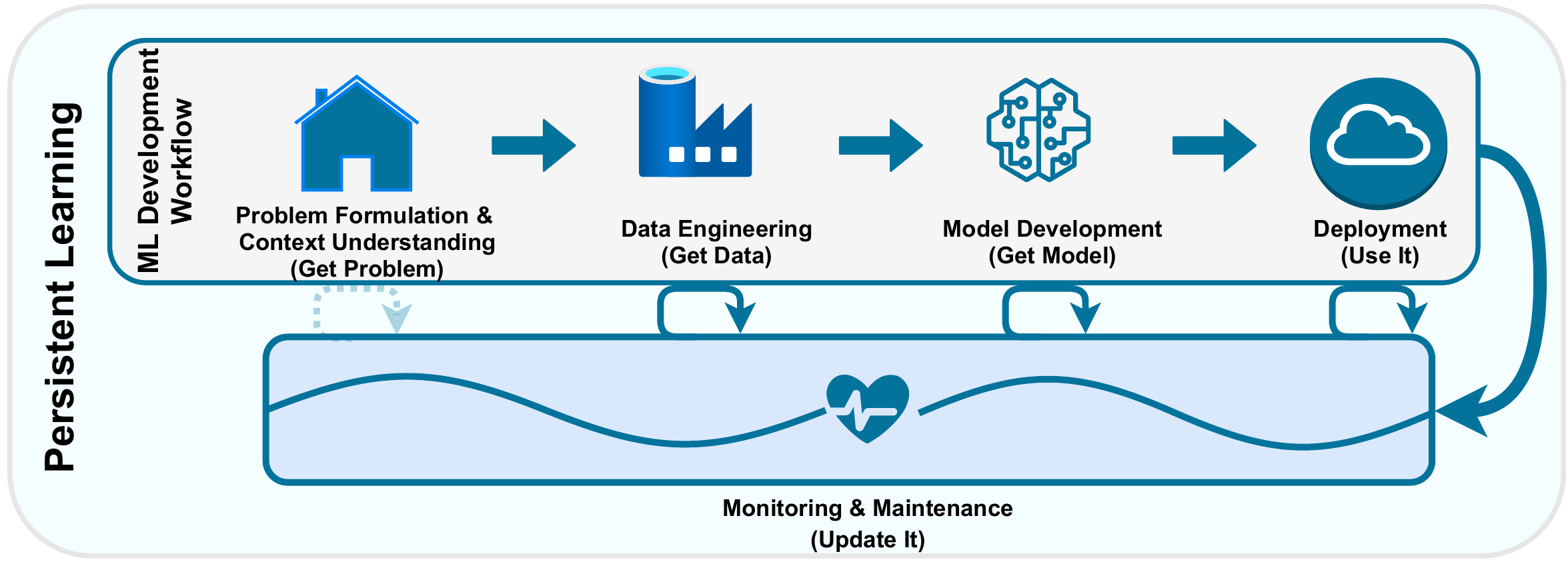}
	\caption{General schematic for a machine learning (ML) workflow, i.e.~the operations involved in producing and maintaining an ML model for an ML application.}
	\label{ml_workflow}
\end{figure}

A comprehensive overview of HCI in AutoML, both current and prospective, needs to be carefully structured. For instance, human involvement in ML applications can be partitioned into two categories: productive and consumptive. While the latter refers to how end-users engage with and benefit from an ML model, the former relates to how such a model comes about. These `productive' practices can be codified in many ways into an `ML workflow' \citep{chcl00, stbu21}, but one particular representation~\citep{kemu20, doke21, scke21} is schematised in Fig.~\ref{ml_workflow}.

Within the depiction of an ML workflow, it is clear that there are several phases of operations involved in developing, deploying and maintaining an ML model. Of these, the model development stage receives the most focus in academic AutoML research, especially in the case of DL and neural architecture search (NAS)~\citep{doke21}, but there have been numerous automation efforts applied to the rest of an archetypal ML workflow. In fact, the capacity to continuously monitor performance and adapt to dynamic changes in data environments has previously been highlighted as a key prerequisite for the transition between AutoML and AutonoML~\citep{kemu20}. Relatedly, there exist many theoretical proposals for supporting online learning~\citep{lamo18}, and initial experimental research towards making AutoML systems `persistent' has recently been undertaken within academia~\citep{bafa21, ceva21}. Meanwhile, in industry, the emerging trend of `MLOps' reflects the importance of automated deployment to real-world demands~\citep{scke21}. In essence, any developers interested in engineering a comprehensive AutoML/AutonoML system must understand the idiosyncrasies of each and every workflow phase, e.g.~the associated formats of human inputs/guidance, the minimal requirements for baseline operations, the possible opportunities for additional human-assisted learning, and so on.

Alternatively, rather than partitioning HCI in AutoML by when an interaction occurs, it is sometimes more informative to consider who is undertaking the interaction. This perspective is particularly natural in commerce and industry~\citep{scke21}, where, beyond the end-users that consume the outputs of an ML model, there are typically numerous stakeholders linked to the production of a model. These may include technicians in the form of data scientists or software developers, business staff in the form of project managers or domain experts, regulatory groups in the form of third-party auditors or government agencies, etc.

Importantly, the obligations and interests of different stakeholders are usually not mappable to individual stages within the ML workflow depicted in Fig.~\ref{ml_workflow}. Moreover, the modes of their interactions may also vary substantially. Some roles will demand fine control over AutoML processes, while others will simply require an entry point for inputs. Some roles will desire a window into the mechanisms involved, while others will only want to be alerted if things go wrong. Whatever the case may be, these requirements must be considered at the fundamental level of algorithm and architecture. It is not optimal to focus solely on predictor accuracy and efficiency during system design, only to service any remaining real-world expectations with a hasty patch-up.

Crucially, it is worth emphasising that the concept of a `user' is intrinsic to the stakeholder view of AutoML.
This may seem like an unnecessary distraction to those that are solely interested in improving the statistical theories underlying ML algorithms, but that attitude ignores the greater ecosystem in which ML operates: human decision-making. For instance, humans may be willing to tolerate disliking 40\% of AI-recommended music, whereas a 20\% false-positive rate for AI-recommended convictions is arguably abysmal.
Simply put, human context matters more than any agnostic accuracy metric. It follows then that successfully translating ML model performance into real-world outcomes is contingent on a set of engagement-related requirements \citep{ardi20, shin21, drwe20, scbi19, ehli21}, which we will bundle here under the title of `user experience' (UX). This includes topics that have recently diffused into academic discussions around ML and AutoML, such as accessibility, transparency, fairness, reliability, and so on~\citep{xiwu21}.

Accordingly, the concept of a user interface (UI) -- the implementation need not be monolithic -- becomes particularly important to AutoML within the stakeholder perspective, as this is where UX can most directly be managed. Indeed, designing intelligent UIs is critical for supporting human-guided AutoML~\citep{giho19, lema19}, where a technical user might, as an ideal, tweak problem settings, explore data characteristics, limit model search spaces, etc. These interactions may also be constrained or facilitated by the manner in which they occur, e.g.~via touch-screen, voice commands, gesture recognition, or even brain signals \citep{xiji13}. In short, the field of AutoML would be well served by greater discussions around the concept of interfacing so that, beyond simply enabling the control of ML operations, users can both inject domain knowledge and extract comprehensible information with ease.

Turning to factors that influence UX, explainability is high up on the list.
This is especially a challenge for AutoML, as the core principle of automation is to decouple humans from certain operations.
It may seem wasteful then, if not counterproductive, to spend research effort in making those processes transparent, consequently encouraging humans to reinvolve themselves. Sure enough, many current AutoML tools are staunchly black-box systems~\citep{xiwu21}, veiling how ML models are built and how predictive/prescriptive outputs are generated. But herein lies the nuance; the aim of AutoML is to remove the \textit{necessity} for human engagement, not the option. Thus, technical obscuration actually impedes ML performance if users cannot understand how to properly insert domain knowledge that would otherwise be beneficial to an ML task~\citep{liwa17}. This is especially a drawback at the current point in time, because human-in-the-loop learning is still often more advantageous than machine-centric ML~\citep{tako16}.

Regardless, even if AutoML systems were to be completely autonomous, their innards untouched by humans, explainability is also necessary to promote trust~\citep{mada19}. Surveys indicate data scientists tend to be sceptical of ML models provided by AutoML tools if there are no mechanisms for transparency and understandability~\citep{drwe20}. Similarly, end-users only follow ML recommendations if the system can show the reasoning behind them~\citep{wani21}. This reticence by people to use results they cannot understand or explain can be frustrating for simple business applications, but it is completely warranted in high-stakes contexts~\citep{ru19}, including medical diagnosis, financial investment and criminal justice. To do otherwise could be disastrous. For example, adverse outcomes have been linked to the COMPAS recidivism prediction model~\citep{dime16}, the BreezoMeter real-time air-quality prediction model used by Google during the California wildfires in 2018~\citep{mc18}, and black-box medical diagnosis models in general~\citep{dujo21}.

Another factor that affects UX, even and especially if stakeholders are not directly aware that they are `using' the results of ML, is fairness. This socially conscious requirement has recently been taken up as an important issue by academic research~\citep{za16, caha20, memo21}, indicating just how far ML has embedded itself into the mainstream, and recognises that predictive/prescriptive accuracy and error may disproportionately affect different people in different ways. Now, granted, there have been efforts towards automating mechanisms for discovering and preventing discrimination in ML models~\citep{habo16}, but the challenge is that there are many possible technical definitions for fairness, often orthogonal and sometimes contradictory~\citep{veru18, sahu19}. Once again, human context matters.
So, it is an open question as to how human oversight can best be integrated within an AutoML system, enforcing ethical requirements upon a mechanised process.

Of course, while every ML algorithm applies its own assumptions, many `unfair' biases are often sourced from biological neurons, i.e.~human brains. These can be injected into learning systems via data and knowledge, manifesting in both information content and sampling.
As a result, human cognitive biases that are internalised can lead to a deterioration in model reliability, and there are many high-profile examples of this occurring~\citep{hu16, ze21}. The severity of these impacts can also vary depending on context. Healthcare is one example of a high-stakes environment, where cognitive biases in clinical practice can have a strong influence on medical outcomes~\citep{sare16, pr19}. Indeed, similar flaws in predictive systems have been shown to hinder social minorities from receiving extra care services~\citep{obpo19}. Accordingly, there is an imperative to more thoroughly consider the nature and implementation of bias mitigation strategies within AutoML.

Fundamentally, the point of all this discussion is that, given a suitable conceptual framework, such as the dual workflow/stakeholder perspective of ML operations, it is possible to engage with many HCI-related issues that, unaddressed, will impede the heretofore surging pace of progress in AutoML. Moreover, this kind of systematic approach does not just clarify the current state of HCI in AutoML; it provides a lens through which the future of this trend in ML can be forecasted. This does not mean speculating on detailed implementations of HCI-related mechanisms, but rather understanding the projected evolution of human-system interactions, particularly as algorithms and architectures become better at their job. Thus, the aforementioned chess analogy remains useful in illustrating this progression as AutoML systems shift further along the spectrum of autonomy~\citep{sifr21}.

However, it is still valuable to conjecture just a little bit further.
AlphaGo~\citep{sihu16} and AlphaZero~\citep{sihu18} are extremely competent at their respective games, but they remain constrained within particular environments. An equivalent AutoML system would essentially be autonomous for every phase of the ML workflow in Fig.~\ref{ml_workflow} except for one holdout: problem formulation and context understanding. Such a constraint is not unexpected, as this phase is likely to be the last bastion of necessary human involvement in ML. Unfortunately, it does stand in the way of numerous AI applications. For instance, there exists plenty of research and development in the field of autonomous vehicles~\citep{haw18,leha20}, yet the challenge of operating in unpredictable and effectively boundless driving environments remains, to date, daunting~\citep{hama20}.
Nonetheless, without delving into the domain of artificial general intelligence, these constraints will eventually relax.
The novel MuZero system~\citep{scan20} is already emblematic of an emerging reinforcement-learning approach that can be applied agnostically to a variety of games with diverse rules, autonomously building competent models from first principles. In theory, cognitive models may eventually supercharge this process even further, enabling ML systems to efficiently transfer knowledge from one problem to another by actually \textit{understanding} context, rather than by outright ignoring it. So, as AutoML truly becomes AutonoML, and then begins to relax into open-world learning: will human-system interactions change yet again?

As is evident by now, there are many important issues to consider in the overlap between AutoML and HCI. This review addresses these topics, marking the final part in a series of monographs dedicated to a systematic and conceptual overview of AutoML~\citep{kemu20, doke21, scke21}. Specifically, with the series having previously focussed on how computers may perform ML/DL in the absence of humans~\citep{kemu20, doke21}, this work aims to recontextualise AutoML/AutonoML back within the ecosystem of human decision-making.
In fact, because ML does not operate in a vacuum within the real world, some organically arising consequences of this interlinkage are already captured by the previous technological survey in the series~\citep{scke21}.
This review, however, dives much deeper into the fundamentals of HCI in AutoML, driven by the following set of questions:
\begin{itemize}
	\item How does HCI currently look like for state-of-the-art AutoML algorithms, especially during the stages of development, deployment, and maintenance?
	\item Do the expectations of HCI within AutoML frameworks vary for different types of users and stakeholders?
	\item How can HCI be managed so that AutoML solutions acquire human trust and broad acceptance?
	\item As AutoML systems become more autonomous and capable of learning from complex open-ended environments, will the fundamental nature of HCI evolve?
\end{itemize}

To best grapple with these questions, the rest of this monograph is structured as follows. Section~\ref{role_inter_current_automl} examines HCI and state-of-the-art AutoML as of the early 2020s. It does so with respect to the workflow/stakeholder perspectives of AutoML, after these are first systematised. Modern approaches to UIs and current concerns around UX, e.g. in terms of explainability and fairness, are also surveyed.
Section~\ref{roles_inter_cmpl_constrained_env} then extrapolates progress in AutoML to where associated systems are effectively autonomous in all high-level ML operations, excluding problem formulation and context understanding. The evolution of HCI with respect to genuine high-performance AutonoML, albeit restricted to constrained environments, is considered. Section~\ref{roles_inter_open_env} follows by pushing this limit, relaxing restrictions and considering ML within open-ended environments. To anchor such a scenario, modern theories for incorporating `reasoning' in ML systems are surveyed and debated. Subsequently, changes to the nature of HCI with respect to these upgraded forms of AutonoML systems are theorised upon. A synthesising discussion is then presented in Section~\ref{open_issues}, identifying existing issues and potential research directions that may hinder or facilitate the successful interplay of HCI and AutoML, both now and in the future. Finally, Section~\ref{conclusion} concludes this review, summarising key findings and perspectives around the roles and modes of human interactions with AutoML/AutonoML systems.

\section{Interacting with AutoML Systems: Current Practices} \label{role_inter_current_automl}

This section is dedicated to reviewing HCI in modern practicable AutoML.
However, it is first worth distinguishing this review from the one that precedes it~\citep{scke21}.
Consider the concept of `AutoML ergonomics': the study of interactions between humans and other elements of an AutoML system, where such an understanding can be used to optimise human well-being and overall system performance.
In the context of AutoML, human well-being includes comfort, safety and productivity while running an ML application.
Under such a definition then, both reviews deal with AutoML ergonomics.
That said, the focus of the prior monograph is the mainstream technological emergence of AutoML, examining how stakeholder needs have shaped both the implementation and application of AutoML systems/packages.
Accordingly, its survey was entirely grounded in concrete products; ergonomic optimisations in HCI were identified as an organic \textit{outcome} of the technological translation from academia to industry.
In contrast, this monograph flips cause and effect.
Specifically, leaning on broader scientific literature, this review partially examines how ergonomic principles for HCI act as a \textit{driver} for designing/informing AutoML interactivity, both present and future.
Of course, the two reviews overlap when discussing the current state of AutoML, but the complementary perspectives of HCI prove illuminating.

With this complementarity in mind, we begin by reintroducing the concepts of an ML stakeholder and ML workflow in Section~\ref{stakeholder} and Section~\ref{roles_dev}, respectively.
This enables discussions around \textit{who} is interacting with modern AutoML and \textit{what} phase of an ML application their interactions concern.
We then start to examine the notion of interfacing with AutoML systems in Section~\ref{multimodality}, listing out essential functions that a UI should support, as well as the modalities in which interactions may be communicated.
This section could almost be considered a discussion of `physical' ergonomics, as it mostly concerns the comfortable and productive communication of human intent via physical means.
Section~\ref{improve_outcome} then highlights significant challenges facing HCI in modern AutoML, reviewing existing attempts to improve trust in ML solutions via efforts to support explainability and fairness.
This section could be considered a discussion of `cognitive' ergonomics, where improving the outcomes of interactions goes hand in hand with optimising the mental processes of a stakeholder, e.g.~situational awareness and decision making.
Finally, Section~\ref{user_interface} synthesises these ergonomic considerations, examining recent proposals for a list of critical requirements that an optimal AutoML UI should satisfy.
Ultimately, through these various discussions, we capture a comprehensive snapshot of how the roles and modes of human interactions with AutoML systems are currently regarded in the early 2020s.

\subsection{Types of Stakeholders}\label{stakeholder}

A popular motivation and selling point for AutoML has always been the democratisation angle, in that associated systems can allow `anyone' to run an ML application, including lone hobbyists and laypeople without data-science expertise.
However, most automated learning systems are embedded deeply within human social environments, with the expectation that various groups of humans will interact and collaborate with these systems~\citep{ehli21}.
Thus, while most AutoML products still lean towards single-user interfaces, it is necessary to consider a broad range of stakeholders, including those whose interactions are relatively indirect.
Indeed, the previous review spoke in terms of business organisations~\citep{scke21}, where each entity within an organisation has its own role and requirements concerning a learning system.
Moreover, the previous monograph categorised these stakeholders with an eye to the outcomes they seek, with the notion of `performant ML' denoting an efficient achievement of these aims.
Here, we similarly provide a classification of stakeholder types, but the focus is now on the stakeholder-system information flows that their roles require.

In condensed form, there are effectively four classes of stakeholders that can engage with an ML application, i.e.~the focus of AutoML operations.
These categories are the technical group, the business group, the regulatory group, and the end-user group.
Naturally, their purview may involve various phases of an ML application, ranging from problem formulation and context understanding to monitoring and maintenance; see Section~\ref{roles_dev}.
Nonetheless, under such a classification scheme, their roles and consequent interactions with an AutoML system are relatively well defined, presented in Table~\ref{table_stakeholder}.
We note that this does not mean real-world job titles and responsibilities need to stay confined within a particular category.
For instance, an ML expert may be a hybrid stakeholder with two parallel agendas: one, work in the technical group to guide AutoML when searching for optimal ML solutions, and two, work in the regulatory group by providing auditing feedback to a government agency about the reliability and fairness of an AutoML system.
The convenient categorisation in Table~\ref{table_stakeholder} simply systematises the interactions that an AutoML system may experience, along with the information flows required to satisfy their intent.

Before proceeding, though, we raise a nuanced point.
Most discussion in this monograph does not consider the top-down construction of an AutoML system, the standard approach in academia/industry, to be an interaction \textit{with} the system.
This assertion remains true even though certain technicians, under advisement from other stakeholder groups, are responsible for enacting this construction.
On the other hand, the \textit{configuration} of the system very much falls within the scope of HCI in AutoML, even if an ML application is not presently being run.
Consequently, one could claim that interactions with an AutoML system to configure a new default parameterisation are part of designing that system.
Specifically, depending on whether the intent of this HCI is permanent reconfiguration or problem-specific customisation, the roles of specific stakeholders can blur between AutoML design and the processing of an ML application.
To some extent, this is simply a matter of terminology; every AutoML system retains some form of a skeleton that is designed/constructed \textit{prior} to any HCI.
Even the most novel of hypothesised frameworks, such as organically emergent AutonoML/AutoDL architectures~\citep{kemu20, doke21}, are anchored somewhere with hard-coded principles.
Ultimately, given that the majority of stakeholders wish to \textit{use} AutoML, not just set it up, we acknowledge the nuance but speak of HCI mainly in terms of running ML applications on an existing system.

Turning now to the categorisation of stakeholders, the first group listed in this section is \textbf{the technical group}, including data scientists, data analysts, ML engineers, software developers, and system engineers.
Traditionally, this group is responsible for developing, deploying, monitoring and maintaining ML solutions, so, at least at present, they are likely to be closely involved anywhere an AutoML system attempts to automate any of these technical operations.
This involvement is especially vital given that ML solutions can be very complicated, often structured as `ML pipelines' of data transformations that we call `ML components'.
Indeed, no existing AutoML package is close to perfect at finding an optimal ML pipeline for every ML problem, so the technical group continues to have a hand in supporting, constraining or outright making many decisions during this process.

The functional roles of a technician can be numerous, with their inputs determining/influencing the use of data samples, feature constraints, model types, hyperparameter bounds, performance metrics, learning/adaptation strategies, computational resources, control/analysis mechanisms for fairness and explanation, and so on.
However, the core focus of these efforts is ensuring an ML solution performs technically well, e.g.~in terms of predictive accuracy and scalability.
Thus, the technical group expects to have access to the ML solution object, if it is not directly returned by an AutoML system, as well as any outputs relevant to technical performance.
Additionally, technicians may also be responsible for verifying that any of the mechanisms/features within an AutoML package are sound and fit-for-purpose, e.g.~bias mitigation or high-performance computing, even if they are not necessarily the stakeholders that will engage with their outcomes.
Finally, we stress that, while the last decade has seen significant advances in AutoML, few modern systems are genuinely fire-and-forget upon being given their inputs.
AutoML can accelerate the search for good ML models, but technicians still lie at the heart of solution-seeking, which is why so many AutoML vendors continue to offer close back-end support, even for products that support user-driven ML.
As such, the best modern AutoML systems are expected to provide feedback that can assist a technician with understanding solution space, e.g.~via model comparisons or other representations of sampled hyperparameters, so that they can better refine these optimisations.
Arguably, the value of these particular outputs may diminish somewhat as AutoML improves its automated capabilities; see Section~\ref{roles_inter_cmpl_constrained_env}.

\begin{landscape}
\centering
\scriptsize{
\begin{longtable}{p{.15\textwidth}p{.42\textwidth}p{.35\textwidth}p{.45\textwidth}}
\caption{Overview of different stakeholder groups and their roles and interactions with AutoML systems} \label{table_stakeholder}\\
\toprule
Stakeholder      & Main roles of human interactions & Human's Input & AutoML's output \\ \midrule
\makecell[l]{Technical group}  & \vspace{-\topsep}
\begin{itemize}[leftmargin=*, label=\textbullet, nolistsep, noitemsep]
\item Collect and explore data
\item Use domain knowledge to guide AutoML algorithms find and select the best learning models
\item Compare different learning models
\item Verify accuracy and performance of AutoML algorithms
\item Assess the explainability, understandability, reliability, accessibility, and transparency of solutions
\item Bias and fairness checking for the obtained learning models
\item Evaluate the security, accessibility and scalability of resulting learning models in production environments
\item Deploy, monitor, and maintain learning models
\end{itemize}  &  \vspace{-\topsep} \begin{itemize}[leftmargin=*, label=\textbullet, nolistsep, noitemsep]
\item Training data
\item Selected features/samples and feature requirements
\item Rules for data pre-processing
\item Search spaces of models and hyperparameters are narrowed down
\item Resource constraints
\item Requirements for model assessment
\item Requirements for model/hyperparameter comparison
\item Criteria for types of explanation, fairness, and transparency of solutions
\item Technical and non-technical constraints
\item Learning and adaptation strategies
\item Settings for operation environments
\end{itemize}             &  \vspace{-\topsep}\begin{itemize}[leftmargin=*, label=\textbullet, nolistsep, noitemsep]
\item Data visualisation and statistics
\item Resulting learning pipelines
\item Performance and accuracy of the obtained learning models
\item Graphical controls/ analyses for AutoML processes
\item Details of the searched learning models and hyperparameters
\item Model/hyperparameter comparison views
\item Explanations of outcomes generated by learning models in different forms
\item Outputs of metrics related to fairness
\item Reports related to functional and non-functional constraints
\item Figures related to the operation of learning models in the real time for monitoring and maintenance
\item The availability of learning models in the production environment and continuous learning and adaptation abilities in real time
\end{itemize}               \\ \midrule
\makecell[l]{Business group}   & \vspace{-\topsep}  \begin{itemize}[leftmargin=*, label=\textbullet, nolistsep, noitemsep]
\item Provide business requirements for the development and design of AutoML systems
\item Provide the acceptance criteria of AutoML systems in terms of functionalities and usefulness
\item Assess and verify the effectiveness of AutoML systems related to business aspects
\item Assess the maintainability, scalability, accessibility, and usability of AutoML systems
\end{itemize}                              &  \vspace{-\topsep}   \begin{itemize}[leftmargin=*, label=\textbullet, nolistsep, noitemsep]
\item Business requirements or constraints
\item Business success criteria
\item Non-functional constraints, e.g., maintainability, accessibility, scalability, usability, resources, and profits
\end{itemize}                       &  \vspace{-\topsep}    \begin{itemize}[leftmargin=*, label=\textbullet, nolistsep, noitemsep]
\item Reports for experimental results for business success criteria and business requirements
\item Results corresponding to each non-functional requirement
\end{itemize}                        \\ \midrule
\makecell[l]{Regulatory group} &  \vspace{-\topsep} \begin{itemize}[leftmargin=*, label=\textbullet, nolistsep, noitemsep]
\item Validate and audit the functional requirements of the systems compared to specification and design documents
\item Validate and audit the non-functional requirements of AutoML systems such as security, safety, reliability, transparency, fairness, ethical issues, social norms, law compliance, and usefulness
\end{itemize}                                   & \vspace{-\topsep} \begin{itemize}[leftmargin=*, label=\textbullet, nolistsep, noitemsep]
\item Test cases/Input data to evaluate and audit the functional and non-functional requirements for resulting learning systems
\end{itemize}  &   \vspace{-\topsep}  \begin{itemize}[leftmargin=*, label=\textbullet, nolistsep, noitemsep]
\item Reports or experimental results for input test cases
\item Views to show reasons or processes to achieve the outcomes for test cases
\end{itemize}            \\ \midrule
\makecell[l]{End user}         &  \vspace{-\topsep} \begin{itemize}[leftmargin=*, label=\textbullet, nolistsep, noitemsep]
\item Use and interact with the deployed learning models
\item Analyse and make decisions based on the recommendations provided by AutoML models
\item Control and supervise the operating procedures of the AutoML system
\item Planning and goal management
\end{itemize}   &  \vspace{-\topsep} \begin{itemize}[leftmargin=*, label=\textbullet, nolistsep, noitemsep]
\item Provide the real inputs for the AutoML system via user interfaces
\item Provide feedback to the AutoML system
\end{itemize}   &  \vspace{-\topsep} \begin{itemize}[leftmargin=*, label=\textbullet, nolistsep, noitemsep]
\item Outcomes for each input data
\item Show reasons or explanations for outputs
\item Continuous learning and adaptation to user behaviours and feedback
\end{itemize}             \\ \bottomrule
\end{longtable}
}
\end{landscape}

The second group listed in this section is \textbf{the business group}, including business analysts, domain experts, and project managers.
While the technical group ensures that the mechanics of an ML application run smoothly, business stakeholders are the ones that embed the AutoML machinery within real-world contexts and problems.
Thus, they mainly contribute to either end of an ML workflow, translating business requirements and sourcing raw data for application inputs, then verifying the suitability of solution outputs for use and deployment.
After all, a quick and accurate ML model is not helpful if it solves the wrong problem or cannot be leveraged for consequent impact.
At a higher level, some business stakeholders are also responsible for deciding whether an AutoML system is worth using in the first place, but, to date, such decisions are either made prior to interacting with the system or in response to running the product.
In the latter case, assessing the value of an AutoML package is often equivalent to assessing the experiential/solution quality of one or more processed ML applications.

In terms of inputs, the business group will typically intend to provide real-world criteria and constraints into an AutoML system.
Ideally, these stakeholders should also acquire business-relevant feedback on the performance of both the AutoML system and the ML solutions it produces/deploys.
Such metrics will typically involve operational costs, e.g.~electricity bills.
Additionally, many businesses will be particularly interested in the final outcomes of an ML application, e.g.~profit from user subscriptions to an ML-driven product, disease cases averted by an epidemiological model, etc.
Such information is more challenging to provide as it requires AutoML services to embed and monitor an ML solution flexibly in a customisable deployment environment.
This caveat leads to an important point; we stress intent and ideals when defining business inputs and outputs.
At the current time, albeit with a growing number of exceptions~\citep{scke21}, most AutoML packages do not provide an easy way to blend a real-world business context with the abstracted mechanics of ML.
It is thus still a common expectation that technical stakeholders serve as a bridge for many interactions between the business group and an AutoML system, e.g.~filtering data inputs by business constraints or interpreting scalability/maintainability requirements as financial costs.
However, should the field of AutoML eventually discover/implement mechanisms for autonomous context understanding, the notion of a technician as an intermediary may become defunct; see Section~\ref{roles_inter_open_env}.

The third group listed in this section is \textbf{the regulatory group}, including governance staff, third-party auditors, and government agencies.
The growing importance of the group reflects the fact that, as ML has become a practicable technology widely adopted by industry, many issues have become unveiled when applying abstract mechanistic ML processes within the nuanced complexity of real-world environments.
Human societies, which ideally have access to the characteristic of empathy, already struggle to quantify the qualities of human existence when making ethical decisions.
Thus, integrating a black-box AutoML system into this decision-making process can be high-stakes and induce more detriment than benefit, even if, by some technical criterion, it is fast and accurate.
Simply put, modern AutoML, as with the ML processes it wraps around, faces issues of trust; see Section~\ref{improve_outcome}.
Consequently, the role of regulatory stakeholders is to ensure compliance around specific social requirements: user safety, model reliability, outcome fairness, cybersecurity, legal adherence, societal expectations, ethical standards, etc.

Now, ML and the automation of its high-level processes are very much frontier technologies, at least with mainstream industry as the target.
Naturally, as with the history of the internet, legislation tends to lag behind technological capacity, so there are currently no standardised regulatory practices or associated entities for the general industrial use of ML, let alone AutoML.
Thus, at present, the focus of the archetypal auditor is solely on the final outcomes of an ML application, much like with the business group.
Typically lacking data-science expertise, they are likely to assess an AutoML system indirectly and post hoc.
Because of this, technical stakeholders do, in practice, take on hybrid responsibilities shared with this group, directly probing ML solutions for technical outputs related to, for example, model bias.
That said, many AutoML vendors have recently invested quite a bit in this space, so there are presently an increasing number of accessible interfaces for regulatory outputs~\citep{scke21}.
In fact, with the way the technology is evolving, the notion of an AutoML auditor may itself eventually arise, testing the suitability of a system for particular classes/domains of ML problems.
In such a case, the regulator would expect not only to interact with informational outputs, e.g.~as part of traditional reactive regulation, but also to provide inputs in the form of test cases, i.e.~proactive regulation.
As a concrete example, consider an epidemiological ML model that optimises a health budget to minimise infection within a population.
A simple scenario test introduced by an auditor may reveal that the ML model removes medical support for the infected, thus `solving' an epidemic by accelerating deaths, which is problematic.
The value of scenario modelling to the regulatory group is clear, and, as surveyed previously~\citep{scke21}, some AutoML services already provide this functionality.

The last group listed in this section is \textbf{the end-user group}, covering any stakeholder that interacts with an AutoML-produced ML solution.
Any associated inputs and outputs regarding an ML model can be abstracted as queries and responses, respectively.
Crucially, end-users do not typically engage with an AutoML framework directly, at present, unless the system wraps around an ML solution or provides an integrated environment for model use.
Even so, it is difficult to concisely summarise interactions for end-users as ML products have so many different use cases.
For instance, it is entirely possible that, with a static ML solution produced by and detached from an AutoML system, its end-user never engages with the solution factory at all.
This disconnect is especially true of users that do not even directly interface with the ML model, e.g.~communities that unknowingly benefit from an ML-based epidemiological decision-maker.
Alternatively, as an example of an often overlooked interaction, many commercial ML-based products require installation on various computational environments, which tends to involve an end-user.
This process could potentially be managed/optimised by an AutoML deployment mechanism, although we are not currently aware of such a feature being supported for widespread use.
It is also possible in certain situations for end-users to receive outputs \textit{about} an ML model, e.g.~simplified explanations detailing why a prediction was made.
Such outputs become increasingly desirable if the role of the end-user starts to blur with the business group, e.g.~a client driving an ML application.

The nature of end-user interactions can become even more complex once the static assumption for ML models is discarded.
If an AutoML system is responsible for maintaining and monitoring an ML solution, i.e.~AutonoML, and if the inputs of an end-user are driving ML model updates and adaptation, then it follows that the user is, perhaps unknowingly, interacting with the AutonoML machine.
Such commentary is perhaps better relegated to the discussion of idealised fully automated systems in Section~\ref{roles_inter_cmpl_constrained_env}, given that there are no popularised ML products powered entirely by an AutoML back-end, let alone an autonomously adapting one.
Nonetheless, it is worth highlighting that end-users do already impact ML models via active learning, e.g.~by correcting a spam-email filter.
Whether or not these adaptive mechanisms should be considered under the umbrella term of AutoML when they do not involve certain core functionality, such as model search, is debatable.
However, it remains clear that end-users, as the intended beneficiaries of an ML model, greatly influence an AutoML system, whether this influence is relayed by the other stakeholder groups or provided automatically via direct/indirect feedback.

\subsection{Roles and Modes Within the Machine Learning Workflow}\label{roles_dev}

While the previous section demonstrates a simple categorisation of stakeholders that interact with AutoML by their roles and motivations, it also shows how difficult it is to concisely capture every interaction and associated input-output flow that each role requires.
This challenge is partially due to the fact that a complete end-to-end ML application involves many different processes.
Thus, to better grapple with the diversity of expected interactions, this section expounds a perpendicular perspective of HCI anchored in the conceptual framework of an `ML workflow', briefly introduced in Section~\ref{intro}, which has also underpinned our three other AutoML focused reviews~\citep{kemu20,doke21,scke21}.

\begin{figure}[!ht]
	\centering
	\includegraphics[width=1\textwidth] {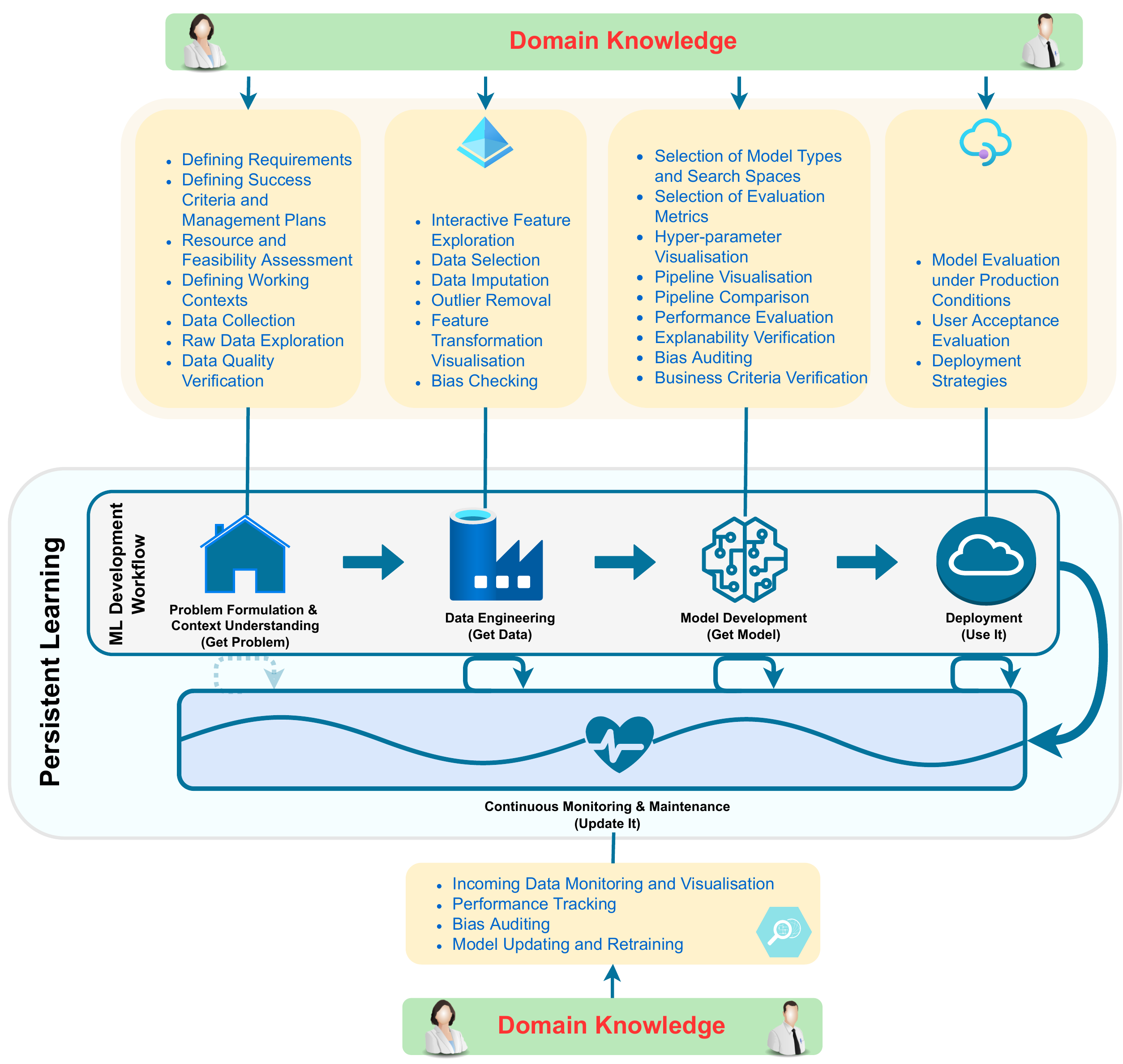}
	\caption{Summary describing the roles and modes of human interactions at various phases of an ML workflow.}
	\label{Fig_2}
\end{figure}

As an initial overview, an ML workflow comprises of five phases, which we discuss as follows: problem formulation and context understanding in Section~\ref{workflow_problem}, data engineering in Section~\ref{workflow_data}, model development in Section~\ref{workflow_model}, deployment in Section~\ref{workflow_deployment}, and continuous monitoring and maintenance in Section~\ref{workflow_maintenance}.
Their typical chronological arrangement is displayed in Fig.~\ref{Fig_2}.
Historically, loosely speaking, the automation of high-level ML operations began with hyperparameter optimisation (HPO) in the model development phase before gradually extending out along the ML workflow.
Indeed, in the years since an ICML-2014 workshop\footnote{https://sites.google.com/site/automlwsicml14/} formally introduced the term `AutoML', elements of data engineering have been absorbed into the endeavour, while, most recently, the notion of `MLOps' has similarly brought attention to automated deployment~\citep{scke21}.
Problem formulation and context understanding are, however, virtually untouched by AutoML, while monitoring and maintenance, prerequisites for autonomous lifelong learning~\citep{kemu20}, are also nascent areas of research and development.
Frameworks that attempt to handle true AutonoML are scarce, although one proposed general-purpose architecture~\citep{kaga09}, experimented with in later studies~\citep{kaga09a,kaga09b,kaga09c,kaga10,kaga11,baga17,sabu18,bafa21}, is an early pioneering example of a successful system attempting to automate the majority of an ML workflow.

Crucially, most academic research into AutoML is driven by the desire to process ML operations independently of human interaction.
It has not been a typical priority in this setting to facilitate HCI.
Only very recently has there been a shift back towards opening up an architecture, driven by the technological emergence of AutoML in the mainstream.
There is now a growing expectation that humans should have the ability to configure/control AutoML processes at all phases of an ML workflow or at least wherever an AutoML package is designed to operate.
The range of these possible interactions are broad; a summary of where humans make decisions during an ML application, both optionally and necessarily, is included within Fig.~\ref{Fig_2}.
These interactions are associated with particular portions of the workflow, and we now proceed to discuss each phase individually.

\subsubsection{Problem Formulation and Context Understanding}
\label{workflow_problem}

The first phase of the ML workflow covers problem formulation and context understanding.
Its purpose is to translate stakeholder aims, typically the goals of a business, into a concrete ML task with clearly defined objectives.
This translation of a real-world problem is a crucial step, given that the core of an ML application is an ML model/algorithm, which attempts to learn, i.e.~approximate, some desirable mapping between query and response space~\citep{kemu20}; the conversion must be appropriate.
As such, this phase not only includes the collation of raw data sources, assessing their quality for driving a learning process, but also involves verifying that an intended project is feasible to approach with ML techniques~\citep{stbu21}.
Technical and business stakeholders are expected to collaborate closely at this stage, fostering a high-level understanding of the problem context and its associated data environment, often undertaking exploratory analysis and, if proactive regulatory stakeholders are also involved, identifying and eliminating bias factors.
Of course, not all related processes during this preparatory phase are data-specific.
Even beyond establishing the requirements and scope of an ML project, e.g.~defining success criteria from both domain and organisational perspectives that ML solutions have to meet, many logistical tasks remain.
These include the formation of development and risk management plans, the examination of prior art to assess novelty and feasibility, the general allocation of human and computational resources, the establishment of collaboration/communication practices among team members, etc.

By and large, existing AutoML frameworks have not focussed on this phase, instead assuming that system operators will have a well understood and narrowly defined ML task all ready to process.
This current state of technology is not unexpected, as the second sub-phase of context understanding is immensely challenging to mechanise, discussed further in Section~\ref{roles_inter_open_env}, while automating the first sub-phase of problem formulation requires installing AutoML with `individual agency'; this veers into topics that are presently science fiction.
Nonetheless, in the absence of mechanised formulation/understanding, several AutoML services exist that support humans in developing their own context understanding, e.g.~by automating algorithms for exploratory data analysis.
Likewise, from the application planning perspective, certain vendors do embed mechanisms to assist collaborative engagement with an ML project, e.g.~team communication channels and version control.
These have been noted as part of the preceding survey~\citep{scke21}.

It is well known that visualisation plays a crucial role in this phase of an ML application, particularly when generating and communicating the results of exploratory data analysis~\citep{pesh08}.
Automated visual analytics can clarify the structure and format of big/complex data, identifying missing values and other issues.
More importantly, these HCI mechanisms can efficiently bring the statistical properties and shallow knowledge content within a data source to the eyes of a stakeholder~\citep{no06}.
In accordance with cognitive ergonomics, this lightens their mental workload, particularly with understanding a problem context, and allows greater productivity in setting up an ML project.
Naturally, there are many options to assist these visualisation processes, including software libraries like matplotlib~\citep{hu07}, toolkits like InfoVis~\citep{fe04}, and other commercial systems such as Power BI or Tableau.
How well these options integrate with existing AutoML systems differs from one to the next.
As an additional side note, while the outputs of visual analytics are, by definition, visual, the inputs for HCI can vary; see Section~\ref{multimodality}.
Vizdom is an example of a progressively interactive system that can be interfaced with via pen and touch~\citep{crga15}.

As aforementioned, collaboration between stakeholders is vital for most sub-tasks within this phase, which remains true of exploratory data analysis.
Business domain experts and data-science technicians must reach a consensus on the data/requirements to propagate further along an ML workflow, especially as they often do not share the same perspectives due to their different roles and responsibilities~\citep{mawa19}.
Interactive visual analytic tools on electronic whiteboards, such as Vizdom~\citep{crga15} or Northstar~\citep{kr18}, reflect one approach for facilitating a cooperative strengthening of context understanding.
When different stakeholder groups are, in concert, able to transform, visualise and analyse various characteristics of raw data, this can leave them better informed about the types of questions to seek ML-based answers to, as well as which features and subsets of raw data to supply for subsequent phases of an ML workflow.
The pooling of domain knowledge can also identify which data sources are low-quality, irrelevant or, worse, counterproductive when relying on ML to seek informative patterns.
Indeed, the early detection of unstable entries, abnormal data distributions and other undesirable outliers can be handled via manual/automatic exclusion~\citep{stbu21}, reducing the risk of poor ML performance.
Importantly, all data-specific operations at this stage involve only a rough form of curation; serious data engineering is left for the next stage of an ML application.

Finally, beyond establishing organisational practices to tackle an ML-solvable problem and curating/exploring relevant data, the preparatory period is also the time to make other ML-specific decisions; this is equivalent to configuring an AutoML system ahead of running an ML application~\citep{yawa18}.
Specifically, constraints can be set within the architecture in anticipation of what the business goals require.
For instance, if explainability is important, model search may be configured to eschew DL networks in favour of decision-tree variants.
As another example, a meta-learning module can be activated if prior experience is judged helpful for a new ML problem.
Technically, such interactions are associated with later phases of the ML workflow, especially configuration-based refinements, but it is still worth noting that, chronologically, an initial approach for solving a problem arises in tandem with its formulation.

\subsubsection{Data Engineering}
\label{workflow_data}

The second phase of the ML workflow covers data engineering.
Its purpose is to produce and provide high-quality data that allows ML algorithms to effectively learn desirable patterns via the formation of ML models.
This phase can be revisited many times during an ML application if subsequent phases detect any issue in the data or otherwise seek refinements.
Importantly, this is also the first stage of an ML workflow that can be heavily mechanised~\citep{kaga09,sabu16,sabu18,fieh21}, e.g.~in terms of data imputation and feature selection.
However, at the current time, AutoML tools are not capable of capturing domain knowledge, which still leaves human interaction indispensable for various important aspects of data cleaning, data routing and feature engineering~\citep{xiwu21}.
Admittedly, while domain experts were somewhat more important in Section~\ref{workflow_problem} for the curation of raw data and its initial exploration, technicians come to the fore at this point, with many data engineering processes more focussed on constructing inflows suited to the abstract and theoretical requirements of ML algorithms.
Nonetheless, collaboration is ideally maintained, and HCI can involve the control and visualisation of many processes: data imputation, sample/feature selection, feature transformation, outlier removal, etc.
With regulatory oversight, this is also an opportune time to check and mitigate bias that may creep in as part of the transformative process.

We re-emphasise, as discussed within Section~\ref{workflow_problem} in terms of context understanding, that visualisation is invaluable for allowing stakeholders to mine data for insights.
Interactive tools, e.g.~Vizdom~\citep{crga15} and Northstar~\citep{kr18}, continue to facilitate easy engagement that allows for the comprehension of data and the formulation of additional hypotheses~\citep{cahu19}.
Granted, the roles of the interaction are slightly different, as stakeholders are no longer exploring data from a business standpoint but a technical one.
For this reason, it is possible to imagine much less input being necessary in this phase than during problem formulation, and many automated techniques have already been developed to assist in, for example, feature selection~\citep{chsa14}.
In fact, the goal of the problem formulation phase is to manifest a workable ML task, so, with the fully automated systems idealised in Section~\ref{roles_inter_cmpl_constrained_env}, human control should theoretically no longer be required in the development of an ML model.
Simply put, an ML application should be \textit{runnable}, in a minimalist manner, without human input.
However, for an ML application to perform \textit{well}, data engineering is required, and, in practice, modern AutoML still requires much oversight in terms of validating transformations, avoiding spurious correlations, and so on.
Visualisation is also beneficial when combined with explanation approaches~\citep{samo19} and interpretability metrics~\citep{scbi19}.
Lessons learned from practical case studies indicate that interpretability and explainability are critical requirements for measuring data quality~\citep{fieh21}.

Turning to tasks involved within the data engineering phase, we note that feature engineering is a hefty one, which, as argued elsewhere~\citep{yawa18}, can be subdivided into two steps: generating features from raw data and improving their discriminative nature, i.e.~ensuring that there is a high level of mutual information between feature and target variables.
The first of these two steps is very difficult to mechanise, given that the relevance of features to an ML problem depends highly on the context understanding developed in the previous phase of the ML workflow.
For instance, a computer has no intuitive reason to convert a raw date into a day of the week, yet humans know well that many behaviours correlate with the weekly cycle; this relatively complex transformation can immediately provide useful information to an ML application.
On the other hand, other simple transformations involving algebra or conditional logic, to boost discriminative capability, are easy for a machine to explore, e.g.~thresholding a numeric `day of the week' variable to serendipitously convert into a binary `is weekend' variable.
Thus, AutoML research has mainly targeted this latter step of feature enhancement~\citep{yawa18}.
Even then, modern mechanisms for automated feature engineering, while substantially explored~\citep{kemu20}, are far from perfect, and, in practice, even this step involves heavy interaction from technicians, with useless/unstable features often discarded by hand.

Of course, technicians are involved in many more sub-tasks related to data engineering.
For instance, as with features, i.e.~the columns of a traditional dataset, instances of data, i.e.~the rows, must also be selected carefully.
Supplying a model with the best sampling of training data requires careful assessment to eliminate uninformative noise.
Histograms, box plots, scatter plots and magnitude-shape plots~\citep{dage18} are all popular graphical tools to detect outliers.
Similarly, attention from data scientists may also be required to handle missing values.
If the relevant data instances are not excluded, presumably with additional domain-expert input, stakeholders must decide on an appropriate method for data imputation.
Depending on data types and learning purposes, missing entries can be imputed by relying on mean, median or prediction values.
These values can be generated by, for instance, learning models~\citep{bisa18}, matrix factorization~\citep{agts13} or a technique involving multiple imputations~\citep{mu18}.
Often, there is a risk of introducing inappropriate artifacts into training data via these techniques, and, in many cases, these will not be identifiable until later phases of the ML workflow, i.e.~after a model has been run.
Technical stakeholders thus need to execute and compare the results generated by different imputation techniques to select the best approaches for specific problem contexts.
Accordingly, it is not uncommon for AutoML packages to facilitate HCI for data review and, where appropriate, the modification of imputed values with domain knowledge.

Finally, it is worth stating that human input does not inherently improve a learning system, especially when humans are responsible for curating data and designing AutoML subsystems.
Problem formulation and data transformation frequently infuse ML model inputs with existing human biases.
These may appear in the form of sensitive attributes with associated causal impacts and may also manifest via the under-representation or over-representation of certain groups~\citep{ntfa20}, e.g.~in features relating to gender, age, race, country of birth, etc.
Naturally, ML models may reproduce or amplify such biases and related discrimination~\citep{kage18}.
A more detailed discussion of the issues involved is delayed until Section~\ref{biases_impacts}.
For now, it is essential to highlight that the end of the data engineering phase is a prime opportunity to identify and mitigate biases within samples, labels and generated features.
Somewhat ironically, however, it is still the role of human stakeholders to catch and correct the flaws of other human stakeholders, generally via exploratory data analysis.
It is not even clear whether this will ever truly be mechanised, as recognising biases may require automated context understanding; see Section~\ref{roles_inter_open_env}.
To be fair, there are a growing number of tools that assist in this pursuit, e.g.~Amazon SageMaker Clarify~\citep{hach21} and AI Fairness 360~\citep{bede19}.
Such tools tend to flag issues via various statistical metrics, e.g.~statistical parity difference measures, class imbalance rate and the conditional demographic disparity in class labels.
However, while many existing metrics exist for structured data, bias detection methods are far rarer for multimodal data that includes unstructured formats such as text, images and video.

\subsubsection{Model Development}
\label{workflow_model}

The third phase of the ML workflow covers model development.
Its purpose is to yield a good model or a combination of models, i.e.~an ML solution, which meets the defined requirements for an ML problem.
Of all the phases in an ML workflow, this one has received the most significant attention in automation efforts.
Numerous AutoML packages tackle the challenge of both HPO and model search, with approaches ranging from simple grid/random search to more sophisticated principled strategies, e.g.~Bayesian optimisation.
However, configuration search spaces are typically vast, especially in the case of AutoDL~\citep{doke21}, and the computing resource requirements in both time and memory are immense.
Moreover, optimal outcomes rely not just on the availability of learning algorithms but also on previously specified business goals and the data that is supplied.
It is well known that there is no gold-standard ML solution applicable to all types of problem, with specialised algorithms outperforming general models for many tasks~\citep{stbu21}.

To some extent, research efforts have explored meta-learning~\citep{lebu15} to improve model search, automatically leveraging information acquired in previous runs.
One such example involves narrowing the search space for a current ML task based on the predictors that performed well for a similar problem~\citep{lega10,ngmu21, ngke21}.
However, there is still no comprehensive solution able to transfer the learned knowledge effectively between different tasks~\citep{xiwu21}.
With an exception of the previously mentioned pioneering automated machine learning framework~\citep{kaga09} where meta-learning has been used as one of the fundamental integrated concepts in multiple ML model development, monitoring and maintenance stages, the AutoML community has only included meta-learning approaches in a handful of other systems, mainly academic, e.g.~warm-starting either the initial hyperparameters of optimisation algorithms~\citep{fesp15} or, as part of Auto-sklearn, the learning algorithms themselves~\citep{fekl15}.
For now, humans are still, in many cases, better at learning from previous experiences and generalising knowledge for new tasks.
The importance of facilitating HCI at this phase of an ML workflow is thus evident, with empirical evidence demonstrating that AutoML outcomes can be improved by the inclusion of domain knowledge in the search process~\citep{tako16, wami19}.
In practice, this manifests as technical stakeholders selecting constraints for the search space, including/excluding hyperparameter configurations and model types, possibly pipelined with preprocessors.
Technicians are also expected to select the performance metric with which to optimise this search, e.g.~classification accuracy or mean squared error.

We refer to previous reviews for a detailed discussion about the technical mechanisms involved in ML model production~\citep{kemu20,doke21}; one can infer from these monographs that an AutoML system with good HCI support should optionally allow a human to interface with all the control/feedback flows it lists.
However, few existing AutoML frameworks provide that easy access, acting as black-box systems instead.
This lack of clarity in operation and the resulting lack of trust in outcome are significant factors preventing widespread uptake and use~\citep{lema19, wami19}.
In fact, while the ML transparency debate has often focussed on understanding model outcomes, it has been suggested that it is an equally important requirement for both data and model-constructing processes~\citep{di17}.
Indeed, the integration of visualisations for input data distributions and feature engineering processes has been noted to improve human trust and understanding with respect to AutoML frameworks~\citep{drwe20}.
The same motivation can be extrapolated to model development.
As a result, various mechanisms for transparency have recently been integrated into AutoML frameworks so that technical stakeholders can more easily interact with the underlying algorithms.
This facet of HCI can be categorised into two main research directions: hyperparameter visualisation and pipeline visualisation~\citep{onca21}.

In terms of hyperparameter visualisation, Google Vizier~\citep{goso17} and VisualHyperTuner~\citep{paki19} are two examples that display how the tuning of hyperparameters, within constraints, affects resulting model performance.
An alternative, ATMSeer~\citep{wami19}, which is built on an existing AutoML framework, extends its visualisation to multiple scales.
Users can configure a search space, analyse sampled solutions in real-time, and explore the effects of tuning variables at the level of hyperparameters, hyper-partitions, and models.
However, this is not necessarily enough.
As mentioned earlier, ML solutions are sometimes sought as pipelines of data-transforming components, where the predictor connects to preprocessors and postprocessors; these can include data cleaners, feature engineers and ensemble aggregators.
Such a setup has an analogue in the layered representation of a DL network~\citep{doke21}.
Unsurprisingly, finding ways to depict these ML pipelines and show them to stakeholders in a speedy manner is key to efficient comprehension and ongoing human guidance for ML solution search~\citep{shzg19}, whether for debugging or search-space refinements.
One example of a pipeline visualiser is AutoAIViz~\citep{wewe20}, which utilises `conditional parallel coordinates' to represent sequential pipelines and their hyperparameters in a two-level hierarchical structure.
Specifically, AutoAIViz depicts a chain of ML components at one level and then the hyperparameters for each component at the next.
However, its utility is limited by the capacity to show only three components in sequence, i.e.~two transformers and one classifier.
It cannot portray long pipelines with optionally stacked predictors, let alone complex structures like directed acyclic graphs~\citep{sabu16, sabu17, sabu18}.

\begin{sidewaysfigure}
	\centering
	\includegraphics[width=0.95\textwidth, height=0.52\textwidth]{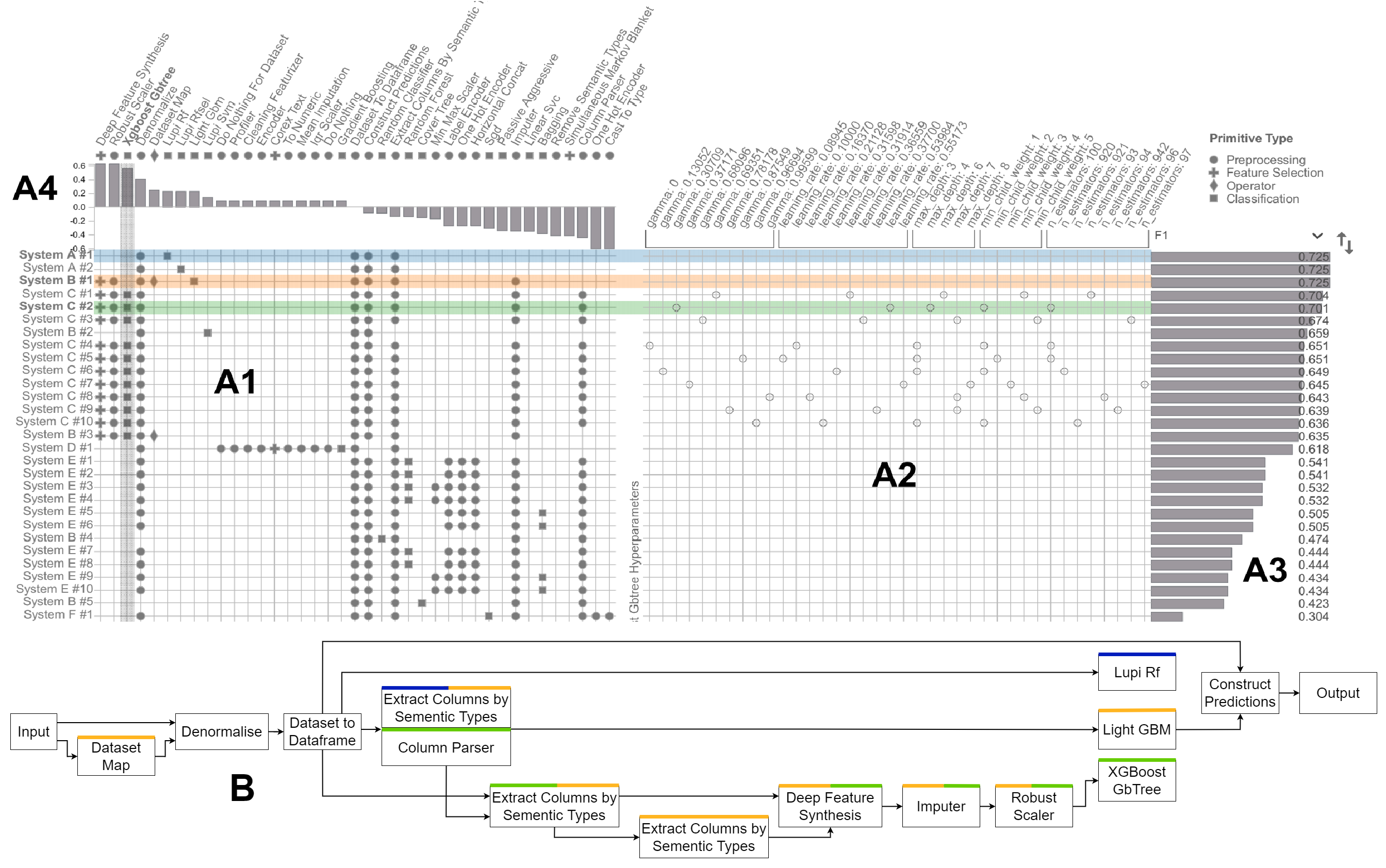}
	\caption{An example of an interactive user interface in \textit{PipelineProfiler} \citep{onca21}. A1: A pipeline matrix showing primitives (columns) belonging to the pipelines (rows). A2: A hyperparameter matrix contains the hyperparameter values corresponding to the selected primitive \textit{Xgboost Gbtree} for different pipelines (rows). A3: Pipeline scores, where users can choose different metrics to rank the pipelines. A4: Primitive contribution panel presents the correlation values between primitives and the pipeline score (in this case, the feature selection method \textit{Deep Feature Synthesis} has the highest correlation with the F1 measure). B: Pipeline comparison panel visually contrasts selected pipelines in terms of structures and the used primitive components.}
	\label{fig_profiler}
\end{sidewaysfigure}

To date, very few visualisation tools exist that can genuinely grapple with sufficiently complex ML pipelines.
One notable anomaly is PipelineProfiler~\citep{onca21}, exemplified in Fig.~\ref{fig_profiler}, which enables the interactive analysis of complex ML pipelines for technical stakeholders.
Such an interface appears to be the way forward for transparent HCI and model development processes in AutoML, supporting perspectives ranging from a high-level overview of pipeline architectures to the low-level depiction of a hyperparameter search space for a selected pipeline.
It is also capable of comparing and assessing multiple ML pipelines across different automated learning systems, although, admittedly, the profiler is only integrated with packages in the DARPA Data-Driven Discovery of Models ecosystem~\citep{lica16}.
Of course, comparisons of ML solutions are fundamental to any process of model selection, whether fully automated or human-assisted, so any tool that captures the contrasts helps facilitate HCI.
Another example that does this is ClaVis~\citep{hemu20}, which trades the ability to deal with complex pipelines for a very comprehensive suite of model-agnostic visualisation capabilities.
Supporting arbitrary classifiers, ClaVis aids users in ranking them by different metrics, provides a scatterplot correlating hyperparameters with model performance, clusters predictors with similar prediction results, generates a history line chart to analyse training behaviours, and constructs confusion matrices for detecting class-specific misclassifications.


It is essential to perform an initial evaluation process with human intervention during model development to assess the quality and performance of models generated automatically by AutoML tools.
In this step, technical stakeholders and business experts assess the models based on performance metrics and comparison to other pipelines. However, they also verify the explainability of learning systems to end users, solution robustness to biases, and the satisfaction of outcomes concerning predefined business criteria.
It has been suggested that learning systems should be assessed on at least seven complementary properties, including performance, robustness, explainability, scalability, model complexity, resource demand, and fairness~\citep{stbu21}.

In addition to performance, explainability is becoming crucial in human-AutoML interactions when interpreting outcomes, thus developing human trust around automated learning systems, enabling the discovery of new insights and supporting informed decisions \citep{ehli21}.
Therefore, one of the roles of humans in the evaluation step is to assess the explanation mechanisms attached to an automated learning system, e.g.~those based on rules, examples involving factuals/counterfactuals, and visualisations.
Certain research~\citep{wani21} has examined rule-based and example-based explanations, assessing their impact on human understanding, their ability to convince, and their effect on task performance when a human makes decisions assisted by learning systems.
Empirical results showed that rule-based explanations have a small positive impact on understanding system outcomes, while explanations based on both rules and examples can make users follow the guidelines more often even when they are incorrect.
However, neither of these types of explanations can help users improve task performance or make correct predictions in new situations.
The authors explained this by a lack of clarification for the rationale of underlying behaviours of the systems.
In contrast to these other types, visualisation is a natural way to overcome this drawback.
One example of a visual analytics framework~\citep{spsc20} for interactive and explainable learning systems is designed to target three user groups: model users, model developers, and business users.
The first two groups, i.e.~model users and model developers, are familiar with using or developing ML models, and so this group needs to understand, diagnose and refine constructed models in a given application context.
In contrast, business users are non-experts in ML and they only care about how to apply learning models for specific domains. 
Therefore, the explanation mechanisms for each group of users with different professional backgrounds require several specific configurations; information flows for each group were discussed in Section~\ref{stakeholder}.

It was previously pointed out that there is a considerable gap between current studies on AutoML systems and requirements regarding ethical factors~\citep{fieh21}.
Therefore, other aspects that need to be evaluated in this phase are the bias and fairness of learned models.
Because learning models rely mainly on data that is collected and created by humans, whatever biases exist in human activities they can also enter the learning models, potentially being propagated and amplified~\citep{ntfa20}.
Therefore, bias factors need to be evaluated and adjusted after building the learning models.
If white-box learning models are used, developers and experts can modify their internal processes.
For example, classification rules can be corrected, probabilities in Bayesian models can be adjusted, or the class label at leaves of decision trees can be changed.
If black-box learning models are constructed, different explanation mechanisms are required to verify and understand the reasoning behind the results.
Their predictions can then be altered if bias factors are detected.
For instance, the proportion of decisions between protected groups and unprotected groups can be adjusted by changing the predictive results of instances close to the decision boundary~\citep{kama18}, and experts can manually check these potential causes of discrimination.
Several existing tools, such as Amazon SageMaker Clarify~\citep{hach21} and AI Fairness 360~\citep{bede19}, also provide metrics to audit the fairness of the trained learning systems.
ML developers and experts can use them to evaluate and correct the learning systems in an attempt to produce fair results.

Finally, in the model development phase, developers and domain experts also need to reevaluate learning systems by the criteria posed in the first stage of the project, both in terms of business and ML.
Technical stakeholders typically expect to check what the models learned by carefully observing the features with a high impact on the outcomes and verifying whether they are reasonable and consistent with business domains.
If any criterion is not satisfied, the model development process should be repeated from the data engineering stage to verify and assess the quality of training data.

\subsubsection{Deployment}
\label{workflow_deployment}

The fourth phase of the ML workflow covers development.
Its purpose is to situate a trained ML model within a computational environment, supplied by sufficient resources, and allow end-users to benefit from its learned knowledge.
This process occurs after the construction and evaluation of an ML model, with technical and business stakeholders deciding whether the obtained model can be deployed for practical use.
A part of this procedure involves technical stakeholders evaluating learned models under production conditions.
First, the operational abilities of the ML solution need to be verified against the available computational resource constraints, working environment, and services.
A potential risk that can arise is that production data is substantially different from training data.
As a result, the actual performance of the learned model has to be incrementally assessed under the operational environment and conditions.
The ML solution may be adjusted according to the deployed hardware and the production environment for each incremental run.
In some essential cases, domain adaptation approaches~\citep{yolo19} can be used to improve the generalisability of the learning models.
While the construction of learning systems is regularly carried out in offline mode, inference and associated decision-making in a realistic environment must often operate online under specific energy budgets and security standards.
In practice, these tasks need to be evaluated by deployment engineers and ML experts.

When the ML model obtained by AutoML tools passes all evaluation criteria on the deployment environment, it may face user rejection, with its usability overestimated during the development process.
Therefore, the ideal practice is to test a prototype with end-users in the deployment phase.
It is crucial to assess whether these stakeholders understand the query responses provided by the trained model.
User guidelines around functions, business rules and domain concepts related to ML models can potentially be generated automatically~\citep{gemo21} to provide end-users with explanations for the functionalities, operational mechanisms and restrictions that drive an ML solution.

With different version control systems for trained models and data, an incremental deployment strategy can minimise undetected errors in the development process.
The system can then be rolled back to the previous version when an error is detected.
In this way, the cost of fixing errors can be minimised.
In practice, various deployment environments inevitably exist in the real world, and each possibly has its own characteristics in terms of latency, energy use, memory, and so on.
AutoML techniques can be used as a mediator for interactions with different layers of a computing system.
For instance, some AutoDL works~\citep{doke21} attempt to automate the search process of hardware configurations for deployment, e.g.~via field-programmable gate arrays~\citep{nako19}.
However, each commercial/industrial entity possibly poses different requirements for the flexibility/rigidity of hardware constraints and ways of provisioning resources.
These constraints need expert knowledge related to hardware, software infrastructure, and computational resources.
They mark the roles of human involvement in the deployment process of ML solutions.

\subsubsection{Monitoring and Maintenance}
\label{workflow_maintenance}

The fifth phase of the ML workflow covers monitoring and maintenance.
Its purpose is to regularly, if not continuously, examine ML models after deployment for signs of performance degradation, which are accompanied by decisions on whether and how to adapt the deteriorating ML solution.
As an example, consider a cybersecurity system that automatically detects and warns of cyber-attacks; the performance of this learning system should be constantly monitored and evaluated by operators.
The system must regularly be fed training data that contains knowledge about new types of viruses and offensive/defensive methods, thus ensuring that the automated system still meets the design standards.
Human roles are critical in such a process, covering verification, supervision, evaluation and maintenance during the operation of the automated systems.
In practice, the main reason for the impairment of learning models over time is the change in data distributions~\citep{gazl14}.
Other reasons include the degradation of hardware and the updating of a system, leading to a shift in the operational environment \citep{stbu21}.

The most effective way to detect the deterioration of a deployed learning model is by continuously monitoring its performance over time.
This process requires continuous evaluation of ML model performance using the same metrics employed during its development process.
All input signals and outcomes of the learning model need to be inspected by end-users and technical stakeholders via log files or visualisations for the monitoring phase.
Specifically, while incoming data can be automatically validated~\citep{scbi19_1}, it should be corrected in real-time with inspection by users.
Statistical information of incoming data should be compared to that of the training data to detect any change in the distribution.
Abnormalities in the learning system can be automatically identified, followed by a notification of the relevant stakeholders~\citep{babr17}.
However, technical stakeholders should ideally have the power to act on or disregard the abnormalities based on their domain knowledge.

Although many different methods can be used to automatically detect changes in data distribution, with types of dynamics detailed elsewhere~\citep{gazl14}, the final decision on when to update the learning system is currently left to domain experts and technical stakeholders based on their knowledge, the operational information provided by the system, the costs to update a model, and the costs of doing nothing, i.e.~leaving model performance to deteriorate.
Ideally, feedback provided by an AutoML system should note significant changes within input data, the number of anomalies, and the degree of any performance deterioration; this knowledge can give technical users the informed authority to trigger a model update.

When updating an ML solution, new data has to be collected to retrain any relevant models so that they can capture an identified change in data distribution.
If a completely new model needs to be developed from scratch, the whole ML development process needs to be repeated, which is time-consuming.
Therefore, most adaptive AutoML research aims to fine-tune models on the new data~\citep{sabu16b,baga17,bafa21}.
In some cases, new learning pipelines will be constructed.
Regardless of the updating approach, deploying the new ML solution without any oversight provided by technical stakeholders and domain experts is risky.
To avoid such risks, the newly updated model should be compared with previous versions to determine whether predefined criteria are still satisfied.
This process should be open to end-user feedback and leverage any domain knowledge provided by relevant stakeholders.

Various automated adaptation strategies discussed in recent studies~\citep{bafa21, ceva21} can be used to automatically adapt and evolve AutoML systems according to a change in data distribution.
However, there are no sophisticated adaptation mechanisms currently employed in AutoML frameworks, although the principles of monitoring/maintenance are starting to bleed in slowly.
DataRobot is one example that introduced a champion/challenger setup in mid-2020, aiming to manage AutoML processes and test predictive models in a production environment~\citep{tom20}.
The champion/challenger approach deploys an original model, called the champion, and several new or retrained models, called challengers, to shadow the champion model.
The challenger models compete against each other for an opportunity to be the new champion.
In practice, the outcomes of this ongoing competition in the production environment are usually reviewed by technical stakeholders, who decide which model should be selected for ongoing usage.
The swap-out could easily be automated, but, presumably for governance-based reasons, they ensure that ``a designated approver will route her final decision to which model will actually become the new standard (i.e., the new champion)''.
In general, humans should monitor and evaluate the results and performance of any updated model because many relevant business factors and domain-specific requirements cannot be gauged from the training data alone.
As a result, performance and fairness metrics for all model versions, new and old, are usually viewed through different visualisation mechanisms, enabling humans to monitor and evaluate the learning system over time.
Of course, continuous learning and continuous adaption with the collaboration of humans may be potential future goals for the development and operation of AutoML systems; see Section~\ref{roles_inter_cmpl_constrained_env}.

\subsection{The User Interface: Many Modalities} \label{multimodality}

Even if AutoML systems acquire a high level of autonomy and adaptive intelligence, which is the focus of Section~\ref{roles_inter_cmpl_constrained_env}, it is difficult to imagine a world where human action is no longer required for any task.
Instead, it is more realistic to see the machinery of AutoML as an assistant rather than a replacement.
After all, at the current time, there is no evidence that computational algorithms can consistently outperform humans when dealing with uncertainty or when coming up with quick solutions to problems that require both cognitive capabilities and implicit knowledge.
Humans and intelligent systems have complementary strengths and weaknesses, so collaboration seems key to optimising ML-based productivity~\citep{empi19}.
Naturally, such collaborative interactions are usually performed via UIs.
Therefore, having discussed the \textit{who} and \textit{what} of HCI in modern AutoML, we now explore \textit{how} these interactions occur in practice.
This section involves examining what basic operations the most complex ML interactions can decompose into, then discussing how these operations are presently/potentially communicated.

\begin{figure}[!ht]
	\centering
	\includegraphics[width=1\textwidth]{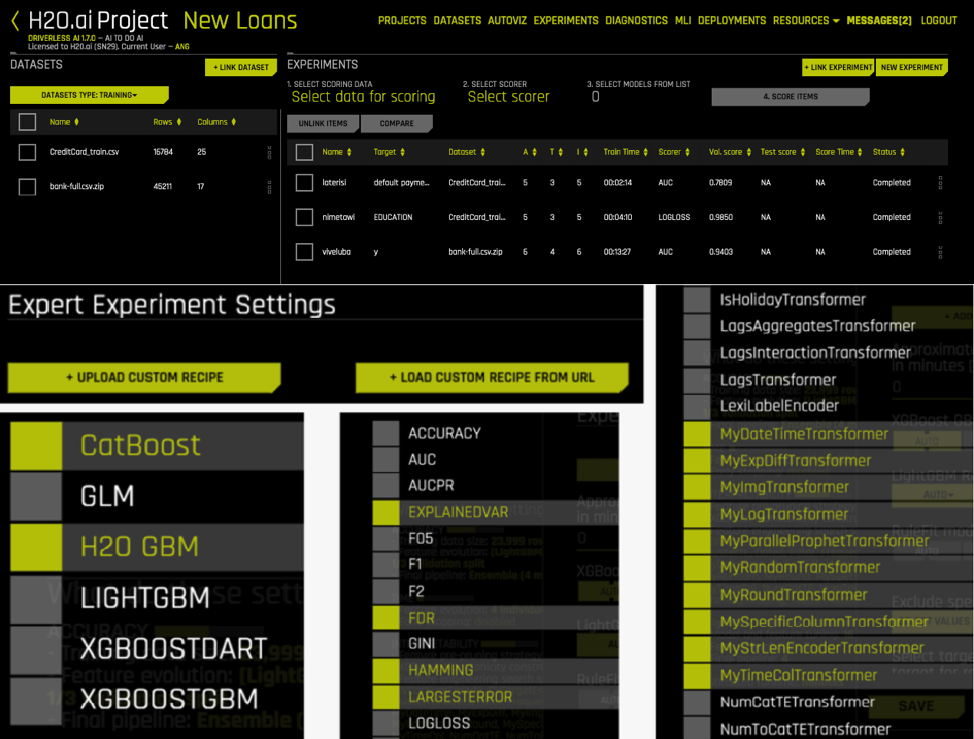}
	\caption{An example of GUIs in an AutoML tool (H2O Driverless AI\protect\footnotemark).}
	\label{example_gui_automl}
\end{figure}
\footnotetext{https://www.h2o.ai/blog/new-innovations-in-driverless-ai/}

To begin with, it is worth noting that many academic implementations of AutoML are programming libraries; stakeholders are often required to code scripts that call these prepackaged automated mechanisms.
However, in industry, most AutoML systems manage the inputs/outputs of HCI via a graphical user interface (GUI).
An example is displayed in Fig.~\ref{example_gui_automl}.
Of course, AutoML frameworks often differ in GUI specifics, yet they all tend to decompose into similar sets of interactive functions.
We examine these here, leaning heavily on previous categorisations related to interactive visual analytics~\citep{yiah07,stga13}.
In short, at the simplest level, stakeholder intent is facilitated by (a) selection, (b) exploration, (c) abstraction/elaboration, (d) reconfiguration, (e) encoding, (f) filtering, and (g) connection.
These techniques were proposed in the context of data exploration and data mining tools, but they remain appropriate for describing HCI in AutoML tools.
We will additionally append the operation of value entry to this systematic listing.

The details of these fundamental operations are as follows:
\begin{enumerate}[label=(\alph*)]
	\item \textit{Selection}: This interaction technique allows users to select visible items of interest via typical GUI elements such as drop-down lists and check-boxes (as shown in Fig.~\ref{example_gui_automl}).
	The intent of \textit{selection} can also be communicated via more sophisticated actions, e.g.~dragging and dropping blocks into boxes, as employed by KNIME~\citep{bece09} and RapidMiner~\citep{hokl16}.
	In this way, stakeholders can parameterise an AutoML system by ordering the use of specific data transformation methods, ML algorithms, hyperparameter ranges, and evaluation metrics.
	\item \textit{Exploration}: This interaction technique allows users to observe and inspect varying subsets of data samples via a GUI.
	The Direct-Walk approach, e.g.~using hyperlinks, is one standard implementation that smoothly navigates a view between informational contexts~\citep{yiah07}.
	For instance, when plots of input data samples are shown, clicking on a specific sample can reveal information relating to that selected sample.
	\item \textit{Abstraction/Elaboration}: This interaction technique allows users to adjust the focus and granularity of data representation on demand.
	Elements of a GUI that support such an operation typically enable navigation between several levels of detail, from an overview to narrow specifics, depending on the needs of a user.
	Such GUI components can take many forms, including tool tip texts that detail specific data items wherever a mouse cursor is moved over, treemaps that facilitate many drill-down operations, and zooming mechanisms to change the scale of representation.
	As an example, ATMSeer~\citep{waya19} presents an overview panel showing the high-level statistics of an AutoML process alongside a separate profiler view; this profiler allows users to traverse down a search history of ML models and hyperparameter configurations at varying granularities of detail.
	\item \textit{Reconfiguration}: This interaction technique provides users with multiple ways of examining data/results through changes in their representation.
	It is equivalent to the adage, ``look at things from a different angle''.
	Often, the \textit{reconfiguration} technique is employed to modify how data items are arranged and aligned, e.g.~sorting columns or rows in tabular views.
	Other examples include using jitter on dense plots to avoid overlap or modifying attributes on scatter plot axes.
	Ideally, a new and clearer perspective allows the extraction of new information from observed data/results.
	\item \textit{Encoding}: This interaction technique provides users with the ability to change the fundamental visual representation of data items, e.g.~colours, sizes, shapes, fonts, and orientations.
	Closely related to \textit{reconfiguration}, the technique of \textit{encoding} goes even deeper into the format of data/results, e.g.~changing from a pie chart to a stacked column chart.
	Another typical example of this operation is using visualisation toolboxes to specify the colour for certain variables.
	\item \textit{Filtering}: This interaction technique allows users to change the set of displayed items via conditions and constraints.
	Specifically, this involves users providing or selecting a condition as input, after which data items are included in or excluded from a view based on whether they satisfy the condition.
	Check-boxes and sliders that allow the selection of conditions/ranges are GUI elements capable of supporting the \textit{filtering} operation and, consequently, dynamic query control~\citep{stga13}.
	\item \textit{Connection}: This interaction technique provides users with the ability to highlight associations or relationships among data items presented within the same view or across different views.
	Such connections can also be automatically found/displayed, e.g.~selecting a data point in one plot and seeing where it sits within another.
	The KNIME system exemplifies this by applying \textit{connection} to data points selected by a brushing method, thus cross-highlighting them within other active plots, e.g.~so stakeholders can see the same sample in both a line and scatter plot~\citep{bece09}.
	\item \textit{Value entry}: This interaction technique is used in most existing AutoML toolboxes to enable the specification of variables, e.g.~hyperparameter ranges or performance-metric thresholds chosen without sliders.
	It may seem like an obvious interaction that all GUIs should provide, especially for the insertion of domain knowledge, but it is not strictly necessary.
	Because all other interaction techniques are sufficient for results analysis, value entry operations can be excluded for systems with default behaviour.
	Of course, if this default behaviour is effective, then the AutoML system can be treated as fully automated; see Section~\ref{roles_inter_cmpl_constrained_env}.
\end{enumerate}

Showing how these various forms of basic operations contribute to interfacing with an AutoML system, Table~\ref{table_modes_of_intr} provides a breakdown of complex interactions across the entirety of an ML workflow.
However, to develop this section, we now consider modalities in which such intent may be communicated with an AutoML system.
Granted, it may initially seem that there is not much to discuss here.
Presently, most AutoML systems are controlled via standard keyboard/mouse user inputs.
Likewise, feedback is typically communicated visually via computer monitors, with warning/completion signals possibly transmitted by audio.
However, ML models produced by the higher-level AutoML systems are often implanted within production/deployment environments where inputs/outputs can take the form of complex representations.
For instance, an ML model query-response cycle could take the following form: gestures to machine code to audio.
This fact becomes important when noting that clean distinctions between an ML model and its encompassing AutoML system are not always possible.
Interfacing with an ML model can be functionally identical to interfacing with an AutoML system, especially those integrating tightly with auxiliary services.
So, forms of interaction with an ML model must be included in any discussion of HCI and AutoML.

\begin{landscape}
	\scriptsize{
		\begin{longtable}{p{.15\textwidth}p{.2\textwidth}p{.3\textwidth}p{.3\textwidth}p{.35\textwidth}}
			\caption{An example of complex AutoML interactions across the ML workflow decomposed into basic operations.} \label{table_modes_of_intr}\\
			\toprule
			Stage      & Complex interaction & Input & Output & Basic operations \\ \midrule
			Problem formulation \& Context understanding & Data exploration visualisation & Raw samples and features & Graphs and plots & Selection, Exploration, Reconfiguration, Encoding, Abstraction/Elaboration, Filtering, Connection \\ \midrule
			Data engineering & Interactive feature exploration & Samples and features & Selected features, graphs and plots & \multirow{2}{*}{\makecell[l]{Selection, Exploration, Reconfiguration, Encoding, \\ Abstraction/Elaboration, \\ Filtering, Connection}} \\ \cline{2-4}
			& Data selection & Samples & Selected samples, graphs and plots & \\ \cline{2-5}
			& Data imputation & Samples, data imputation methods & Imputed samples & Selection (types of imputation), Value entry \\ \cline{2-5}
			& Outlier removal & Samples with outliers, outlier handling methods & Outliers are removed; graphs and plots & Selection (types of outlier handling, types of visualisation), Exploration, Filtering, Connection \\ \cline{2-5}
			& Feature transformation visualisation & Samples and features, feature transformation techniques & Transformed features, graphs and plots & Selection (types of feature transformation, types of visualisation), Exploration, Filtering, Connection, Abstraction/Elaboration, Encoding, Reconfiguration \\ \cline{2-5}
			& Bias checking & Samples and features, bias checking measures & Outcomes regarding the bias measures, graphs and plots & Selection (bias checking measures, types of visualisation), Exploration, Filtering, Encoding \\ \midrule
			Model development & Selection of model types and search spaces & Model types, parameters & Selected models and value ranges of the selected parameters & Selection (models and parameters), value entry (parameters) \\ \cline{2-5}
			& Selection of evaluation metrics & A list of evaluation metrics & Selected evaluation metrics & Selection \\ \cline{2-5}
			& hyperparameter visualisation & Parameters and visualisation types & Graphs and plots & Selection, Exploration, Abstraction/Elaboration, Filtering, Connection, Encoding, Reconfiguration \\ \cline{2-5}
			& Pipeline visualisation & Pipelines and visualisation types & Selected pipelines, graphs and plots & Selection, Abstraction/Elaboration, Exploration \\ \cline{2-5}
			& Pipeline comparison & Pipelines & Graphs and plot for the selected pipelines & Selection, Abstraction/Elaboration, Exploration, Connection \\ \cline{2-5}
			& Performance evaluation & Performance metrics & Selected performance metrics and results, graphs and plots & Selection, Exploration, Filtering, Abstraction/Elaboration \\ \cline{2-5}
			& Explanability verification & Explainable metrics & Selected explainable metrics and results, graphs and plots & Selection, Exploration, Filtering, Abstraction/Elaboration, Encoding \\ \cline{2-5}
			& Bias auditing & Testing samples and features, bias checking measures & Outcomes regarding the bias measures, graphs and plots & Selection (testing datasets, bias checking measures, types of visualisation), Exploration, Filtering, Encoding \\ \midrule
			Deployment & Model evaluation under production conditions & The constructed learning models and production conditions & Results, reports, and plots & Selection (of conditions and measures), Exploration (of results), Abstraction/Elaboration \\ \cline{2-5}
			& Deployment strategies & Deployment methods and environments & The deployed learning systems & Selection, Value entry, Exploration, Abstraction/Elaboration \\ \midrule
			Monitoring and Maintenance & Incoming data monitoring and visualisation & Incoming data and requirements & Operating outcomes, graphs and plots & Selection (measures and data), Exploration, Filtering, Value entry, Abstraction/Elaboration, Encoding, Reconfiguration \\ \cline{2-5}
			& Performance tracking & The learning systems, input data, performance metrics & Results, graphs and plots & Selection (performance metrics, visualisation types), Exploration, Filtering, Abstraction/Elaboration, Encoding \\ \cline{2-5}
			& Bias auditing & Testing samples and features, bias checking measures & Outcomes regarding the bias measures, graphs and plots & Selection (testing datasets, bias checking measures, types of visualisation), Exploration, Filtering, Encoding \\ \cline{2-5}
			& Model updating and retraining & Updating and retaining methods & The updated learning models & Selection, Value entry, Exploration, Filtering
			\\ \bottomrule
		\end{longtable}
	}
\end{landscape}

\begin{figure}[!ht]
	\centering
	\includegraphics[width=1\textwidth] {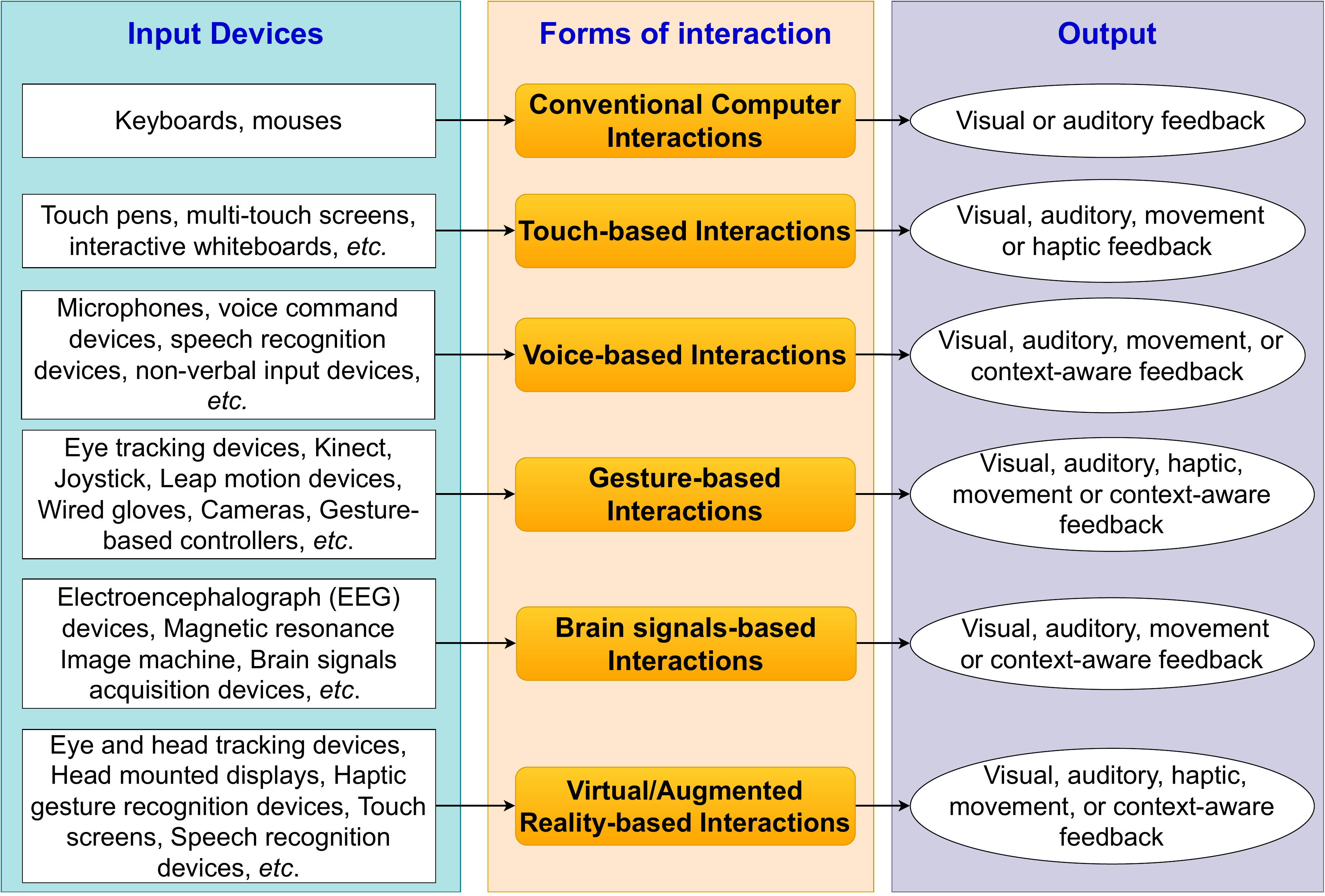}
	\caption{Communication modalities for HCI with ML solutions and AutoML solutions, exemplifying input devices and forms of output feedback.}
	\label{Fig_3}
\end{figure}

Of course, although modern AutoML systems do not use exotic modes of communication for control/feedback, this is not due to any technical barrier, as is the case with the fully automated systems idealised in Section~\ref{roles_inter_cmpl_constrained_env}.
The technology is there; it is merely an issue of what stakeholders currently desire/require.
So, we take a broad scope when discussing the present/potential communication modalities with AutoML systems, summarised in Fig.~\ref{Fig_3}.

Beyond the conventional keyboard/mouse and monitor/audio feedback loop already mentioned, touch-based interactions are perhaps the next most pertinent to AutoML discussions, especially given the popularity of smartphones, tablets and other touch-based accessories.
Already, touch-based interactions have been widely deployed for communication with learning systems in aviation~\citep{meda12, gaca15, rove17}, industrial production~\citep{gosc14}, and autonomous vehicles~\citep{waja17, bede17}.
Likewise, the AutoML community has already begun exploring this domain of communication, with interactive UIs designed for touch screens as part of, for example, the Vizdom~\citep{crga15} and  Northstar~\citep{kr18} packages.
Unsurprisingly, this modality, if implemented well, has several benefits.
For instance, it is easier for touch screen devices to field multiple sources of inputs than traditional computer accessories, thus supporting greater collaboration between technical users and domain experts for both data exploration and progressive model development.

Another modality of interest is voice, given that humans typically speak faster than they type.
Voice-based communications can be less efficient at certain forms of control as they do not inherently provide spatial information, e.g.~like a cursor.
However, if translated appropriately, they can be particularly efficient for value entry and, in well-defined interfaces, selection.
Indeed, voice has previously been used to facilitate HCI between learning systems and users, including those with intellectual disabilities.
Example domains include navigation~\citep{roca17}, robotics~\citep{gusy17}, and self-driving cars~\citep{huxi19}.
Voice has also been considered as a supplementary mode of HCI alongside other modalities~\citep{lele17}.
Of course, voice interactions require learning systems to have an automatic speech recognition module that can transform vocal content into computer-readable outputs.
Ideally, the systems should also have a way to synthesise responses and a mechanism to make those responses intelligent, i.e.~partake in dialogue.
Recently, AutoML methods have been applied to the automated construction of speech recognition modules.
For example, Deepgram AutoML~\citep{st20} is the first framework designed to make training/deploying speech recognition modules for practical applications more efficient.
However, for now, this is as far as the field of AutoML and the topic of voice have crossed over; we are not aware of any substantial AutoML systems with voice-operated control.

Vision-based interfacing is another emergent form of communication between humans and learning systems, seen as both natural and cost-effective~\citep{mali20}.
The use of gestures and eye gazes belongs to this category, having been proved capable of controlling robots and other machines.
It relies on computer vision approaches to detect and understand ocular motion or the configuration of human hands.
This communication modality has two requirements: hardware devices to capture visual structure/motion and learning algorithms to interpret commands inherent in these visions.
Within this category, gesture-based interaction is already widely used for many practical applications of learning systems.
For example, real-time control of robotic arms with hand gestures has already been shown possible~\citep{rash10}, and, elsewhere, an interactive semi-autonomous control interface has allowed human-robot collaboration to facilitate various coarse motions in robotic arms~\citep{qufo14}.
Similarly, HCI based on ocular activity is also used in different practical sectors, e.g.~for monitoring and control in the field of teleoperation when the hands of a user become busy~\citep{mali20}.
Relevant domains include mobile robots~\citep{cage18} and the control of smart wheelchairs~\citep{wavi16}.
While AutoML does not appear to have leveraged vision-based interfacing yet, such a modality seems to be a natural fit with certain operations, e.g.~tracking gaze during data/results analysis to abstract/elaborate/explore.

In recent years, interfaces based on brain signals have become potential methods of control/interaction with vehicles, robots, and ML models~\citep{mali20}.
In this type of interaction, it is essential to acquire electroencephalogram (EEG) signals through specialised devices and analyse them.
In most cases, an ML module will be used to translate these electrical signals in a human brain into intent or cognitive stages, after which these mental commands are used to control and communicate with applications~\citep{lido21}.
Robotic teleoperation is one domain where neural control seems promising, but many challenges are still to be overcome.
Decoding neural activities and mapping these signals to motion commands join a list of complications, including latency, noisy/erratic low-dimensional user commands, and asymmetric control signal inputs~\citep{muve17}.
In practice, it seems the most promising approach is to distinguish between levels of operation, using human-in-the-loop supervision to communicate high-level intent and facilitate task planning, then use robotic systems to automate lower-level processes with reduced reliance on user control~\citep{crgo02}.
Given that neural control is still in its infancy, it is difficult to predict how it would factor into an AutoML system, e.g.~whether it would allow the communication of finely resolved intent or whether it would be better reserved for directing a maximally autonomous system; see Section~\ref{roles_inter_open_env}.

Augmented Reality (AR) is the final modality we list, argued to be appealing for real-time HCI with automated learning systems due to its enhanced UX~\citep{bach20}.
An intuitive interface based on AR can strengthen the interactions between users and a working environment, in which virtual components are inserted into human perception of the real world.
Such a UI enables users to leverage their senses and engage more closely with information residing in the working environment.
In effect, the productivity/quality of human decision-making can be improved.
Thus, AR has been used for many applications in the industry of smart manufacturing~\citep{bach20}, with interfaces usually employed for the quick maintenance of manufacturing systems.
Integrated with continuous monitoring mechanisms, an AR interface can provide maintenance technicians with on-site context awareness and safe operational procedures~\citep{tate17}, as well as real-time instructions from remote experts~\citep{faxu20}.
Interfaces based on AR are also appealing for their instructional capability, explored previously via on-site technical training programs that were designed to teach trainees real-environment skills quickly~\citep{meme15}.
The AR modality has also enabled technical users to train machine operations~\citep{caqi20} via real-time role-playing interactions that convey intent to robots without the need for offline training~\citep{cawa19}.
Many other practical applications related to collaboration between humans and automated learning systems can be found elsewhere~\citep{bach20}.
In addition to AR, Virtual Reality (VR) can likewise be used to build interactive training applications for end-users in many domains, e.g.~games, medical applications, rehabilitation, and industrial/commercial training.
Methods that automatically construct, adapt and assess training content in VR environments, based on learning models and expert knowledge, are reviewed elsewhere~\citep{vaga16}.
Unsurprisingly, given the abstract nature of many ML operations, AR/VR has not yet been substantially explored in AutoML.
However, it is not hard to imagine `tangible' AR/VR objects eventually representing system objects/mechanisms, e.g.~virtual blocks for the components in an ML pipeline.

Finally, to cap off this section, we briefly look beyond the current state of AutoML and communication modalities that, in some form, are within easy reach.
First, we note that most of the above discussion assumes humans dictate the information communicated to an AutoML system.
Likewise, the assumption is that they can operate with all the feedback they receive.
However, as AutoML systems are expected to operate more autonomously in managing their ML tasks, they will likely become much more complex.
Consequently, beneficial HCI does not just rely on supporting many convenient communication modalities.
AutoML UIs should also be context-aware, providing users with \textit{only} the information and interactivity that are essential for their current situation~\citep{gosc14}.
Context-awareness requires learning systems to be equipped with various sensors, aggregating raw data into higher-level context information that can be used to understand and adapt UI inputs and outputs, respectively, in real-time.

Additionally, with AutoML systems projected to evolve from relatively simple tools into more multi-faceted entities, there is no reason to imagine that all the previously listed modalities cannot be supported.
Consider an autonomous car as an analogous example.
Drivers can interact with the car using touch screens.
Ocular tracking systems can be used to analyse distracted behaviours and promote safety.
Gesture-based control is usually deployed to interact with vehicular navigation systems, as well as video, audio, and air conditioning systems~\citep{huxi19}.
Voice-based interactions can, via inputs, be used to control entertainment devices or detect emotional responses to the driving environment, and, via outputs, voice can convey instructional information back to the driver.
Moreover, it is known that understanding the cognitive/physiological state of a driver is crucial to building effective driver assistance systems and autonomous vehicles.
Hence, interactions based on brain signals have also been proposed to detect, analyse and monitor driver state~\citep{zhes21}. Additionally, AR-based interactions can be used to increase driver awareness by highlighting elements present in a natural environment, e.g.~road signs, crosswalks or pedestrians~\citep{bogi20}. Perhaps, similarly, an AutonoML system may eventually be much more closely integrated with daily human activities across a full spectrum of communication modalities, possibly being called upon, at will, to apply ML techniques to practical problems in the environment of a user.
Such speculation is left to Section~\ref{roles_inter_open_env}.

\subsection{Improving the Outcomes of Interactions}\label{improve_outcome}

Having discussed the \textit{who}, \textit{what} and \textit{how} of HCI in AutoML, we now delve into the \textit{why}.
For many years, particularly in academia, the development of AutoML theory/technology has been driven by the goal of making machines take over ML operations from humans.
So, why has there been such a recent shift towards facilitating better HCI between users and AutoML systems?
There are several reasons, but they all acknowledge that, however advanced and autonomous AutoML machinery becomes, ML is ultimately performed for humans.
If a human is not satisfied by the outcomes of their interactions with an AutoML system, the system will simply fall out of use.
So, this section looks more closely at the big issues bringing attention to HCI in AutoML as of the early 2020s.

Simply put, beyond matters of practical usability, the greatest obstacle for AutoML uptake is trust; see Section~\ref{outcome_trust}.
It is clear that the explainability of AutoML systems is vital to confident stakeholder engagement, discussed in Section~\ref{outcome_explain}, and we examine current approaches towards boosting this quality in Section~\ref{explanation_type}.
Of course, even with a better understanding of AutoML operations, humans still bring cognitive biases to their interactions with ML-processing computers, and we detail these in Section~\ref{biases_impacts}.
Indeed, fairness in the outcomes of ML applications is likewise crucial to fostering trust, elaborated in Section~\ref{outcome_fair}, and we examine current approaches towards mitigating biases in Section~\ref{mitigate_bias_collab}.

\subsubsection{Towards Trustworthy AutoML}
\label{outcome_trust}

Apart from matters of utility and performance~\citep{scke21}, AutoML systems and the ML models they produce are only used for practical applications -- without controversy -- if they are explainable and fair.
Fundamentally, mechanised algorithms need to be trustworthy, and trust has previously been considered an adhesive quality for interactions between humans and ML systems~\citep{shin20}.
It is argued that the purpose of building trust between humans and intelligent systems is to allow stakeholders to foresee the behaviours of learning systems in the presence of risk~\citep{jama21}.
Here, we review what affects human perception/trust of automated systems from psychological and societal perspectives.

Recent survey results indicate that stakeholders continue to hold a negative view of decisions generated automatically by ML algorithms because of their potential risks~\citep{arhe20}.
Experiments show that this negativity is minimised when users understand the characteristics and operations of algorithms, suggesting that transparency is a crucial factor in making AI systems more ubiquitous~\citep{shin20}.
Another recent study similarly indicates that the accommodation of explanatory means within learning systems boosts trust~\citep{shin21}, allowing users to, via transparency and accountability, comprehend underlying decision-making processes.
In fact, contrary to assumptions that practitioners may not want to involve themselves in the details of an ML algorithm, research suggests that users usually do wish to understand how algorithms work, at least to assess whether the outcomes are fair and reliable~\citep{wopo18}.
Ultimately, if one reviews the literature, it quickly becomes apparent that trust in ML is closely linked to stakeholder satisfaction in several interrelated concepts: fairness, accountability, transparency, and explainability (FATE)~\citep{shin21}.
The higher the satisfaction, the greater the trust.

Admittedly, the FATE acronym is not necessarily exhaustive.
For instance, people frequently assess explanations based on prior knowledge and beliefs, not just their comprehension of learning algorithms.
Thus, the concept of causability has been proposed to measure the quality and effectiveness of explanations to users; the associated metric assesses whether stakeholders attain a strong causal understanding of results within a specified context~\citep{hola19}.
Other experiments~\citep{shin21} make a stronger argument, stating that, apart from simply being a metric for explanation quality, causability is an antecedent of explainability and forms an essential element in trust.
The data science community will likely continue to debate the best way to systematise these concepts.
However, it appears clear that facilitating FATE and causability in AutoML systems would strengthen the quality of collaborations between humans and user-centred learning systems.

In considering the design of trustworthy systems, some have discussed leaning on the notion of interpersonal trust in sociology~\citep{jama21}, as divided into two types: intrinsic and extrinsic.
It is suggested that intrinsic trust appears only when users can successfully understand the actual reasoning process of learning systems, one that is consistent with either the logic of human reasoning or the priors held by a user.
In contrast, extrinsic trust can be achieved through the persuasiveness of model outcomes and other behaviours; without mechanical transparency, a trustworthiness evaluation scheme needs to rely on this.
Naturally, one criterion for building extrinsic trust is ensuring that the distribution of samples used in evaluating the quality of a trained model matches the distribution of unseen data when the models are deployed in practice~\citep{jama21}.
In simple terms, data used to test an ML model should represent the real world.

A prerequisite for extrinsic trust, where the logic of an AutoML system is hidden, is plausibility.
In general, humans need to believe that model outcomes do not outright contradict their understanding of an ML problem and an associated data environment.
At the very least, seeming contradictions should be quickly resolved with the consideration of accessible insight.
Otherwise, implausibility can dramatically weaken understandability~\citep{fukl20}.
That said, plausibility is a highly subjective aspect based on personal assessments of model utility and explanations.
A seeming contradiction is not a sign of falsity; it only indicates that the model is currently inconsistent with a stakeholder pool of knowledge/understanding.
In truth, as a crowdsourcing study shows~\citep{fukl20}, the plausibility of explanations is greatly affected by human cognitive biases and fallacies; see Section~\ref{biases_impacts}.
Thus, to distinguish whether the error of plausibility arises from human or machine, debiasing mechanisms are necessary to integrate within a trustworthy AutoML system; see Section~\ref{mitigate_bias_collab}.

To conclude, trustworthy AutoML systems will ideally contain mechanisms to facilitate fairness, accountability, transparency, explainability, and causability.
Intrinsic trust can be fostered by ensuring users understand internal operations or associate them with human reasoning, while extrinsic trust can be attained by seeking consistent forms of outcome evaluation and promoting plausibility.
Many of these aspects are closely bundled with explainability, so we proceed to dive into this topic in upcoming subsections.
However, by and large, modern AutoML systems can easily fall victim to human cognitive biases, e.g.~via the input of training data or other interactions, which can adversely affect performance.
Therefore, after the explainability discussion, the subsequent subsections focus on human bias factors, fairness, and design principles for HCI interfaces with unbiased protocols.

\subsubsection{The Importance of Explainability}
\label{outcome_explain}

A recurring criticism that modern ML algorithms have faced, including state-of-the-art DL, is that they are hard to trust because one can never be sure when they will incur nonsensical or hazardous errors~\citep{mada19}.
The decision-making process in these algorithms is typically obscured, meaning that algorithm designers themselves cannot explain their outputs.
Similarly, most modern AutoML tools are also black-box systems that do not provide much human-understandable detail on how they select ML pipelines and associated hyperparameters~\citep{xiwu21}.
This difficulty in interpreting and assessing the inner processes of an AutoML system makes it hard to trust the resulting ML solutions that they produce~\citep{wami19}.
Indeed, humans do not usually adopt systems that are not explainable, accountable, tractable and trustworthy~\citep{zhli18}, especially if the reasons leading to outcomes are unverifiable.
It follows then that developing explainable/interpretable learning algorithms is key to promoting confidence in their usage~\citep{jama21}.
Some recent evidence supports this by recognising that engineering transparency characteristics into AutoML systems would significantly increase stakeholder understanding and, correspondingly, trust~\citep{drwe20}.
Of course, the topic of explainability is very much a part of HCI, as explanation is an interaction between a learning system and a human~\citep{liaj21}.

The detailed benefits of integrating explainability mechanisms into AutoML systems are many.
For instance, interpretability is a prerequisite to improving the fairness of learning models, it aids the anticipation of system behaviour under challenging circumstances, and it facilitates ease of HCI~\citep{ardi20}.
Indeed, while we focus on trust in this section, the importance of explainability to practical operations should not be discounted.
With a better understanding of how black-box systems generate results, it becomes easier for stakeholders to diagnose errors and refine the operation of the systems.
When errors are understood to occur, particularly early, users can take appropriate actions such as reporting failures, updating systems, adding more training data, and rerunning an ML application with reconfigured settings.

Naturally, to construct learning models that are effective and explainable, many studies lean on insights drawn from the social sciences to understand explanation as it relates to human psychology~\citep{mi19}.
Seemingly, a good explanation does not only show why an ML model makes a prediction/prescription, X, but also explains why the model selects X over an alternative, Y.
Therefore, the use of counterfactuals in learning systems can play a critical role in achieving this objective because they enable people to identify and infer cause-effect or reason-action relationships between events, thus justifying past results and predicting future outcomes~\citep{by19}.
Rule sets can be used to build such counterfactual explanation summaries~\citep{rala20}, providing ML-leveraging decision-makers with a high-level global comprehension of underlying models.

Ultimately, given how critical explainability is to AutoML trust and uptake, the explanation process must be robust.
Explainable systems must forge human-interpretable reasoning for the decisions they make, e.g.~in terms of ML pipeline selection or input-specific predictions/prescriptions, via the extraction/collation of knowledge from raw data that is compact, precise, and complete.
This knowledge must be transferred to stakeholders via easily understandable means, e.g.~in the form of rule sets and visualisations.
A deeper dive into modern-day approaches for boosting explainability is presented in Section~\ref{explanation_type}.
However, while transparency and clarity are prerequisites for trust, they do not inherently induce it.
Certainly, user-friendly explanations can support the plausibility of model outcomes, expose any unfairness inherent in an ML solution, and enhance performance/productivity via strengthened user engagement.
Nonetheless, explainability on its own is simply a metric for information quality.
Belief in that information relies on human cognitive biases and how they are managed via HCI processes; these discussions begin in Section~\ref{biases_impacts}.

\subsubsection{The Methodology of Explainability}\label{explanation_type}

Having motivated the importance of explainability to user engagement with AutoML, we now review typical approaches that are either in use or proposed to support this characteristic.
In the literature, two main criteria are usually used to classify explanation methods for ML systems: scope and stage~\citep{vilo20}.
In terms of scope, explanation methods can be divided into global and local types.
Global explanation methods aim to understand the entire inferential process of a learning system when producing predictive results, examining conditional relationships between model output and predictive features within a complete dataset.
For local explanation, the purpose is to explain why the learning system generates a specific outcome for a single sample.
In contrast, the notion of stage examines when the techniques are applied, i.e.~ante-hoc or post-hoc~\citep{vilo20}.
Ante-hoc techniques aim to consider and enforce explainability from the beginning of model development, while post-hoc methods leave the training process of ML models alone; understanding the behaviours of a model are left to external explainers incorporated at testing time.
A deeper review of ML explanation methods is out of scope for this monograph, and interested readers are referred to recent surveys on the topic~\citep{mo20, ardi20, vilo20, lipa21}
Here, we focus mainly on the formats that humans require of explanations produced by automated learning systems to explain how and why specific predictions/prescriptions are made.
These output formats include rules, textual explanations based on examples/rationales, and visual explanations.

It has been argued that explainability should be studied as it arises within a purely human context, i.e.~a shared meaning-making process involving both explainers and explainees~\citep{up21}.
In such interactions, people usually use why-questions to obtain reasons for decisions.
Often, these are asked in comparative fashion: ``Why was decision X made rather than decision Y?''
This contrast enables explanations to be expressed as differences between an existing phenomenon (fact) and a hypothetical (foil)~\citep{li90}.
Moreover, a contrastive explanation tends to make communicating information in regular life both concrete and concise.
A similar approach can be applied to AutoML, distilling an abundance of information into compact discrimination between two or more choices that an ML model faces.
This contrastive technique can be used in rule-based explanations, example-based textual explanations, and visual explanations.
For instance, consider a bank customer that is 35 years old, has black hair, earns \$4500 per month and has two children.
It is one thing to state that the bank rejects home loan applications ``IF \textit{customer age} < 30 AND \textit{monthly income} < \$5000''.
It is another thing, arguably more informative, to explain that the particular customer would \textit{not} be rejected if their monthly income was at least \$5000 greater. The use of contrastive explanations cuts through obfuscating detail and helps stakeholders identify the differentiating boundary between fact and foil, i.e.~which specific conditions would make the outcomes change.

Now, as is becoming a trend within this monograph, we note that visualisation is a natural and attractive way to communicate the outcomes of an AutoML system and generated ML models to a human.
This assertion applies to explanatory information as well.
Many graphical tools exist, attempting to unveil the internal operating mechanisms of an AutoML framework.
These include Google Vizier~\citep{goso17}, VisualHyperTuner~\citep{paki19} and ATMSeer~\citep{wami19}, which deploy graphs and scatter plots to illustrate searchable configuration spaces for HPO.
Similarly, as depicted earlier in Fig.~\ref{fig_profiler}, PipelineProfiler portrays and compares ML pipeline structures with their parameters~\citep{onca21}.
Depending on the problem context, features essential to model predictions can also be highlighted via various graphical means.
Heat maps with different colours can be used, in the field of computer vision, to emphasise special regions of an image~\citep{risi16}.
Likewise, in the field of natural language processing (NLP), they can bring attention to specific words in a paragraph~\citep{stge17}. 

Another intuitive form in which to present explanations to humans is the use of textual descriptions, which can be conveyed in written or spoken format~\citep{sebe18}.
They can, of course, be bundled with supporting evidence and contrastive situations to increase plausibility.
For instance, consider an explanation for a predictive ML model that rates upcoming films: ``Movie A directed by XYZ is likely to score a good rating because it is an action film. Three out of four action films directed by XYZ were previously rated good, i.e.~C, D, E but not F. Movie B directed by XYZ is likely to score a bad rating because it is a drama film. One out of three drama films directed by XYZ were previously rated good, i.e.~G but not H or I.''
Evidently, the use of examples aids in describing past phenomena and generalising knowledge to new scenarios~\citep{nesi72, peza97}.
More fundamentally, though, textual explanations in natural language are appealing for HCI due to a couple of simple reasons~\citep{ehha18}.
Firstly, they are intuitive to laypeople, who cannot necessarily grapple with the inner structure of learning models and data characteristics in technical terms.
The use of contextually appropriate textual explanations thus significantly boosts communication effectiveness with non-experts.
Secondly, humans can often digest the information contained within natural language quickly, allowing for collaborative interactions between learning system and user occurring in near real-time.
Both these reasons essentially boil down to the fact that the use of natural language is inherently comfortable and expressive for humans.
Of course, textual explanations can always be supplemented with visualisation techniques, should deeper and more technical analyses be required~\citep{sebe18}.
For now, most modern AutoML systems do not have sophisticated explanatory mechanisms that use natural language, but research is undoubtedly ongoing in the broader data science community.
There have been efforts, for instance, to train a computational model that can automatically transform the internal states and action data of an autonomous system into human-centred textual explanations~\citep{ehta19}.

Natural language, however, often uses many filler words that do not allow for information to be contained compactly; rule-based explanations are an alternative.
Using rules to explain data-driven learning models allows for concise structure and logic, yet humans still find them relatively easy to understand~\citep{lipa21}.
They can provide both global and local explanations for the predictive outcomes of an ML model~\citep{fu05}, i.e.~in the form of a rule set for a query-response space and an individual rule for a specific example, respectively.
Rules are typically constructed in the form of `IF $\ldots$ THEN $\ldots$' statements together with AND/OR operators to represent value combinations of input features and predictive outcomes.
For instance, in line with a previous example, a home-loan application rule can be expressed as follows: ``IF \textit{Age} > 60 AND \textit{monthly income} < \$6000 THEN Output = Rejected''.
The parts before and after `THEN' are called antecedent and consequent, respectively.
The antecedent usually comprises one or more premises combined with logical operators.
However, rules do not need to be crisp, i.e.~adhere to classical logic, where the combination of premises leads to an entirely true or false consequent.
Explanations can be expressed as fuzzy rules, where relations between antecedent and consequent are true to a non-binary degree~\citep{gu01}.
Indeed, the integration of fuzzy rules into learning algorithms can aid with reasoning/explaining the internal logic of black-box ML models such as neural networks~\citep{pane01} and support vector machines~\citep{nuan02}.
Fuzzy-based sets of premises and conclusions can also be used to build arguments, which are essential components in a defeasible reasoning process.
Defeasible reasoning aims to build a common-sense qualitative understanding in highly uncertain contexts, where the availability of new information can lead to the fallibility of preconditions and the withdrawal of previously generated conclusions.
Given that ML models are usually imperfect approximators of a desirable function, the integration of defeasible reasoning into data-driven learning systems is appealing, potentially improving capabilities for knowledge representation/extraction, grappling with incomplete/uncertain information, and communicating the internal processes of systems/models~\citep{lo16}.

\begin{figure}[!ht]
	\centering
	\includegraphics[width=\textwidth]{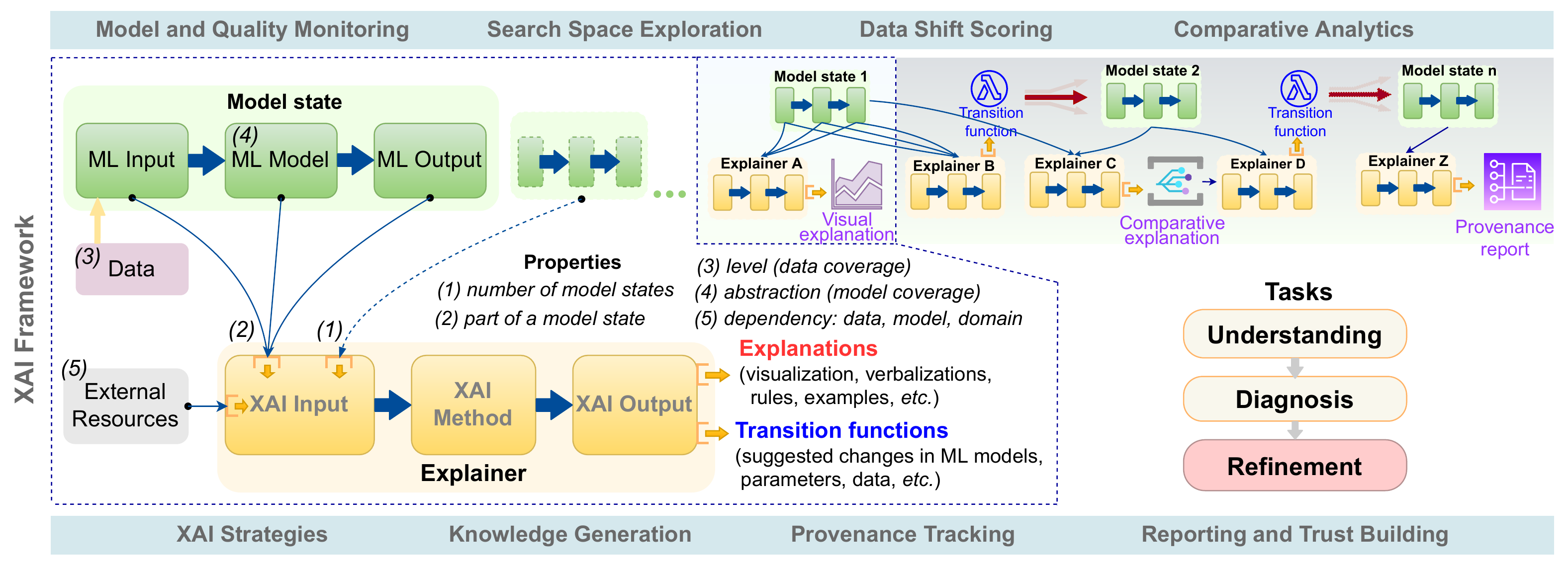}
	\caption{A schematic of an iterative explainable ML framework aiming to understand, diagnose, and refine ML models~\citep{spsc20}. Each `explainer' object, possessing five properties, takes in one or more model states and data characteristics as inputs, then deploys explanatory approaches to produce an explanation or transition function as outputs. The framework also includes global monitoring/steering mechanisms, listed in the long cyan blocks, to manage the explainers.}
	\label{fig_xai_framework}
\end{figure}

In practical applications, different types of explanation can be integrated into automated learning systems using a single unified framework.
One such example, illustrated in Fig.~\ref{fig_xai_framework}, is ExplAIner, which employs both visual and textual explanations for outcomes generated by learning models~\citep{spsc20}.
These assist in continuously interpreting, diagnosing and refining ML models during their operation, serving as ideal supporting mechanisms for technical stakeholders.
The resulting explanations also aid other stakeholder groups in understanding why an ML model made particular predictions/prescriptions.
In detail, the ExplAIner framework incorporates so-called `explainer' objects that transform model configurations and input data into one of two outputs: one, explanations in different forms of visualisation/verbalisation, or two, transition functions that suggest new configurations for model refinement.
Eight global monitoring/management mechanisms also support the framework: search space exploration, model quality inspection, the comparative analysis of solutions, reporting processes for trust-building, a module for knowledge generation, etc.
Naturally, given the growing importance of HCI and data-driven learning systems, ExplAIner is but a single example, with many modern AutoML frameworks beginning to integrate various explanation approaches for ML solutions.
For instance, Google Cloud AutoML Tables~\citep{go21} provides users with feature importance explanations to visualise which features contribute most to the training process of models and individual predictions.
Similarly, the Mimic Explainer package integrated with the Azure AutoML suite~\citep{az21} likewise supports feature-importance explanations.

\subsubsection{Human Cognitive Biases and Their Impacts}\label{biases_impacts}

Explanation mechanisms can boost human comprehension/reasoning related to AutoML operations.
Unfortunately, it is argued that, while these mental processes are typically rational and heuristic, they are subject to distortion via cognitive biases~\citep{waya19}.
This is not a new concept; the term `cognitive bias' was promulgated in the 1970s to describe flawed patterns of thinking in making judgements and decisions~\citep{tvka74}.
Specifically, human cognitive biases can be defined as unconscious errors in cognition resulting to issues with memory, attention and other mental facets that can impair the rationality/accuracy of judgements and decisions~\citep{ru21}.
They are considered the results of a brain attempting to simplify a complex world to help make decisions quicker.
Confirmation bias, anchoring bias, availability heuristics, the misinformation effect and self-serving bias are all common examples of cognitive bias.
A more extensive list can be found elsewhere~\citep{bl16}.

As humans play a critical role in operating AutoML systems, influencing processes ranging from training-data preparation to the constraining of search spaces, cognitive biases can be propagated through to the resulting ML models.
Technically, the building of AutoML systems can also embed biases within mechanisms, but these can be harder to `correct' if they are non-configurable.
The critical point here is that human biases can potentially impact ML application outcomes, so it is essential to identify and manage them.
Bias mitigation can help boost the reliability of an AutoML system and, accordingly, increase human trust in the machinery.
Of course, we note that the results of an ML model can often seem more plausible if they \textit{conform} to existing biases, so this is not a trivial topic.
Somewhere along the line, a connection needs to be hypothesised between a possible bias and a negative outcome, either potential or realised, e.g.~unfairness; see Section~\ref{outcome_fair}.
In any case, this subsection simply discusses common biases that exist in ML algorithms, as well as human cognitive biases that can adversely affect a human understanding of AutoML outcomes.

Importantly, the ecosystem of HCI is complex, and human biases can distort ML outcomes at various phases of information flow.
Some biases may be introduced at the very beginning by system/algorithm designers, while others may take effect at the very end, i.e.~when users interpret model outputs.
A recent keynote talk~\citep{ge19} attempts to categorise biases associated with learning systems into three groups: data bias, automation bias, and algorithmic discrimination.
Data bias refers to how training data is selected.
If the input data provided to the learning system is biased, then the output is likely to be biased.
For instance, if an automated recruitment system is trained on candidate resumes for which males dominate the elite level of a career path for historical reasons, an ML model may incorrectly learn that male employees perform intrinsically better than female employees.
This gender bias is just one context-specific example of problems arising with class-imbalanced data.
Of course, in the workforce example, no bias has necessarily been introduced in the data-preparation phase; it is simply that the real-world environment does not represent the equitable function one wishes an ML model to learn.
However, data bias is possible even if a real-world distribution is otherwise unbiased, simply because humans can curate raw data with uneven sampling.
Whatever the case, the bias in this data can propagate down an ML workflow and integrate into its outcomes~\citep{har20}.
Other forms of data bias include data heterogeneity, where data is pooled from various sources with different formats, e.g.~text, images, audios, and videos.
Many ML algorithms often work with one data source that is assumed to be independent and identically distributed (\textit{i.i.d.}).
Data heterogeneity usually violates this assumption~\citep{lich17}, weakening the validity of an algorithm.
Yet another form of data bias links to the issue that high-dimensional data often results in overfitting, with ML algorithms subject to the curse of dimensionality~\citep{do12}.
This issue is usually tackled by reducing dimensionality via feature engineering/selection, but this procedure introduces biases of its own.

Next up is automation bias, which can be considered a fallacious `appeal to authority', the authority here being an AutoML system.
Specifically, humans tend to accept suggestions produced by automated decision-making systems, even in the presence of contradictory information~\citep{ge19}.
Unsurprisingly, an over-reliance on automation leaves users exposed to the many potential biases inherent in such systems~\citep{arhe20, habo16, za16}.
The concept of `death by GPS', i.e.~accidents caused by uncritical acceptance of driving commands, is one dramatic example of automation bias.
Admittedly, it can seem antithetical to promote trust by removing a bias already correlated with trust, but misplaced faith is dangerous, often leading to extreme scepticism in a technology when that faith is broken.
It is also not contradictory to claim that AutoML faces trust-based obstacles in uptake while also stating that automation bias is widespread; most contemplation around engaging in a service occurs, for better or worse, before that engagement begins.
Regardless, the risks of automation bias are clear, especially in high-stakes applications, and ideal AutoML systems must either remind users to remain critical or otherwise facilitate both explainability and accountability.

Finally, algorithm bias refers to systematic errors within high-level AutoML mechanisms and low-level ML algorithms.
Such errors are often an unavoidable part of generalising patterns and approximating functions.
The issue is that any systematic errors in an ML model, including an inability to capture nuance within a problem context, can lead to adverse outcomes, e.g.~discrimination via the preferential allocation of privileges to those with wealth~\citep{chva21}.
As a side note, definitional boundaries are fuzzy in the broader literature, so algorithm bias often overlaps with data bias, sometimes referring to the replication of unethical societal biases hidden within training data.
There are many examples of such an occurrence in real-world applications.
For instance, previous research found evidence for racial bias in the US health care system, with favouritism shown to white patients over sicker black patients~\citep{obpo19}.
It was discovered that an associated algorithm incorrectly correlated high health care costs with the sickest patients, prioritising them for intensive-care programs.
On the surface, this conclusion does not seem unreasonable, i.e.~using health costs as an indicator of health needs.
However, because of systemic societal racism, black patients are less likely to receive the health care they need, thus spending less money than white patients.
Effectively, the model did not consider the nuance of health care availability.
In such a way, ML algorithms can learn, amplify and even legitimise institutional biases~\citep{ge19}.
Granted, given that algorithmic biases are systematic and repeatable, one of the best countermeasures is to know what assumptions underlie a procedure before employing it.
However, this remains a significant challenge for modern AutoML, which simultaneously opens up ML technology to non-experts while, for the most part, obscuring its internal processes.

Thus far, except for automation bias, we have mainly focussed on errors distorting model outcomes.
However, the final step of an ML application is often conveying predictions/prescriptions to a stakeholder, and cognitive biases can certainly affect understanding these.
Therefore, ideal AutoML systems are expected to optimise the communication of model results, not just the model itself.
This assertion is especially true as, nowadays, learning algorithms are increasingly seen as important socio-technical artefacts~\citep{ki17} embedded in many societal/organisational structures, associated with a diversity of relationships, incentives, and roles in society~\citep{arhe20}.
Accordingly, a recent study argues that explainable AI must be designed to express helpful explanations while avoiding misunderstandings due to typical biases faced by human reasoning, e.g.~representativeness bias, availability bias, anchoring bias, and confirmation bias~\citep{waya19}.
Another study similarly discusses the potential impacts of cognitive biases -- it highlights twenty popular types -- on the way humans interpret learning models, particularly inductively learned rule-based systems~\citep{klba18}.
Ultimately, careful design of how rules and other forms of explanation are represented seems to be the best path for avoiding human cognitive biases during AutoML-specific HCI.

\subsubsection{Fairness}
\label{outcome_fair}

Biases have many consequences, but both researchers and the public have recently highlighted one particular concern related to ML applications and, by association, AutoML systems: fairness.
Admittedly, the concept of fairness is challenging to define, with many varied attempts in the literature, and there is still no consensus on the matter~\citep{sahu19}.
This state of affairs is because fairness is not an objective statistical notion but a subjective socioethical concept~\citep{memo21}, one that can change over time and between problem contexts.
Indeed, different cultures can have discordant perspectives on what fairness means and how important it is.
That said, for decision-making, fairness is principally defined as the absence of bias/favouritism towards an individual or group based on an inherent characteristic~\citep{ch17}.
These `protected' characteristics usually include age, religion, gender, skin colour, ethnicity, etc.
Other definitions of fairness may focus on either individuals or groups, seeking similar treatment for different entities.
This discussion already leans into a deep topic, with distinctions sometimes drawn between equality and equity; the former allocates every entity the same resources/opportunities, while the latter acknowledges varying circumstances and aims for equal outcomes.

From a technical view, many metrics have been proposed for fairness in ML.
Consider a binary classifier with predictions divided into two groups, each associated with a typical confusion matrix consisting of true positives ($\mathrm{TP}$), false positives ($\mathrm{FP}$), true negatives ($\mathrm{TN}$) and false negatives ($\mathrm{FN}$).
Attempts to ensure fairness may seek to satisfy the following constraints as defined across the two subgroups: (1) demographic parity, i.e.~equal $\mathrm{TP}+\mathrm{FP}$, (2) predictive parity, i.e.~equal $\mathrm{TP}/(\mathrm{TP}+\mathrm{FP})$, and (3) equalised odds, i.e.~equal $\mathrm{TP}/(\mathrm{TP}+\mathrm{FN})$.
However, barring exceptional circumstances, it is generally impossible to satisfy the equality of all such metrics at the same time~\citep{klmu17}, and the above three are often cited in the `impossibility theorem' of machine fairness.
Therefore, trade-offs and prioritisations based on problem context are required to apply fairness in practical ML.
In the abstract sense, this may mean, for instance, choosing to prioritise equality/equity for either individuals or groups but not both.
Unfortunately, a theoretically fair ML model may still lead to biased decisions in practice because the decision-making process is ultimately subject to human interpretation and operational environments.
Worse yet, while it depends on the dataset, enforcing outcome equality between subgroups often comes at the cost of overall model accuracy.

While fairness is difficult to achieve in practice, unfairness is much easier to identify.
For instance, one can define six forms of discrimination~\citep{memo21}.
Direct discrimination occurs when an outcome, e.g.~demographic parity, is influenced by the protected features of individuals, while indirect discrimination is based on neutral non-protected features that are not independent of protected features.
An example is a zip code, which, despite codifying spatial information, is often associated with ethnic and socioeconomic factors.
Systemic discrimination refers to policies or behaviours discriminating against a subpopulation, e.g.~if a stakeholder intentionally excludes people with disabilities during data curation.
Statistical discrimination assumes individuals within a group are represented by the average qualities of that group.
One can also classify discrimination as explainable or unexplainable, depending on whether an imbalance in outcome across groups, e.g.~predictive parity, has a clearly understood cause.

The important point here is that, due to the subjectivity of fairness, the concept is critical to the topic of HCI in AutoML.
Stakeholders must communicate to the machine what they deem to be fair, while the machine must communicate to the stakeholders whether the model satisfies this requirement.
Fortunately, this endeavour can springboard off existing development in explanatory mechanisms; AutoML systems only need to incorporate some methodology to detect/identify discriminatory outcomes.
To that end, one research effort proposes to use a causal model to capture underlying biases, assessing relationships between protected features and ML model outcomes~\citep{kulo17}.
Indeed, supported by explicitly causal models, the impact of ethically sensitive attributes on other attributes and ML outcomes can be presented to decision-makers, thus evaluating fairness.
Of course, given the increasing attention to trust-related matters in the AI community over recent years, several AutoML vendors have already made moves in this area.
Commercial frameworks such as H2O Driverless AI~\citep{iy19} and DataRobot AutoML~\citep{da21} now include fairness metrics so that developed ML models can be evaluated in this respect.
Nonetheless, the broad survey of existing AutoML tools closely linked to this monograph~\citep{scke21} indicates that this is very much a nascent topic, with few packages currently implementing debiasing modules.
More sophisticated interfaces for the analysis of fairness will likely arise as the conversation on equitable ML continues.

\subsubsection{Bias Mitigation Through Human-computer Collaboration}\label{mitigate_bias_collab}

In real-world applications, ML models do not exist in a vacuum; end-users either interact with them directly or are impacted somehow by their outcomes.
Thus, any biases internalised by an AutoML system can have adverse flow-on effects on society, e.g.~by hindering the ability of individuals that are discriminated against from partaking equitably in socioeconomic activities.
Naturally, there have been various ways technicians have approached mitigating these biases, and traditional methods fall under three main categories: preprocessing, in-processing, and postprocessing~\citep{daon17}.
Unsurprisingly, as confirmed by experiment~\citep{har20}, combining methods at all three stages of processing can be the most effective alleviation of biases inherent within propagating data.

Before modelling, preprocessing approaches explore data transformations to eliminate discriminating factors within incoming data.
In-processing methods focus on the ML model instead, imposing fairness constraints on learning algorithms during model training.
Postprocessing is carried out after an ML model has been generated, usually by reassigning sample labels in holdout sets used for testing \citep{apga14}.
However, at present, the overall effectiveness of these techniques still seems to be lacking~\citep{chva21}.
A significant obstacle is still the identification of biasing factors, which is why explainability research is currently so appealing in this field.
Explanatory approaches boost accountability for model outcomes that usually slip under the radar for more conventional, albeit rigorous, probing methods.
For instance, the integration of factual/counterfactual explanations into AutoML systems may prove useful for managing the risks arising from automation bias, clarifying reasons for decision-making outcomes and indicating which scenarios/conditions may change the results.

Now, it is challenging to decide what unbiased fairness is.
Technically, as discussed in Section~\ref{outcome_fair}, it is actually impossible to satisfy all reasonable metrics for fairness.
Therefore, generating an ML model with outcomes that are genuinely representative of a problem context and data environment requires asking a simple question: what matters?
On the other hand, answering this question is no easy feat; it depends on the nuances of the social contexts in which an ML application is run/deployed, the risks associated with data curation/preparation, and the dangers of mishandling these risks.
Until context-intelligent AutonoML systems arise, as speculated in Section~\ref{roles_inter_open_env}, human judgement and expert knowledge will always be required to determine standards of bias/fairness in the design, implementation, operation and deployment of ML models.
This human expertise may be sourced from many disciplines, including the humanities, social sciences, law, and ethics.

Of course, stakeholders must have sufficient HCI-enabled understanding/oversight of AutoML system internals to identify the reasons for discriminatory behaviour and avoid vulnerable groups suffering from bias.
This scenario of necessary human intervention happens more often than some automation purists may expect.
For instance, researchers investigated a model designed to assess employee churn at the Xerox corporation~\citep{mada19}, which predicts how long workers remain in employment.
The learning system identified that a key feature inversely correlating with employment duration was commute time.
This association seems reasonable, and a consequent decision to maximise employee retention might be simply to prefer locals.
However, after careful consideration, managers identified that commute time is indirectly a protected variable, as the company was located in an affluent area and employees with a lower socioeconomic status were based further out.
Subsequently, this criterion was eliminated from model inputs.
It is thus evident that biases can easily slip through an ML application without close human monitoring and other collaborative interactions. 

The upshot is that AutoML systems should ideally be designed with human-computer collaboration in mind, at least for the modern era.
A practical approach to human-in-the-loop decision-making involves algorithms producing recommendations or sets of options while humans double-check or select ML solutions that are both good and fair.
This mode of interaction does admittedly increase human responsibility in designing, implementing, operating, monitoring and evaluating both AutoML systems and their generated ML models, so frameworks must ideally facilitate stakeholders not just identifying the biases of others but also their own.
To this end, clarity is critical.
Working towards transparent algorithms/mechanisms and explainable results can grant humans better judgement over how much weight to give any ML solution.
As discussed in Section~\ref{explanation_type}, this is partially achieved by ensuring AutoML interfaces support a diversity of analytic tools and visualisations with which human biases may be identified.
Additionally, just as how ensembles of ML predictors compensate for individual weaknesses~\citep{ruga05}, it is worthwhile considering, where practical, the comparison of predictive algorithms alongside human decision-makers.
While this does not always show whether human or machine is correct, discrepancies indicate that an error exists somewhere, which at least justifies the need to deploy resources for producing explanations and verifying rationale.
At the current state of research and development, these are just several achievable ways that AutoML systems can be kitted out to leverage human-computer collaboration in avoiding/managing biases.

Granted, while explainability clarifies AutoML processes/outputs and can highlight biases, how these explanations are interpreted still depend heavily on human cognitive behaviour.
For continued progress in this area, it is crucial to examine how humans do causal inference~\citep{ardi20} and how they grapple with issues of ambiguity and uncertainty in learning algorithms~\citep{va20}.
Indeed, examining the heuristic operations of a human brain can lead to the improved design/development of ML models that are both causal and explainable~\citep{shin21}.
Similarly, identifying factors that influence the perceptions and attitudes of humans concerning automated decision-making systems is essential beyond just the technical process of generating solutions and optimising predictive performance; these factors indicate how solution outcomes should be framed and communicated to users~\citep{le18}.
Some researchers have elaborated on how explanation modules can allow decision-makers to validate their thoughts and alleviate cognitive biases~\citep{waya19}.
For instance, an AutoML system could construct/show prototype model queries corresponding to different model responses, thus mitigating representativeness bias and the illusion of similarity.
Availability bias could be alleviated by displaying the prior probability for outcomes.
Anchoring bias could be reduced by using rule-based counterfactual explanations to highlight input-feature importance or lack thereof.
Confirmation bias could be tackled by discouraging backward-driven reasoning, e.g.~by prioritising input attribution in a UI display over posterior probability or class attribution.
Many proposals exist, drawing inspiration from empirical results in cognitive science, with yet another study~\citep{klba18} detailing techniques to address twenty common cognitive biases that impact the understanding of rule-based learning systems.
In time, it is expected that these mechanisms will become increasingly common as part of AutoML technology.

\subsection{The User Interface: Key Requirements}\label{user_interface}

Over the previous sections, we have broadly reviewed the \textit{who}, \textit{what}, \textit{how} and \textit{why} of HCI in AutoML as it is currently seen in the early 2020s.
One final question remains: given all the underlying concepts/themes that we have covered, are there any general principles for designing a good AutoML UI?

Evidently, a recurring theme in this review is that visualisation via GUI components is a natural way of fostering human understanding and interactivity around AutoML processes and generated models.
Many recent studies support this notion, stating that interactive visualisation has a crucial role in the comprehension, diagnosis and iterative improvement of numerous learning models~\citep{liwa17, stga13}.
Visual analytics can fill the gap between human knowledge and the insights generated by explainable AutoML systems~\citep{spsc20}.
Thus, many existing AutoML frameworks, particularly those marketed to the general public, focus on building a variety of GUIs; these aim to facilitate effective non-expert HCI at every supported phase of an ML workflow \citep{scke21}.
Of course, visualisation also supports technical stakeholders, enabling the inspection/control of system operations through which the learning process and ML solution search can be managed.
Expectations have reached a point where experts and ML developers prefer to visualise and compare multiple ML pipelines at a time rather than inspecting them one by one~\citep{onca21}.
Additionally, these stakeholders want to know how and why AutoML algorithms construct ML pipelines for specific problems.
Indeed, as concluded by think-aloud experimental investigations~\citep{drwe20}, outcome visualisation and the provision of performance metrics are some of the most critical aspects an AutoML system should support to promote trust.
Moreover, as long as human-in-the-loop collaboration remains integral to AutoML technology, explainability has additional feedback effects in enhancing the performance and reliability of ML solutions.
All of these considerations provide the context in which robust UI design principles should be fashioned.

Notably, the frontier state of AutoML technology means that there is still a dearth of research into systematising the key requirements of a UI.
However, with an expanded perspective, it quickly becomes apparent that there are many proposals in the literature covering guidelines/approaches for effective HCI with learning systems~\citep{amwe19}.
Many of these studies cover defensive mechanisms for avoiding undesirable actions made by a learning system, e.g.~via the implementation of verification processes and control mechanisms~\citep{ho20}, the mitigation of user-misleading interactions with adaptive systems~\citep{ja07}, and the enhancement of transparency/explainability for mechanised behaviours~\citep{kubu15, raco18, risi16}.
However, research prior to the current surge of attention on trustworthy ML also exists, generally considering operational matters, e.g.~balancing automated services with direct user control~\citep{ho99}.
Historically, adaptive UIs have also received significant attention in both design principles~\citep{desc16} and evaluation, e.g.~in terms of predictability~\citep{gaev08}.
However, it is argued that AI designers and developers continue to struggle with the creation of intuitive UIs that foster human understanding, belief and effective collaboration with automated learning systems~\citep{giho19}.
A well-designed UI fits squarely into the topic of ergonomics, primarily cognitive; it needs to provide essential elements that assist users in reducing workload, improving work performance, and ensuring safety.
Case studies indicate that the most successful UIs are usually associated with a well-defined structure of HCI based on obvious principles, validated theories, and actionable guidelines~\citep{sh20}.

\begin{figure}[!ht]
	\centering
	\includegraphics[width=\textwidth]{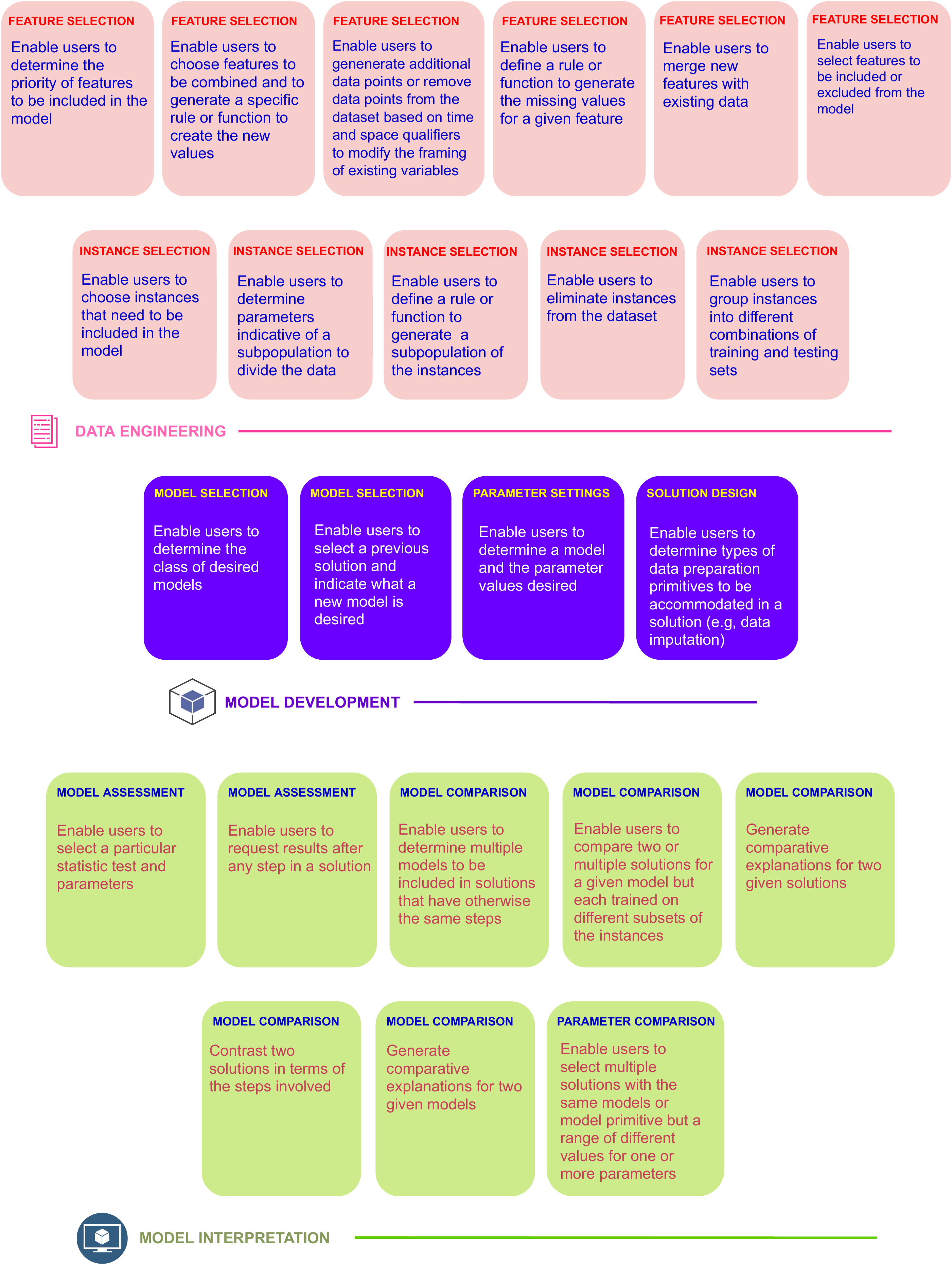}
	\caption{Conceptual schematic of HCI-related functionality that an AutoML UI should support~\citep{giho19}.}
	\label{automl_ui_design}
\end{figure}

Although UI design for learning systems, examined primarily from a communication-modality perspective in Section~\ref{multimodality}, essentially remains a topic for the broader data-science community, there have been a handful of initial AutoML-specific forays into this discussion.
One study claims that, although numerous applications may be suitably served by ML that is fully automated, humans usually possess rich domain knowledge that is valuable for constraining available data and guiding solution search procedures~\citep{giho19}.
Therefore, the authors introduced a set of UI design principles for human interactions with AutoML systems, schematised in Fig.~\ref{automl_ui_design}.
These requirements were proposed based on how stakeholders work with AutoML algorithms and what their expectations for interactions are.
Three main categories of tasks were identified, loosely corresponding to the core of the ML workflow we describe in this monograph: data engineering, model development, and the interpretation of obtained learning models.
Additionally, the authors argued that human interactions with AutoML frameworks are both iterative and nonlinear.
Specifically, human engagement is iterative because, over time, ML models are produced/examined, AutoML settings are reconfigured, and model search/production is rerun.
Human engagement is also nonlinear because stakeholders may desire to skip steps at any time, relying on defaults for necessary operations and ignoring optional ones, or they may decide to revisit previous phases of an ML workflow at will.
Therefore, UIs should ideally support these requirements.
Unlike traditional fire-and-forget approaches to AutoML in academia, there is also a strong focus in the proposal on post-production evaluation, e.g.~the UI must allow users to examine a specific ML pipeline, complete with its detailed steps of production, for comparison against challenger models.
Accordingly, the work stresses that effective HCI with AutoML frameworks is critically dependent on \textit{both} transparency and usability.

\begin{figure}[!ht]
	\centering
	\includegraphics[width=\textwidth]{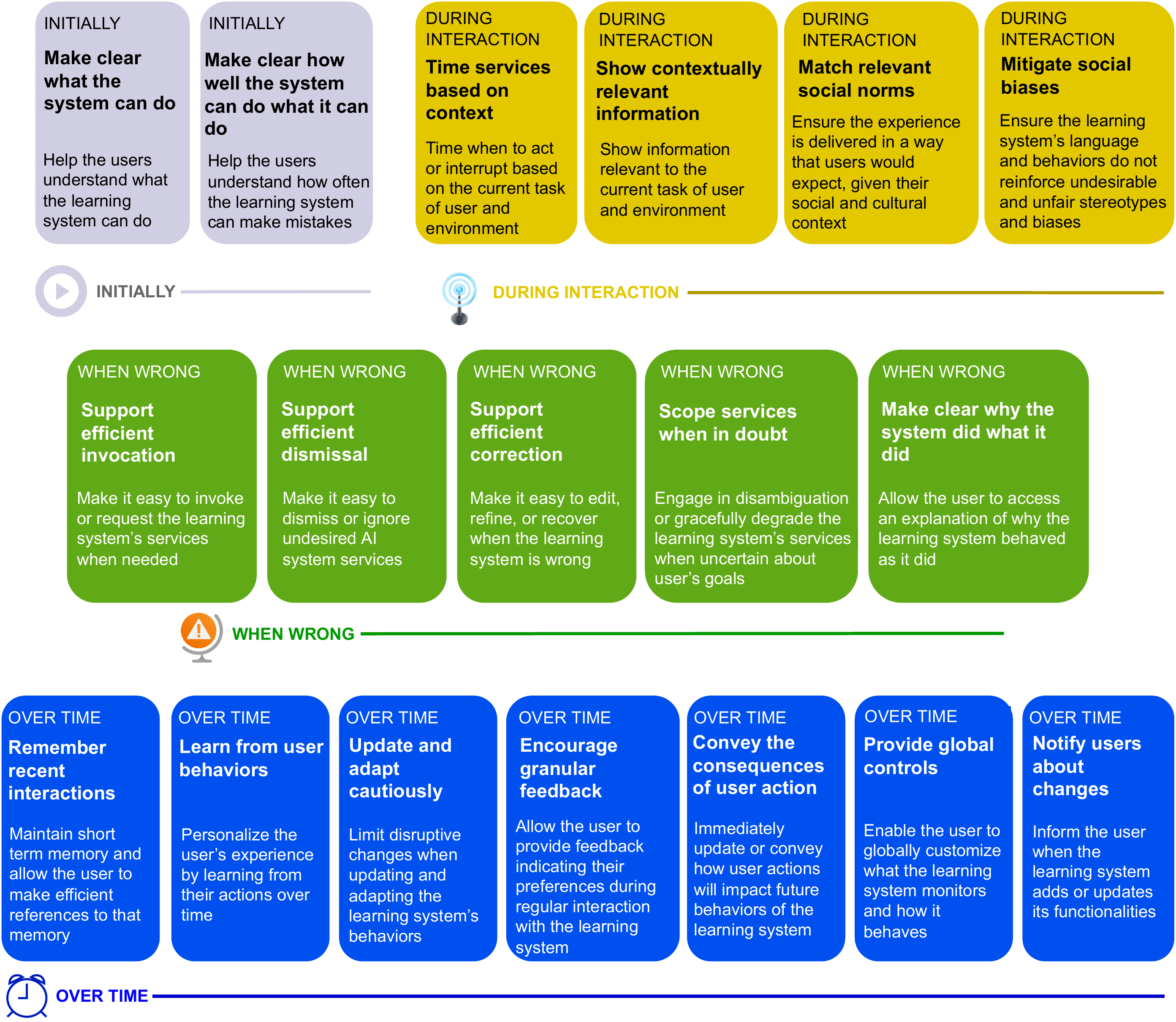}
	\caption{General design guidelines for effective interactions between humans and learning systems~\citep{amwe19}.}
	\label{interaction_design}
\end{figure}

Of course, as of the early 2020s, the conversation on HCI in AutoML has just started.
Thus, the broader concerns facing the AI community have not yet percolated into direct considerations of AutoML-specific UIs.
For instance, the proposal mentioned above~\citep{giho19} still has a strong focus on technical performance, while establishing trustworthy ML requires grappling with many other factors, e.g.~fairness, biases, and ethical concerns.
Also, it is essential to remember that, at present, AutoML is still a long way from its major goals, i.e.~comprehensive automation of the entire ML workflow; see Section~\ref{roles_inter_cmpl_constrained_env}.
This fact is evident in that the proposed UI requirements did not systematise HCI in deployment environments, let alone consider ongoing monitoring and maintenance related to changing ML models~\citep{he19}.

Unsurprisingly, there are some significant challenges on the horizon for AutoML-specific UI design.
By now, it should be clear that humans usually use automated decision-making systems if machine behaviours can be controlled or predicted; stakeholders expect to interrupt anything incomprehensible and potentially dangerous in a timely manner~\citep{sh20}.
Therefore, guidelines for UI designs should factor in human understanding and control, forcing AutoML systems to explain what they did.
Bonus points can be awarded for adaptive UIs that learn from user behaviour and better facilitate desired interactions in the future.
However, two main complications challenge effective UI for collaborative interactions~\citep{yast20}: uncertainty around the capabilities of an automated learning system and complexity related to system outputs.
Presumably, efforts to grapple with these obstacles will continue to filter in from the broader literature, with Fig.~\ref{interaction_design} exemplifying one such study, codifying 150 design recommendations from varied academic/industrial sources into 18 general design guidelines~\citep{amwe19}.
These guidelines cover best practices for immediately clarifying system capability, managing interactions, tackling errors, and dealing with ongoing operations.
Similarly, the IBM Design website\footnote{https://www.ibm.com/design/ai/} also provides guidelines for design foundations, albeit from broader social perspectives discussed in Section~\ref{improve_outcome}, e.g.~ethics, fairness, and accountability.

Regardless, it seems inevitable that these low-level abstractions and implementations of UIs will, in time, solidify within the AutoML technology that is publicly available.
As mentioned before, the steady expansion of the automation endeavour along the entire ML workflow, combined with escalating concerns around trust in ML, will require system developers to grapple with these facets one way or the other.
It may be informative then to zoom out of the details for a while and forecast, with a high-level perspective, what happens to HCI once humans no longer \textit{need} to be involved in any system operations.

\section{Interacting with AutoML Systems: Constrained but Fully Automated}\label{roles_inter_cmpl_constrained_env}

Bluntly speaking, the concept of HCI was never fundamental to the initial drive for AutoML.
After all, a primary motivator for automation is ensuring that operations of interest can be processed with as little human interaction as possible.
To this end, the field of AutoML has made great strides, yet it remains far short of its end goals.
In most modern AutoML frameworks~\citep{scke21}, humans still participate heavily in configuring the system by, for example, controlling the selection of data transformations, ML models, and HPO algorithms~\citep{yawa18}.
Moreover, not only is the technology \textit{semi}-AutoML in practice, many existing packages focus only on subsections of the whole ML workflow, e.g.~feature engineering, HPO, or NAS~\citep{zhzh21,kemu20,doke21,scke21}. 
Therefore, at present, most AutoML services are seen as no more than tools to assist technical users in developing ML models for practical problems.
Accordingly, ergonomic approaches to HCI have arisen as a necessity, not a desire, primarily to support the technical control that semi-AutoML packages require.
That is not to say that there are no HCI mechanisms devoted to the broader social issues discussed in Section~\ref{improve_outcome}.
However, the roles and modes of human interactions are predominantly driven by getting AutoML to work and work well.

Of course, the democratisation angle of AutoML remains.
The field strives to enable domain-knowledgeable stakeholders without a technical background in data science to engage in and benefit from ML theories/practices~\citep{shzg19, giho19, xiwu21}. 
To pursue this agenda and truly remove the need for an ML technician, an AutoML system must be virtually end-to-end, capable of processing almost the entire ML workflow independently.
We exclude the stage of problem formulation and context understanding from this requirement; see Section~\ref{roles_inter_open_env}.
Essentially, given an input dataset, a genuine AutonoML system must be able to find/construct an optimal ML pipeline on its own, deploy it, and then maintain its performance over time within a production environment.
It should also be capable of dealing with diverse input types, such as images, sounds, text, and tabular data. 
As of the early 2020s, there are only a few AutoML products that are sophisticated enough to even approach these requirements, such as AutoML by DataRobot~\citep{da21}, Driverless AI by H2O~\citep{iy19}, Fully-AutoML~\citep{zhzh21}, and Auto-sklearn 2.0~\citep{feeg21}.
The first two examples, AutoML by DataRobot and Driverless AI by H2O, are commercial tools marketed as automating the entire ML workflow, from data engineering to the MLOps phase, complete with simple forms of monitoring/maintenance.
The other two examples, Fully-AutoML and Auto-sklearn 2.0, only focus on data engineering and model development.
All are impressive tools/systems, but, at the current time, it would not be advisable for a client to leave their ML project entirely up to the machine without any oversight.

Granted, we do not necessarily advocate the removal of human intervention.
However, there is no reason not to expect the default behaviour of AutoML frameworks to get better and better over each new generation of research and development.
So, we now imagine that an AutonoML system with sufficient technical capability is finally developed, one that can only solve well-defined ML problems -- its understanding of context is `constrained' -- but one that can do so with full automation.
Does the spirit of HCI change?
To consider this question, Section~\ref{search_strategy} first discusses the possible ways in which AutoML theory/technology will progress to the ideals of AutonoML; improving search strategies is crucial to this advancement.
Section~\ref{human_roles_constrained} then considers what the roles and modes of human interactions are likely to be when engaging with these fully automated systems.
Finally, we consider what lies beyond constrained ML problems in Section~\ref{beyond_constraints}, motivating an incoming discussion on AutonoML within open-ended environments.

\subsection{Served by Superior Search Strategies}
\label{search_strategy}

In general, the core aim of ML is to find a mathematical model that best approximates some desirable query-response function.
Naturally, a great deal of human effort goes into concretising this desirable function and the associated data environment in which it hides.
For now, we treat this initial phase of an ML workflow, i.e.~problem formulation and context understanding, as solely the domain of human cognition.
In so doing, we assume that, by the time an ML problem is passed to an AutonoML system, it is well defined and firmly constrained.
Given this assumption, it is reasonable to state that all remaining \textit{operational} human interactions go into supporting the search for an optimal ML solution.
These can involve transforming the data environment, defining searchable spaces for HPO, supplying the right computational resources, considering how best to adapt models for a changing desirable function, and so on.
It stands to reason then that full automation of the ML workflow depends critically on improving search strategies, for every AutonoML mechanism, to the point that humans need not guide them.

Historically, advances in optimisation theory/practice have already had an enormous impact on data-science applications and technology.
Notably, these advances have often shown that expert knowledge is unnecessary -- sometimes inhibitory -- to effective solution search in various narrow domains, e.g.~protein-folding~\citep{juev21}, Atari-based games~\citep{gusi14}, Go~\citep{sihu16,sisc17}, Chess~\citep{sihu18}, and so on.
A pioneer of modern reinforcement learning, Richard Sutton, argues in a blog post that ``the biggest lesson that can be read from 70 years of AI research is that general methods that leverage computation are ultimately the most effective, and by a large margin''~\citep{su19}.
Certainly, machines are not beset by human cognitive flaws and limitations, nor do they struggle with the same learning curves~\citep{me19}; again, we emphasise this is in the context of narrowly defined ML problems.
Modern computational resources are also vast, even if they fall short of the postulated exaFLOP that a human brain operates at, and leveraging them efficiently/intelligently can achieve significant outcomes.

As an example of how AI exceeded human skill by relying on superior search strategies, consider the game of Go.
Traditionally, ML efforts at mastering games have relied heavily on human expertise.
Even AlphaGo~\citep{sihu16}, the first computer program to beat a professional Go player, involved a deep neural network trained in a supervised manner on previous games played by human experts.
However, it also drew benefits from self-play; after a certain level of proficiency, it played millions of games against itself and used reinforcement learning to improve its capabilities.
During this self-play phase, it was essentially provided only the rules of the game and a metric to evaluate board state, e.g.~win-loss.
In effect, it ceased to rely on handcrafted domain insights and human heuristics for making moves, eventually developing its own tactics for optimal gameplay.

Of course, an initial expert-knowledge bootstrapping can still influence a subsequent self-play process.
So, the successor, AlphaGo Zero~\citep{sisc17}, eliminated human expertise embedded in the training data entirely.
Its search environment was constrained by the context of the problem, i.e.~the rules of the game, a performance metric, and the input structure of data corresponding to the board state.
Then, starting from entirely random moves and initialised parameters, the system reached significant proficiency at the game, able to beat top human players within days, and achieved all this without any human data or guidance.
Without human cognitive biases and preconceptions related to the game, AlphaGo Zero was effectively able to explore non-standard tactics beyond conventional Go knowledge.
Since then, AlphaZero~\citep{sihu18} has only continued this trend of advancement, able to employ a single network architecture to search and learn at expert level for multiple games, i.e.~Go, Shogi, and Chess.
These achievements demonstrate the potential of generic automated search-and-learn algorithms to surpass human performance in narrowly defined problem contexts.
Unsurprisingly, self-play strategies have also been engineered into AutoML systems, e.g.~AlphaD3M~\citep{drkr18}.

Crucially, the above example shows that the key to significant advances in automated search relies on high-level insights, e.g.-the notion of adversarial self-play to explore a space logically but, to humans, unintuitively.
In a similar fashion, ignoring revolutionary changes in computational hardware, AutoML is unlikely to make sufficient progress towards full automation without significant advances in theory/practice.
Search spaces for HPO are typically immense and multi-dimensional, meaning that random search has often been found competitive with principled optimisation methods~\citep{gasa18}.
A possible saviour then might be found in the notion of meta-learning, whereby meta-knowledge, experience built from encounters with other ML problems, is used to supercharge advanced search strategies, e.g.~by warm-starting searches or recommending parameter values.
This principle is highly appealing, grounded in lifelong biological learning, and is likely a hard requirement for opening up cross-context capabilities for AutoML; see Section~\ref{roles_inter_open_env}.
Unfortunately, the current accumulated successes of meta-learning have been inconsistent at best, e.g.~in terms of recommending search spaces for ML models~\citep{ngke21,albu20}.
It is presently unclear which settings of an AutoML system should be tuned similarly for similar ML problems, nor is it clear how to define that similarity between ML problems effectively.
Additionally, while the field currently exists in an era of big data, the AI community may not have enough instances of \textit{datasets} to begin building robust meta-models.

Despite all the obstacles, it seems obvious, conceptually speaking, that meta-learning should be central to advanced AutonoML.
As early as 2009, a generic architecture was proposed to fully automate the process of developing, deploying and maintaining online predictive models~\citep{kaga09}.
It was an attempt at AutonoML before the term `AutoML' was even propagated in the mainstream.
Relevantly, meta-learning was considered an integral part of ML solution search alongside local learning and ensemble learning, with optional support for the insertion of expert knowledge.
Although modern AutoML systems are arguably more advanced in various aspects, this framework remains a possible hint of how fully automated AutonoML may eventually appear.
Its generic nature allows different data-engineering methods, predictive models and adaptation strategies to be easily plugged in, with automated interaction channels among these components explicitly defined.
It supports both batch learning and incremental learning in a single framework and, uniquely among modern AutoML offerings, provides multiple complex adaptation approaches for lifelong learning.
That said, the research did not have an agenda of forced automation, so the architecture also describes interaction methods between system operators and various elements within the whole framework.
Certainly, technicians can access specific modules and manually intervene in system operations at will, with most interventions relating to process control, parameter configuration, and performance evaluation.

There are a handful of other systems that also incorporate meta-knowledge, such as the Fully-AutoML framework~\citep{zhzh21}.
Without human guidance, this system can search a coarse-grained space of data-engineering and model-development components to construct an ML pipeline for different types of input data, e.g.~image, text, video, speech, and tabular.
Most importantly to this discussion, the proposed framework has an offline phase, during which an evolutionary algorithm searches for pipeline configurations with good empirical performance on various data sets, subsequently storing them within a given repository.
These pipelines are employed as lifelong knowledge anchors and are associated with various meta-features, including descriptive characteristics for the datasets they solve, as well as ML task type, domain, and so on. 
The promising ML pipelines generated from previous experiences are then used to initialise a population of solutions that an evolutionary algorithm works with to propose a good candidate for a new ML problem.
Recommended pipelines can consist of ML components tasked with various operations: data preprocessing, feature engineering, model selection, hyperparameter optimisation, and ensemble aggregation.
Similarly, though not by evolutionary means, Auto-sklearn 2.0~\citep{feeg21} also employs meta-knowledge to warm-start ML pipelines of preprocessors/predictors and their hyperparameter configurations.

For now, meta-learning mechanisms remain rare amongst AutoML services, so it is difficult to assess whether their present implementation is generally effective or simply proof-of-concept.
Indeed, it is time-consuming and difficult for humans to curate an appropriate collection of meta-knowledge, while automated curation risks the inclusion of low-quality meta-data.
In addition, the effectiveness of meta-learning appears very sensitive to many factors, with empirical results indicating that minor differences between previous data sets and a novel ML problem can degrade performance severely due to biases in data~\citep{bama19}.
Nonetheless, the promise of meta-learning to accelerate and enhance search strategies continues to draw research, which grows ever more sophisticated.
For instance, to enhance the generalisation of meta-learning, a Wasserstein Generative Adversarial Network~\citep{arch17} was recently proposed to act as a surrogate model for the generation of meta-data~\citep{liti21}.
A self-adaptive meta-model trained on this artificial meta-data was then investigated, seeking a good trade-off between accuracy and run-time.

Overall, while the above examples are only initial forays into AutoML-specific meta-learning, there is no in-principle reason to doubt that search strategies will continue to improve.
In time, default AutoML operations will likely become sufficiently performant to no longer warrant technical oversight.
However, not all phases of an ML workflow are equal.
The pathway is arguably the longest for the stage of continuous monitoring/maintenance, and previous comprehensive reviews found research/development efforts here to also be the most nascent~\citep{kemu20,doke21}, even though many real-world ML models operate in dynamically changing environments.
In general, the design and implementation of adaptive learning systems face six essential challenges~\citep{zlbi12} that remain incompletely addressed: making adaptive systems robust and scalable, tackling realistic data, enhancing usability and trust, effectively integrating expert knowledge, considering different application demands, and translating algorithms into software.
Unfortunately, this remains largely irrelevant to modern AutoML, as a survey of packages/services found that most still focus on static data~\citep{scke21}.
Those that do not constrain their focus tend to have simple mechanisms for adaptation, although there is insufficient evidence to judge whether that simplicity is a negative.

Regardless of the present state, it is inevitable that AutoML will eventually have to grapple with adaptation, if only because real-world applications demand it.
Indeed, in the wake of the generic proto-AutonoML architecture proposed in 2009~\citep{kaga09}, related efforts have shown the utility of complex adaptive ML systems in the process industry, specifically with soft sensors~\citep{kaga10, kaga11, baga17}.
Consequently, some researchers have decided that, if AutoML packages are fire-and-forget, perhaps it is possible to call them on demand as part of an adaptive strategy~\citep{sabu16b}.
This thread of research was recently progressed even further, exploring the artificial combination of several open-source AutoML toolboxes and a drift-detection mechanism~\citep{ceva21}.
However, while valuable, such approaches remain workarounds; genuine AutonoML systems will probably need to internalise adaptive mechanisms if they are to operate in any sophisticated manner.
Accordingly, other efforts have explored how best to automatically develop/deploy flexible adaptive strategies~\citep{bafa21}.
Ultimately, the takeaway here is that there are many interesting technical challenges left for researchers/developers to tackle in the field of AutoML.
However, to contemplate HCI, it is not unrealistic to fast-forward and envisage the end-point of full automation, at least in terms of well-defined ML problems.

\subsection{Roles and Modes When Human Involvement Is No Longer Required} \label{human_roles_constrained}

\textit{Technical and business stakeholders shift their focus to problem formulation and context understanding.}
This reprioritisation is the obvious consequence of engineering AutonoML frameworks capable of fully mechanising all ML workflow phases dedicated to ML solution search.
Technicians, in particular, are freed from a wide range of tedious tasks, including managing the selection/generation of features, recommending model types to evaluate, and guiding search through HPO configuration spaces.
Instead, computers are assumed to automatically run these operations with sufficiently good performance, subject to the bounds of computing memory, time, input types, and solution-acceptance criteria.
Accordingly, human resources can be redirected to more innovative zones, e.g.~refining ML task definitions/constraints, better understanding how problem and context interrelate, sourcing more informative data sources, etc.
In essence, stakeholders focus on building the best sandbox in which AutoML will proceed to play.
For example, the operational environment of learning models playing Chess or Go is restricted by specific invariant game rules, alongside input-data structures representing the size and type of game boards involved.
For diagnostic models used in radiological imaging, the learning environment may be constrained to specific disease types and scan formats.

The redirection of human resources can be particularly valuable, as some settings may require substantial effort in defining the problem context, especially for complex predictive/prescriptive models that manage multiple intertwined tasks.
For instance, the `brain' of a self-driving car will be subject to many factors such as physical response times, traffic laws, speed limits, road signage, lane markings, and weather conditions.
The point here is that stakeholders no longer instruct the automated algorithms on how to solve a problem; they merely dedicate time and effort to ensuring that definitions/constraints are sound and watertight.
These include appropriate metrics for evaluating outcomes and performance, e.g.~in a reinforcement-learning setting.
The AutoML system is then expected to leverage available computing power and conduct search/learning processes without the need for any human intervention, possibly supported by meta-knowledge.
Without directly inheriting the biases and limitations of human intelligence and cognition, AutoML may discover entirely new ways of solving a problem that humans have never seen before~\citep{me19}.
Technicians can, of course, revisit a search history to validate results according to predefined criteria and then choose whether to accept them. 

The other clear advantage of an AutonoML framework that is self-managing and effectively end-to-end is that the role of an end-user can also shift.
With the operation of ML applications ceasing to require technical expertise, ML solutions no longer have to be built for an end-user; they can be built \textit{by} an end-user.
This assertion is simply the logical culmination of the democratisation drive behind AutoML.
Research suggests that it is already feasible with current AutoML technology to give end-users a degree of that agency, allowing them to build models for their own domain-specific datasets and initial settings~\citep{leku19}.
Admittedly, the empirical results of the study are preliminary and only involve basic models.
Nonetheless, this role is likely to evolve in parallel with UI improvements and simplifying representations of AutoML.
For instance, if stakeholders currently desire explainable ML models from AutoML systems, they are often forced to interact with low-level details, deciding whether configuration spaces should include/exclude predictors such as decision trees or multilayer perceptrons.
In contrast, for fully automated AutonoML frameworks, end-users should need to do no more than state a preference for explainability, leaving all the rest up to the system.
Many such advancements are, in principle, technically trivial, e.g.~translating one marked checkbox into a series of true/false attributes associated with ML algorithms in the codebase.
The onus in these particular cases is on developing robust ontologies to map human desires into machine-interpretable operations.

\textit{The strengths and weaknesses of operators and AutonoML systems become complementary.}
This eventuality is closely related to the previous one.
Rather than humans and machines both wrestling with solution search, entities are now able to solely prioritise the strengths of their biological/mechanical intelligence, covering for the weaknesses of each other.
In recent years, this perspective has become particularly popular, eschewing AutoML as an outright replacement for human roles.
Instead, the purposes of AutoML are increasingly seen as augmenting the capabilities of users~\citep{luad21} and countering the constraints/limitations of human cognition~\citep{kova21}.
Naturally, it is still the role of domain experts to precisely define/constrain a problem and its context, as the presently hypothesised form of an AutonoML system has neither individual agency nor a versatile understanding of concept.
However, in the ideal case, operators do not expect to directly interfere with the internal mechanisms of their AutonoML `assistant'.

For an example of this duality, consider the case of medical diagnostics, where an ML task may be to map particular patients to sets of treatments, with careful consideration of potential side effects~\citep{liaj21}.
Maintaining concentration for extended time periods is challenging for humans, especially with high-stakes monotonous tasks, e.g.~reading through large amounts of clinical data.
Computers do not suffer this form of processing fatigue.
On the other hand, despite their extremely effective search strategies, AutoML systems have neither the rich domain knowledge that humans have acquired nor the cognitive mechanisms to form connections between concepts.
This lack of context-awareness can make it impossible to understand the implications of data and outcomes.
For instance, as noted by the medical diagnostics study, human experts can identify that a missing data feature might be a critical indicator for unmeasured information or hard-to-quantify objectives concerning a minority subgroup expressed in a dataset.
Specifically, this is an outcome of humans applying the context understanding that technically proficient AutoML systems are incapable of~\citep{xiwu21}.

The primary remit of an ideal but constrained AutonoML system is also confined to four out of five phases of an ML workflow.
One can obviously speculate, but, for now, such systems do not have the same flexibility to engage with a real-world data environment as humans.
For example, clinicians can adjust rigid prescriptions fired out from an ML solution by collaborating with patients and taking in their feedback.
Clinicians can also come up with creative ways to alleviate side effects.
This argument for rejecting automation as an outright replacement for human roles is common.
For instance, even though an automated learning system could theoretically observe and interpret X-ray images, the expertise and capabilities of a radiologist go far beyond these functions~\citep{ar20}.
A radiologist will regularly handle patient-facing tasks, e.g.~ultrasound, fluoroscopy, biopsy or consulting, and they will also handle multi-disciplinary work, e.g.~auditing results and reviewing discrepancies~\citep{ha20}.

Ultimately, a human brain and an idealised AutonoML system, without context understanding, are both complementarily valuable for decision-making.
In critical high-stakes fields, these decisions should likely remain the responsibility of qualified humans, but, even there, AutonoML predictions/prescriptions may be supplementarily helpful as long as they are weighted accordingly.
In essence, humans and AutonoML systems will ideally reach higher performance in both speed and accuracy when acting together than apart.
For example, an automated driving support system may be better at keeping a car within a lane and identifying sudden hazardous situations.
At the same time, human drivers may be better equipped with context understanding to make significant routing decisions or select responses to certain hazards~\citep{liaj21}.

\textit{Humans become mentors and supervisors to AutonoML systems.}
This eventuality ensures the effectiveness, safety and trustworthiness of such systems~\citep{xiwu21}.
After all, proficient search is not perfect search.
We have just noted how a lack of context understanding is limiting, and the discussion around fairness ambiguities in Section~\ref{outcome_fair} supports this, but there are still technical complications in ML that will likely never fade.
For instance, without a complete analytical description of a search space, it is impossible to know if a locally optimal ML solution is near a global optimum.
There are many other challenges, and these, including necessary trade-offs, are mentioned in a previous review~\citep{kemu20}.
Moreover, it may not even be the AutonoML system that is guilty of flawed ML solutions; Section~\ref{biases_impacts} highlighted that it is easy for human cognitive biases to be internalised by a machine.

The upshot is that responsible usage of AutonoML requires oversight by whichever stakeholders are running an ML application.
Beyond just validating outputs, humans also need the ability to review an entire ML workflow to comprehend/justify why specific models and hyperparameters were selected and how predictive outcomes were derived~\citep{xiwu21}.
This careful appraisal reduces the chance of technical flaws arising in the system, e.g.~a performance metric that does not work as advertised.
It also improves the outcomes of AutonoML along the lines discussed in Section~\ref{improve_outcome}, e.g.~ensuring that system outputs are fair, nondiscriminatory, ethical, legal, and harmless to minority groups in society.
In some cases, this will likely involve the correction of data bias to avoid its propagation.

So, how is this different to proposals made for modern-day AutoML?
It is simply a matter of degree.
The freeing up of technical labour means that more human resources are available for this oversight, improving the verification of outcomes according to various pre-defined criteria, e.g.~safety and security.
This migration of labour has already been seen to a limited extent with current AutoML tech, where the effective automation of the model-development phase has shifted stakeholders to monitoring/auditing outcomes alongside focussing on problem definition, data collection, and deployment~\citep{yawa18}.
It stands to reason then that human resources will continue to be redistributed once AutonoML systems become proficient at transforming data, optimising ML models, embedding ML solutions within production environments, and maintaining them.
Naturally, the flow-on effects of increased oversight include boosting confidence in the use of AutonoML and promoting further uptake.

\textit{All stakeholders become capable of improving AutonoML performance.}
We again emphasise that full automation in constrained environments means that an AutonoML system can consistently find, deploy and maintain reasonable solutions to arbitrary but well-defined ML problems.
It does \textit{not} mean that solutions cannot be refined by including expert knowledge that a meta-learning module may not have encountered yet.
So, it is a legitimate role for technicians to continue guiding AutonoML search, especially if they are privy to useful external information.
Certainly, empirical results show that, following the inspection of HPO search spaces and ML models available to an AutoML tool, technical users are reasonably effective at identifying hyperparameter ranges in which better ML solutions may exist~\citep{wami19}.
Any decent modern AutoML package will support the subsequent exploitation of a refined search space.
However, a consequence of full automation is that the line between ML experts and non-experts is now blurred.
As mentioned earlier, in terms of designing robust ontologies, a significant portion of this ideal automation can be achieved with good HCI design.
Most low-level technical functions are already present in a system; their operations merely need to be organised and bundled into accessible high-level interactions.
Thus, if an AutonoML system is automated correctly, it should be far easier for end-users without an ML background to generate solutions and provide feedback via such interactions~\citep{amca14, yasu18}.
For instance, while technicians are capable of wrestling directly with complex HPO subspaces, a search space could be constrained by simply having a non-expert approve/reject a model recommendation.
This process is similar to having a single evaluation inform a Bayesian optimiser where it should next sample solution space.

Granted, the AI community is in the midst of an ongoing debate about whether computerised learning should adopt human expertise or strike it out alone.
We noted in Section~\ref{search_strategy} that ML performance in certain applications skyrocketed once the limitations of human knowledge/intelligence were eliminated from the associated approaches.
Some have even argued that it is general methods leveraging computation, not human expertise, which is responsible for the greatest successes in AI over the past 70 years~\citep{su19}.
However, others argue that all successful AI systems have ``required substantial amounts of human ingenuity'', even if that intelligence has supported automated algorithms in varied ways, e.g.~by engineering specialised network structures and defining particular training regimes~\citep{br19}.
Unsurprisingly, these counterarguments raise many examples showing where ML technology falls far short of human capability.
For instance, a modern recognition system in an autonomous driving car was easily fooled that a traffic stop sign with white stickers was a 45 mph speed limit sign.
The neural networks involved could not identify nuances that humans easily handle, instead relying on generic patterns, i.e.~stop signs are red and speed limit signs are white.

Extrapolating current research progress in AutoML and AI, it is likely that full technical automation will be attained ahead of any proficiency in context understanding; see Section~\ref{roles_inter_open_env}.
So, in practice, it is likely that stakeholders will continue to benefit from encoding their rich expertise for use by AutonoML systems, e.g.~via suggestions of promising model architectures or the refining of constraints.
This requirement may even be essential for high-stakes applications to avoid errors based on a lack of prior knowledge.
A hint of how improved knowledge insertion may work can be gleaned from research involving an ML application that predicts patient mortality risk in an intensive-care unit~\citep{gefr20}.
The authors of the study were keen on automatically extracting problem-specific expertise from domain experts and integrating this into ML models via human-computer collaboration.
Specifically, a gradient boosting model was used to transform raw input data into a set of simple rules representing sample subpopulations, each identified by some condition being satisfied.
Clinicians were instructed to assess the relative mortality risk for each rule-associated subpopulation compared to that of the whole.
The differences between human assessments and those made by data-driven ML models were then computed and ranked.
Finally, a new model was generated, penalising unreliable rules with high levels of human-computer disagreement.
In this way, the ultimate model hopefully contains the best of both worlds, having incorporated human expertise.
Notably, the clinicians in this study did not need to understand ML to improve the model, nor did they have to generate data instances to represent their compiled knowledge.
One can only presume that an ideal AutonoML system will similarly facilitate the direct inclusion of expert knowledge from those that do not understand ML.

\textit{Continuing legal responsibility forces humans to ensure AutonoML systems are transparent.}
This assertion comes down to the fact that, while the debate around AI and legal liability has recently surged liability has recently surged~\citep{ki16}, the full automation discussed in this section is unlikely to trigger any revolutions in jurisprudence.
A proficient ML model optimiser/deployer/maintainer is still a machine following specific orders, i.e.~it solves a strictly defined ML problem.
It does not have the individual agency to formulate a problem of its own.
So, while humans may find themselves content to let AutonoML do all the technical work, legally conscious individuals will be forced to maintain stringent oversight.
This interaction may include overriding decisions made by AutonoML systems in emergencies.
In short, legal responsibility ensures that human involvement cannot be entirely removed~\citep{bobr17}.
A simple analogue to an AutonoML system, perhaps one day a direct example, is an autopilot system in a commercial flight.
Such a mechanism helps manage tedious and repetitive tasks, thus reducing pilot fatigue and operational error during long flights.
However, although the manufacturer of such a system may be liable for mechanical failures, no social contract has granted the autopilot responsibility for the entire flight.
It is merely a machine that assists a human in solving a problem, i.e.~flying, and its assistance is to be discarded when no longer helpful to the actual pilots.

Now, admittedly, jurisprudential issues become murkier if an autonomous system becomes as reliably capable as a human for handling general decision-making tasks.
Other than individual agency, where is the distinction?
If a self-driving car crashes with the expected frequency of a standard driver or even more rarely, is the manufacturer any more liable than the parents of the driver?
Such debates are out of scope for this monograph, let alone this section.
An AutonoML system perfected for constrained environments cannot be expected to handle concepts beyond the bounds of a formulated problem any more than an autopilot is expected to manage every malfunction and tropical storm.
Indeed, as is presently the case with AutoML~\citep{xiwu21}, human involvement will likely be necessary for even fully automated AutonoML to satisfy all the rigorous requirements of a practical application.
The consequence of all this is that legal liability will be just another socioethical factor that prevents society from accepting black-box designs for AutonoML.
Human operators will expect transparency to justify their usage as appropriate and legally defensible.
This notion entails having access to AutonoML processes, understanding them, and knowing system limitations.
It also means knowing to what extent model outcomes can be trusted and in which contexts they can be appropriately applied~\citep{liaj21}.

\textit{AutonoML teaches humans new tricks.}
Thus far, the implicit focus of developing fully automated AutonoML systems, with the condition of constrained contexts, has been to produce/deploy/maintain a `good' ML model.
However, once the limitations of human heuristics have been set aside, it can become valuable to observe \textit{how} the AutonoML system performs its job.
Does it chain ensembles when constructing an ML model?
Does it provision computational resources across a cluster in an unbalanced fashion?
Does it adapt local elements of an ML solution to blend short-term and long-term memory?
All of these are potential insights for how to conduct ML applications better.
Of course, some may argue that if the AutonoML system performs so well, then maybe automation bias becomes justified; see Section~\ref{biases_impacts}.
There may be no point in humans learning how to do ML better, particularly if the AutonoML system records these insights with a robust meta-learning module.
However, as stated before, we do not treat this idealised form of AutonoML as perfect.
Stakeholders may identify patterns, trends and correlations that the system misses, feeding them back in for future runs.
Likewise, human cognition may be able to distil AutonoML complexities, e.g.~important features arising from deep neural networks with arcane connectivity, into conceptually simplified pieces of knowledge.
Even disregarding the use of tricks learned from AutonoML to improve AutonoML, such insights may still feed back into improving problem formulation.
This outcome is especially useful when AutonoML has no context understanding.
For instance, if an AutonoML system consistently applies dimensionality reduction to throw out dataset features prioritised by data-collecting surveys, perhaps valuable knowledge lies in the neglected features; this may inform future directions for context refining and data curation.

Although AutoML is too nascent to suggest how often it will teach humans how to do ML better, there are many analogous examples among automated learning systems and within various fields.
Medicine is but one area that is ripe for ML-derived insights, in which new drugs and treatments can be found by searching and matching elements in ways that are entirely unintuitive to human experts~\citep{me19}.
Unfortunately, in many cases, associated learning systems cannot provide human-understandable explanations of how novel solutions have been found.
The onus is then on human experts to validate the results, which requires high levels of creativity, problem-solving, cognitive skills, and domain expertise.
However, once confirmed, lessons can be further extrapolated into novel discoveries~\citep{hafe17}, enriching existing knowledge within a specific domain.
As mentioned in Section~\ref{search_strategy}, the games of Go and Chess are also arenas where automated learning systems have outperformed humans with unconventional insights.
For instance, in Chess, AlphaZero~\citep{sihu18} has managed to surprise players with its focus on innovative strategies, e.g.~h-pawn thrusts~\citep{miya20}.
Similarly, Lee Sedol, world champion and a competitor of AlphaGo, reportedly improved their playing style after examining the moves AlphaGo found during their fifth match~\citep{me19}.
Thus, it seems realistic to assume stakeholders will also find valuable insights when exploring the search history of an AutonoML system, e.g.~novel modelling approaches, innovative architectures, varied resource provisioning strategies, and ML pipelines that are well suited to particular kinds of problems~\citep{xiwu21}.
The integration of explainability mechanisms discussed in Section~\ref{explanation_type} will presumably further enhance this gleaning of insight.

\subsection{What Lies Beyond the Constraints}
\label{beyond_constraints}

Thus far, we have envisioned a state of AutoML technology where a machine can autonomously solve a tightly defined/constrained ML problem.
This scenario is equivalent to fully automating every stage of an ML workflow from data engineering onwards.
However, it is unlikely that the endeavour will end there.
Many real-world problems are fluid and cannot be neatly constrained.
Alternatively, even a firm singular high-level agenda may translate into multiple ML tasks.
For instance, an ML application dedicated to optimising an investment portfolio may aim to predict the best neighbourhoods for property purchases one day and forecast energy market trends the next.
In both cases, one could try to set up multiple ML applications for sub-problems or problem variants across different contexts.
However, not only is this arrangement highly redundant, cross-context information becomes difficult to leverage.
For example, active learning may identify that a client is only interested in `green' suburbs, producing knowledge that would influence better recommendations for energy-company investments.
Thus, with respect to ML problems, the grand aim is clear: one AutonoML system to process them all.

Going beyond constrained environments is, of course, no easy feat.
Automated systems can struggle within problem contexts possessing weak structure, consistency, guidance, and rules.
In such semi-structured and unstructured environments~\citep{kova21}, rare/undesirable events and unforeseen changes in goals may substantially distort, one, a function that ML is attempting to approximate or, two, the space of data that samples this function.
Importantly, we emphasise that the changes must be dramatic and related to context; simpler forms of data/concept drift are presumably handled by the adaptation mechanisms of a fully automated AutonoML system.
In essence, challenges only arise when knowledge acquired via proficient search/learning on supplied datasets cannot address emerging situations.
That is not to say that all of these challenges require complex countermeasures in modern ML.
For example, if a visual model distinguishing between fish and coral starts to encounter something challenging to classify, the simplest method is to attract the attention of humans and have them introduce a new label, e.g.~oceanic plastics.
Unsurprisingly, many mechanisms have been proposed to automate such a process, covering novel data detection~\citep{jugh19}, automated labelling~\citep{saja21}, and the growth of models to accommodate unexpected forms of new data~\citep{gaba00, khru21}.
However, no mechanism currently exists for managing the entire spectrum of situations in semi-structured or unstructured environments where a domain and inputs are inconsistent over time.
Even ML adaptation in structured environments is currently onerous, with humans often involved in reviewing performance degradation, assessing the effectiveness of automated adaptation mechanisms, and confirming a model update if necessary.
So, adapting to changes in unstructured environments can only be more labour-intensive, time-consuming, and costly.

Herein lies a fundamental weakness of traditional ML.
Inductive learning requires a \textit{lot} of data with repeated presentation of concepts in order to develop a decent model.
The era of big data has undoubtedly made this feasible for many ML applications, but the process can still be intensely time-consuming and energy-intensive, even on modern computational hardware.
Moreover, it can be tough to collect sufficient data for many other ML applications, even with synthetic support by a generative model, and this challenge amplifies for environments where concepts can be fluid, e.g.~those that are complex, messy, semi-structured or unstructured.
This state of affairs is unfortunate as, given sufficient computing resources and data, automated search and learning can easily outperform models constructed with expert knowledge~\citep{thgr21}.
This sentiment recurs in the literature, and the superiority of automated search/learning over expert knowledge-based systems is exemplified by research into adaptive soft sensors for well-defined industry process problems~\citep{kaga09, kaga10, kaga11, sabu16b}.
At the same time, the immense costs are also well-evident, particularly when considering state-of-the-art DL approaches used for non-tabular input data, e.g.~images, sounds and text.
Deep neural networks have proved greatly impressive by unlocking new applications and pushing the known limits of predictive accuracy.
However, they can be substantially less computationally efficient than knowledge-based learning systems; a training time of weeks and even months is not unusual.
Consider the celebrated NLP model, GPT-3, which cost about \$4 million to train~\citep{thgr21}.
Although a flaw was retrospectively identified in the preparation of training/testing data, i.e.~concerning overlaps, the developers decided against retraining the model due to its high cost~\citep{brma20}.
This hindrance to fundamental progress in ML can be found across the board.
In medicine, an AutoML algorithm supporting the construction of an automated diagnostic system needs to search through as many variables as possible to find out which features are important, requiring a large amount of data and inducing high computational costs.
In contrast, models trained with expert insight, e.g.~with radiologists or oncologists identifying essential features based on their experience, can be fashioned much faster and, if the domain knowledge is accurate, reach a high level of performance very efficiently~\citep{thgr21}.

As of the early 2020s, the debate around the future of ML and broader AI continues.
The pivotal successes of DL have seemingly become less frequent and more incremental.
Without any revolutionary acceleration of computational hardware on the visible horizon, e.g.~photonic or quantum computing, some argue that data-driven learning should progress towards knowledge-based learning.
However, there are also good reasons why the term `expert system' faded from the data-science lexicon in the 1990s.
There are numerous examples where expert insights failed to deliver better solutions than data-driven learning, e.g.~in toxicity prediction~\citep{buga10a} and biomarker-based water-pollution prediction~\citep{buga10b}.
The human cognitive biases discussed in Section~\ref{biases_impacts} can also have much more negative impacts on knowledge-based systems over data-driven learning, e.g.~via the misidentification of relevant variables and inclusion of faulty rules.
Thus, modern proponents of knowledge-based learning rarely hark back to the extreme of traditional expert systems; the suggestion is that the best of both worlds should be fused.
For instance, perhaps a mechanism based on human cognition/reasoning may be what enables the automation of context understanding, supporting the optimisation of ML pipelines for fluid ML problems with a flexible leveraging of domain expertise.
Indeed, maybe this is the key to strengthening the adaptability of an AutonoML system and enhancing its general applicability.
Whatever the case, whether by a purist form of induction or by a fusion with knowledge-based approaches, AutonoML research will eventually grapple with learning in open-world environments.
The question arises: will this force HCI to evolve even further?

\section{Interacting with AutoML Systems: Open-ended Environments}\label{roles_inter_open_env}

Research along traditional lines of ML seems all but destined to advance AutoML until the technology can solve ML problems to a high degree of performance without human involvement.
This was the focus of Section~\ref{roles_inter_cmpl_constrained_env}.
However, it is much less clear how to take AutonoML further.
Data-driven AutoML systems that excel at searching in constrained environments are unlikely to handle moving between dramatically different domains and tasks. 
Extrapolating present-day approaches, these systems will struggle to extract learned knowledge from one task and apply it to another.
In contrast, humans are very proficient at acquiring new knowledge from minimal data, abstracting and generalising with typical ease.
A major long-term challenge then is translating capabilities typical of human reasoning to automated learning systems~\citep{liaj21}, e.g.~generalisation, an understanding of normativity, the detection/use of causal factors, and so on.
Success in this endeavour may bring AutonoML more in line with a genuine assistant than a tool, and HCI may start to be framed more in terms of collaboration over mere operation.

\begin{table}[!ht]
	\centering
	\caption{The prospective evolution of automation in ML.}
	\label{diff_sections}
	\begin{footnotesize}
		\begin{tabular}{L{3.2cm}C{2.2cm}C{2.5cm}C{2.3cm}}
			\toprule
			\multirow{2}{*}{\makecell[l]{Stage in \\ ML workflow}} & \multicolumn{3}{c}{Need for human involvement} \\\cmidrule(lr){2-4}
			& \makecell[c]{AutoML: \\ Current practices \\ (Section \ref{role_inter_current_automl})} & \makecell[c]{AutoML: \\ Constrained but \\ fully automated \\ (Section \ref{roles_inter_cmpl_constrained_env})} &  \makecell[c]{AutoML: \\ Open-ended\\environments \\ (Section \ref{roles_inter_open_env})} \\ \midrule
			\makecell[l]{Problem formulation \\ \& context understanding} & Yes & Yes & Partly \\ \midrule
			Data engineering & Yes & No & No \\ \midrule
			Model development & Yes & No & No \\ \midrule
			Deployment & Yes & No & No \\ \midrule
			\makecell[l]{Monitoring \\ and maintenance} & Yes & No & No \\ \bottomrule
		\end{tabular}
	\end{footnotesize}
\end{table}

Crucially, as displayed in Table~\ref{diff_sections}, this advance in AutonoML comes primarily down to one section of the ML workflow.
More specifically, leaving problem formulation in the domain of humans, we consider the challenge of automating the second sub-phase: context understanding.
This aim is a lofty goal, but plenty of research/debate exists around this topic in the broader literature.
We merely recontextualise these discussions within the field of AutoML.
Progress in this area would allow AutonoML to understand problems better, consider them from different perspectives, and require far fewer human interventions in complex open-ended environments.
Systems would be particularly suited to managing clusters of similar tasks, but they would also be capable of adapting to novel unseen contexts, transferring knowledge where relevant.
As an example, an AutonoML system for a lane-keeping driving task might generalise its learned knowledge and adapt to various encounters beyond what it was trained for~\citep{leha20}, e.g.~in terms of road profiles, terrain, and lighting conditions.
Of course, this theoretical discussion will remain carefully grounded for AutoML practitioners and current perspectives, striving to suggest what forms of HCI to prepare for when designing the next generation of AutoML.
We will avoid delving deeper into general-purpose learning systems, which verges on artificial general intelligence (AGI).

So, to address the topic of AutoML and open-world learning, Section~\ref{open_ended_learning} will first explore what makes open-ended environments so challenging for ML.
A possible pathway to improving AutonoML will be discussed in Section~\ref{possible_path}: the incorporation of human-style reasoning.
Current thoughts on the plausibility of this future will be reviewed in Section~\ref{plausibility_debate}.
We will then assume the advance to automated context understanding is inevitable, proceeding in Section~\ref{human_roles_open} to consider how HCI between stakeholders and AutonoML systems is likely to evolve.
Finally, Section~\ref{optimisation_interaction} will examine what scope there is to optimise interactions for this novel state of technology.

\subsection{The Challenges of Learning in an Open World} \label{open_ended_learning}

Most popular ML algorithms in the present day operate with a closed-world assumption, where models are generated from a given training set without considering historically acquired knowledge or any other contextual insights~\citep{chli18}.
Therefore, ML algorithms need a large amount of training data to effectively learn and optimise predictive/prescriptive performance.
Frequently, for supervised learning, the training data is considered a representative sample of what will be encountered during testing and production.
Thus each testing/production sample is typically only classified into classes learned from training sets~\citep{fewa16}.
Admittedly, some methods may be sophisticated enough to construct new classes for poorly classifiable data.
However, the real world is open-ended, where no data perfectly reflects all the dynamic changes and events that may occur within a data environment.

In effect, there are unlimited possibilities and no fixed rules to anticipate all requisite information needed to maintain predictive/prescriptive performance in any given situation.
Accordingly, it is virtually impossible to yield a perfect training set for every function that an ML model needs to learn.
The challenge then is deciding how best to loosen the constraints of well-defined problems so that AutonoML systems can work effectively in real-world scenarios with the aim of continuous lifelong learning.
Operating conditions and contexts usually change over time within an open-ended world, so the key to progress seems to rely on improving the first phase of the ML workflow: problem formulation and context understanding.
Ideally, the goal is to equip AutonoML systems with advanced mechanisms capable of understanding operating contexts better, enabling more autonomous operations supported by non-technical collaboration with human operators.
This gradual relaxation of closed-world constraints is a common agenda across the broader field of AI.
Many ongoing data-driven research efforts aim to acquire, generalise and transfer learned knowledge across different input spaces of the same task or related tasks within open-world scenarios.

One thread of research considers the situation where a single set of training data does not cover every angle of an ML problem.
Specifically, multi-view learning samples a data environment from multiple perspectives, in which each view is represented by a distinct set of features, thus enhancing the performance of an ML model.
For instance, training data for an ML problem processing a video source can be formed from two different perspectives, i.e.~the image signal and the audio signal.
Both views are non-overlapping, yet the signals will be correlated in a highly complex manner, and both will need analysis to thoroughly understand the meaning within the data source.
This research direction has many motivations, but one of them is enabling an automated system to examine and leverage every angle of provided data without changing the problem definition.
Numerous approaches for multi-view learning are listed elsewhere~\citep{su12}.

Another thread of research investigates continuously changing environments, although still mainly within the confines of a previously defined ML task.
The goal is to enable an AutonoML system interacting with its environment to learn from its experiences, thus accumulating, organising and transferring knowledge over long periods.
This long-term form of continuous learning, accommodating new knowledge while retaining previously learned insights, is also known as lifelong learning.
Many approaches tackling lifelong learning draw inspiration from biological systems, and they have been reviewed elsewhere~\citep{pake19}.
We re-emphasise that, to a large degree, lifelong learning is already expected to be a characteristic of constrained but fully automated AutonoML; see Section~\ref{roles_inter_cmpl_constrained_env}.
It is the \textit{extension} of adaptive mechanisms to the phase of context understanding -- consider the dotted looping arrow from the monitoring and maintenance stage in Fig.~\ref{ml_workflow} -- that defines the lifelong learning required for open-ended environments.

As Section~\ref{search_strategy} indicated, an ideal approach to information transfer between contexts is the use of meta-knowledge, where knowledge about how learning was achieved for one ML problem should prove educational in enhancing the performance of ML within a similar setting/domain.
This process of meta-learning, along with its achievements and applications, has frequently been reviewed~\citep{va11, lebu15, albu20}.
In the context of AutoML, meta-learning can be used to recommend good values for system variables, ranging from low-level HPO search-space constraints~\citep{ngke21} to high-level mechanisms, e.g.~which optimiser to use.
A closely related notion is transfer learning~\citep{paya09, albu19, zhqi20}, which typically involves copying and pasting an entire ML pipeline or DL network with a solid performance on one ML/DL problem into the solution space of another.
Most of the transferred pipeline/network is often kept invariant to subsequent retraining, with only the interfacing components/layers being tunable.
In theory, this approach is reasonable, as, for example, it seems that initial layers for feature identification/processing in a deep neural network should not change dramatically whether a model is identifying animals or objects.
Presumably, the last layers correlating feature combinations with concepts are what vary.
However, both meta-learning and transfer learning suffer the same issue.
Their effectiveness depends on the number of commonalities between domains.
For human learning, this makes intuitive sense; the skills acquired from swimming do not readily apply to playing a guitar.
The challenge lies here: the degree of similarity between problems/contexts remains massively difficult to define robustly.
Thus, a softer form of transfer learning exists, called domain adaptation~\citep{sush15}, which typically does not have to care about varying feature spaces, only distribution/domain shifts.
Domain adaptation is not open-world learning in the way this section defines it, but it is still an incremental loosening of problem constraints, the successes of which may feed into broader progress.

By now, it is evident that there is plenty of research activity in broadening the adaptive applicability of ML and, by proxy, AutoML.
However, these threads continue to struggle with training/adapting ML models for real open-world problems despite many remarkable achievements.
In a recent IEEE Spectrum interview, distinguished ML researcher Michael Jordan stated that, ``for the foreseeable future, computers will not be able to match humans in their ability to reason abstractly about real-world situations''~\citep{pr21}.
Unlike ML algorithms, humans can learn effectively and adaptively in open-ended environments from only a few presentations of concept~\citep{chli18}.
Naturally, there is no consensus as to how this is done, e.g.~whether this capability is unlocked by the processing power of the brain, the advanced mechanics/architectures of biological neurons and their networks, or simply the accumulation and effective leveraging of knowledge over a lifetime of learning.
Perhaps it is a mixture of all three.
Nonetheless, phenomenologically, these abilities appear to manifest as the use of semantic representations and high-level reasoning.
Thus, if ML/DL algorithms do not have the complexity yet for such high-level phenomena to arise emergently, then perhaps the shortcut of engineering these qualities into an AutonoML system will make do.
Such a mechanism would make AutonoML better suited for open-ended environments by enabling the accumulation and organisation of historically learned knowledge, employing it via the form of reasoning on new ML tasks/problems~\citep{basa17}.

The core requirement of open-world learning systems is the ability to identify and learn unknowns in order to become more knowledgeable.
In particular, AutonoML systems may need to construct models of the world to support prediction and planning, explanation, abstraction, and the flexible generalisation of knowledge from observations~\citep{laul17}.
This capability goes beyond simple pattern recognition based on many previously determined tasks.
For instance, toddlers often observe the world and generate a mental model regarding how it works, then they take action and adopt the outcomes of that action to refine their mental model~\citep{st21}.
These steps are repeated until they sufficiently interpret the world.
Indeed, through continuous training on non-stop streams of data encountered every day, supported by ongoing background optimisation during sleep time, toddlers accumulate rich knowledge to help them better perceive and understand various contexts in the real world.

Immediately, this hints that the very nature of HCI within AutonoML might need to change.
Although Section~\ref{roles_inter_cmpl_constrained_env} suggested meta-knowledge bases could be required for closed-world AutonoML, these would likely be developed/expanded \textit{reactively} while a system is encountering/solving a problem, with maybe a few one-off sessions of offline processing.
On the other hand, a model of real-world contexts and their relationships is at least an order of magnitude more complicated to generate.
In essence, AutonoML systems may need to \textit{proactively} develop these context models, in which case they are unlikely ever to be switched off.
This paradigm is vastly different from what Section~\ref{role_inter_current_automl} describes as the most common present use of AutoML, i.e.~a tool to be picked up when an ML application requires it.
In fact, given how complex and expensive it will likely be to generate a context model for the real world, it may be that one will be shared between multiple instances of an AutonoML system, with each contributing to it in their off-time.
Accordingly, the systems are likely to be `always on', which has implications for the \textit{constancy} of HCI; see Section~\ref{human_roles_open}.

Now, it is argued that one of the strengths of human-level intelligence that suits it for open-world learning is the ability to generate information from raw data and systematically organise it into abstract high-level concepts~\citep{ga20}.
This knowledge synthesis occurs at a different level to the typical curve-fitting that ML applies to raw data.
Substantial portions of the data science community advocate this reasoning-based form of learning as a way forward for more capable AI; this includes AutoML.
Additionally, in an open-ended environment, AutonoML must be equipped to make decisions with incomplete and uncertain information.
Such scenarios have previously been called commonsense informatic situations~\citep{mc07}.
Thus, although meta-learning and transfer learning based purely on data-driven approaches have shown promise for open-ended problems, the challenges of learning in an open world seem to require more sophisticated strategies leaning towards human-based reasoning.
We cannot state with certainty whether this is the best pathway for advancing AutoML, but the intensity of recent research activity aimed at integrating high-level reasoning mechanisms into data-driven learning paradigms warrants review, if only to characterise/contextualise the advanced systems that HCI may have to accommodate.

\subsection{One Possible Future: AutoML and Reasoning}
\label{possible_path}

The path to automating context understanding and relaxing the restrictions on open-world learning may require AutoML to eventually integrate some form of mechanism for reasoning.
Three components of human intelligence are particularly relevant to this endeavour: (1) the synthesis of conceptual knowledge from data, (2) reasoning based on this conceptual knowledge, and (3) cognitive mechanisms that revise and adjust the acquired conceptual knowledge.
So, to review research efforts/opinions in this area, Section~\ref{human_reasoning} begins by examining how humans reason and make decisions.
As detailed by Section~\ref{paradigm_shift}, this potentially motivates a shift in how ML is approached, evolving from a data-driven focus to one that is knowledge-driven.
Various types of reasoning will then be elaborated by Section~\ref{reasoning}, before Section~\ref{cognitive_models} concludes with a brief overview of how artificial models of cognition may be generated.

\subsubsection{How Humans Make Decisions}
\label{human_reasoning}

Reasoning is a complex concept, and neuroscientists continue to debate many facets of how the human brain engages in cognition.
Accordingly, this section is not intended to be a comprehensive review of current literature on the topic.
However, there is a rough consensus that human cognition operates on multiple levels, although the jury is out on whether these levels are discrete modes of operation or merely distinct phenomena along a continuous spectrum of neural processing.
For the purpose of contemplating open-world AutonoML, the low-level details do not appear to matter.
Thus, we focus here on the popular dual-process model of cognition, also known as the Kahneman theory of thinking fast and slow~\citep{ka11}.
According to this theory, human reasoning and decision-making are enacted by collaboration between two `systems' of thinking.
System 1 involves thinking that is fast, instinctual, associative, intuitive, and low-effort.
System 2 involves thinking that is slow, deliberate, logical, rational, and reasoned.
Several discussions of AI have been grounded in this theory~\citep{rolo19}.

In greater detail, System 1 is said to involve humans using experience, biases and evolutionary heuristics with unconscious cognitive shortcuts to generate quick responses~\citep{sl14}.
Within the context of AI, this form of thought is often associated with the pattern identification and curve fitting of standard ML.
It is not that \textit{generating} the approximating model is fast; the speed usually refers to ML inference, where a novel instance of data can be quickly converted into a prediction/prescription by the approximating function.
A human example of System 1 processing is rapid object identification, usually employing inductive reasoning and the representativeness heuristic to leverage similarity with previously seen objects~\citep{waya19}.
Naturally, the priority of speed over robustness for System 1 means that determinations tend to have weaker validity over other forms of deductive reasoning.
The results of System 1 can also lack explainability, being based on heuristics, unconscious biases, memories, pattern matching, and other shortcuts.
Pareidolia is an obvious example of errors in associative thought.
Nonetheless, despite a tendency to be loose with accuracy and precision, System 1 remains essential for efficiently updating and maintaining personal models of the world~\citep{ka11}, constantly integrating a stream of incoming information into background beliefs~\citep{sobe19} so that effective responses can be formulated for many daily stimuli.
Admittedly, unlike most forms of traditional ML, it is argued that System 1 in humans is also capable of conducting basic causal inferences to interpolate/extrapolate in the case of incomplete knowledge~\citep{bofa20}.
This causal reasoning applied by System 1 thinking is kept at the level of data, while more complex assessments of causality are thought to occur at higher levels.

In Kahneman theory, this leads to System 2.
Theoretically, when facing a novel problem that is too complex to be solved by System 1, such as complicated mathematical calculations, System 2 is triggered and takes over.
It accesses extra cognitive resources and sophisticated logical reasoning mechanisms to form thoughts in a series of steps~\citep{ka11}.
During this procedure of extending explicit knowledge and beliefs, System 2 thinking is slow, step-wise, deliberate, analytical, rule-based, controlled, and resource-heavy.
In the case of humans, complex reasoning processes cannot be utilised without prior knowledge acquisition from many different domains.
This expertise is required to systematically capture/use concepts that are semantic, logical, or mathematical~\citep{waya19}.
It is also needed to manipulate symbolic and abstract rules~\citep{grgi18}.
Because inference here can be much more involved than with inductive approaches, System 2 is argued to subdivide large tasks into multiple smaller parts.
High-level cognition also appears to prioritise long-term memory, while instinctive low-level cognition typically leverages short-term memory~\citep{ka11}.

As with humans, any AutonoML frameworks that incorporate the principles of Kahneman theory are expected to have distinct levels of cognition acting in collaboration.
Proposals in the field of AI vary, but System 2 is often envisaged as being supported by the fast-thinking System 1.
For instance, given the vast solution spaces that System 2 can encounter, heuristics provided by System 1 are well-placed to provide constraints and focus only on the most promising subspaces~\citep{bofa20}.
In the opposite direction, once complex concepts are sufficiently established and refined by System 2, some of this knowledge may be translated/simplified into a form that System 1 can use.
This development is often associated with the notion of once-complex tasks becoming `second nature', i.e.~no longer requiring concentrated thought.
Examples of this are many, e.g.~developing motor control for bike riding or learning to read a natural language.

The overall hypothesis here is that it takes sophisticated thinking/reasoning mechanisms, such as the collaboration of two distinct levels, to provide humans with a competitive advantage in cognition and decision-making over, for instance, animals.
Indeed, animals may have a significantly less developed form of System 2 thinking, aligning with a predisposition for instinctive behaviour.
In contrast, humans can reason at multiple levels of abstraction, adapt efficiently to dramatically new environments, understand wide-ranging contexts, translate/generalise knowledge-driven skills from one problem to another, etc.
Accordingly, to enhance the capabilities of AutonoML, it appears desirable to engineer frameworks with these multi-level principles of cognition.
Of course, computer architectures are not identical to biological brains, so future AutonoML engineers may have the flexibility to improve upon the designs of nature.
For instance, in humans, Systems 1 and 2 are often categorised by local and global forms of processing, respectively, i.e.~in terms of which parts of a brain are involved.
A natural question to ask is as follows: would an integrated AutonoML framework~\citep{kemu20} mimic these distinctions?
Additionally, it is suggested that the cheap form of System 1 processing is done in parallel within a human brain, while System 2 requires plenty of attention/concentration, operating serially.
Sure enough, conventional wisdom is that few humans are effective multi-taskers when it comes to deliberative thought.
Nevertheless, perhaps an AutonoML framework does not need to be similarly restricted, and computers may support the parallelisation of high-level cognition.

Importantly, we note that, while multiple levels of cognition do advance human capabilities, cognitive biases at all levels can impact solutions, forcing sub-optimal outcomes and incorrect decisions~\citep{waya19}.
Typical heuristic biases and their negative impacts have previously been highlighted when studying the medical decision-making process of clinicians~\citep{pr19}.
Granted, as discussed in Section~\ref{mitigate_bias_collab}, these same factors may be mitigated in autonomous systems by the integration of explanatory mechanisms and collaborative interactions with stakeholders.
Nonetheless, it is probably a mistake for AutonoML to directly mimic how humans make decisions.
More realistically, biological processes should be studied as an inspiration when designing the behaviour of AutonoML systems in the future.

\subsubsection{The Paradigm Shift from Data-driven to Knowledge-driven}
\label{paradigm_shift}

As hinted by Section~\ref{open_ended_learning}, the reason why humans are so effective at open-world learning may partially be due to one fact: humans virtually never start learning from scratch.
This statement may seem patently untrue, as humans frequently encounter new situations for the first time.
However, unlike a modern AutoML system, they are hopefully not rebooted prior to each new encounter.
The above assertion does not even refer to the persistence of a traditional meta-learning module in the human brain, e.g.~one that aids learning how to ride a mountain bike based on prior experience with a road bike.
No, long-term experience has equipped humans with much more abstract general knowledge~\citep{la17}, e.g.~numbers, time, space, physics, psychology, etc.
These concepts represent pervasive principles inherent to real-world environments, e.g.~the mental/behavioural states of entities, the observed physical laws of reality, and even more abstract properties like compositionality and causality~\citep{boba17}.
Such abstractions form links between disparate environments, meaning that even an apparently novel problem context can be, to some degree, understood by a human.
For instance, most humans recognise that jumping off a skyscraper without a parachute is typically a bad idea.
They usually do not have to experience/observe repeat presentations of this concept to reach enlightenment.
Thus, the integration of fundamental abstract concepts into ML systems, such as notions from physics and psychology, may be essential for elevating AutonoML to open-world learning~\citep{la17}.
To date, debate continues around how best to achieve this, but a notable portion of the data science community fears that the modern standard of approaching every ML application from scratch and relying only on data-driven approaches is likely to be ``an alarming mistake''~\citep{mada19}.

A recent study argues that most modern forms of ML, especially DL, are poor at generalising knowledge beyond the space of training data, meaning that they cannot be trusted in open-ended environments~\citep{ma20}.
Consequently, the study motivates the development of an autonomous system that can regularly acquire, represent, and modify abstract knowledge.
The goal is for the system to use this learned knowledge in deducing, constructing and updating internal cognitive models of an external working environment.
To be fair, many present-day ML/DL algorithms effectively generate knowledge within a constrained environment.
However, they fail to capture many aspects, e.g.~the way that the physical world works, how objects interact, and how causal relationships manifest in dynamically changing environments~\citep{zhwu16}.
Several research efforts have honed in on this last point specifically, attempting to statistically extract causal relations from changes in data distributions~\citep{bedera19} and automatically learn modular structures representing the dynamics of a working environment~\citep{gola19}.
The concept of causality is essential to human reasoning, often impacting the way new concepts are generated and learned~\citep{re03}.
In addition, causal knowledge also enables human beings to specify what is critical and infer what is unknown~\citep{prah07}.
For this reason, equipping AutonoML systems with mechanisms for the extraction of causal relations from data may be an initial step towards open-world learning.

Now, assuming that the AutonoML agenda does begin to extend from data-driven approaches to knowledge-driven learning, a fundamental question is as follows: how should abstract concepts at a higher level than data be represented?
There is an obvious answer: with symbols.
Indeed, animal cognition is argued to use symbolic representation for the storage and retrieval of information~\citep{gaki11}.
This theory is particularly well-supported for humans, with plenty to suggest that symbolic reasoning is vital to deliberative cognition~\citep{ma20}; consider how often the communication of thought processes is framed in deductive logic operating on symbolic concepts.
Therefore, equipping AutonoML systems with a deeper understanding of an open-ended working environment seems likely to rely on integrating human-like mechanisms that acquire, represent, organise and manipulate abstract symbolic knowledge.
In effect, open-world AutonoML may need to embrace a hybrid learning paradigm that can discover, extract and generalise abstract knowledge from large and noisy datasets while also reasoning with this learned knowledge to better tackle new ML tasks.
It is certainly a recurring opinion, in both the past and present, that AI architectures could potentially benefit from the best of both worlds~\citep{ne80, ma19}, i.e.~the powerful large-scale learning ability of data-driven methods and the generalisable applicability of abstract knowledge in symbolic format.

Another potential strength of knowledge-driven learning is an improved ability to handle uncertainty and missing values.
Crucially, in real-world contexts, data is usually dependent on imperfect observation, thus being subject to noise and incompleteness.
Admittedly, the typical point of an ML algorithm is already to fit and interpolate/extrapolate, leveraging model-specific choices for how inductive reasoning is applied.
However, in an open-ended environment, it is much easier to encounter a novel observation that has no comparable instance of data on which to anchor inductive reasoning.
Thus, an ideal open-world AutonoML system should lean on deductive reasoning to fill in the gaps, making assumptions, inferences and deliberation based on prior conceptual knowledge, updating and revising where need be.
Relying on pure search, no matter how proficient or supported by association-based meta-learning, is unlikely to be efficient or effective.

As an example, consider an ML application tasked with predicting the socioeconomic classes of people -- each dataset is specific to a nationality -- based on their yearly expenditures across a range of consumer goods.
Searching through an infinite space of possible data transformations to maximise classification performance seems infeasible.
Selecting data transformations based on similarities between country-specific datasets may be better, but it faces the usual challenges of traditional meta-learning described in Section~\ref{open_ended_learning}.
In contrast, linking a country to a knowledge-base of economic data may present a more expansive yet well-reasoned pool of data transformations, e.g.~normalising expenditures based on purchasing power parity and domestic price indices.
Now, consider that a new dataset differs not in nationality but feature space, e.g.~income streams are listed instead of spending.
Traditional AutoML would likely need to search for an ML pipeline from scratch, while meta-learning would have no prior experience to recommend data transformations.
However, conceptual knowledge, such as linking income/expenditure feature labels to appropriate economic data, may preserve the performance of the original ML pipeline; an AutonoML system may simply attach a new transformation related to country-specific household debt levels.

All of these knowledge-driven improvements boil down to relaxing the need for HCI in the first stage, second sub-phase, of the ML workflow in Fig.~\ref{ml_workflow}.
By working towards automated context understanding, humans no longer need to be as rigid and constrained when defining a problem statement.
Concurrently, AutonoML systems and the ML solutions they support become much more robust and capable.
For instance, it is worth noting that, although the famed GPT-3 language model~\citep{brma20} achieves impressive outcomes on NLP tasks, many limitations have been attributed to it, e.g.~ingrained biases, a propensity for cherry-picking, a lack of semantic coherence in outputs, and unreliability~\citep{dale21}.
It is far from a system with an open-world level of intelligence because it is primarily based on textual statistics and probabilistic matching.
It has no discernible understanding of underlying concepts, causal relationships, and other facets with which human cognition could easily grapple.
We stress the word `discernible' here, given how contentious the phenomenon of understanding is; consensus theories of human cognition attribute both the fast and slow thinking discussed in Section~\ref{human_reasoning} to the same fundamental concept of a biological neuron.
Nonetheless, many argue that appreciating the statistical properties of a concept manifesting is not equivalent to understanding that concept~\citep{ma20}.
To acquire true understanding, or at least mimic the functional phenomenon, AutonoML seemingly needs to operate with prior abstractions of concept under uncertain conditions.
Naturally, if deductive inference is based on previous truths, a question then arises: what kinds of prior knowledge should be made available to an autonomous learning system?
Data scientists have argued that these should encode principles about time, space, causality~\citep{mada19}, compositionality, intuitive physics, and psychology relating to both individual agents and groups~\citep{la17}.
Whatever the case, this `mental model' of the universe will need to be curated/integrated carefully, given the significant potential for computational bloat.

A final comment is that one of the big HCI issues of modern AutoML, i.e.~the topic of trust that was discussed in Section~\ref{outcome_trust}, will remain relevant for knowledge-driven learning.
Outcomes need to be interpretable/explainable, especially when the risk-managing constraints of a closed world fall away; see Section~\ref{human_roles_open}.
Fortunately, symbolic logic and similar forms of reasoning are already typically perceived as more transparent and believable than inductive extrapolation due to various potential properties, e.g.~logical consequence and deductive closures.
Indeed, there are many human-centred reasons why knowledge extraction has been proposed to enhance communication between users and an autonomous learning system~\citep{ga20}.
Improved explainability is just another potential bonus of integrating knowledge-driven approaches into AutonoML systems.

\subsubsection{Types of Reasoning}
\label{reasoning}

Even though the implementation of an open-world AutonoML system is unclear, in terms of whether mechanisms should be discrete and prescribed~\citep{kemu20}, a functional dichotomy is becoming apparent.
Traditional ML is associated with System 1, fast inference, inductive principles, and data-driven learning.
Automated context understanding is associated with System 2, slow inference, deductive principles, and knowledge-driven learning.
Traditional ML typically attempts to interpolate/extrapolate approximate probability density functions for an entire domain, processing large amounts of data.
Automated context understanding prospectively attempts to encode concepts from a problem setting and data environment into formal logic, often compacted representationally into symbolic logic.
Each has its own particular strengths and weaknesses.

For an open-ended world, it is infeasible to make a nuanced global fit of a function, so the capabilities of ML are weakened.
Accordingly, an AutonoML system should potentially support local-space approximations with inferences formed from global-space abstractions of knowledge, thus extrapolating beyond supplied data distributions without needing large training samples~\citep{trga16}.
Many other inspirations can be drawn from human intelligence, which is effective/efficient for open-world learning.
Humans frequently reason when considering action strategies, taking into account many relevant actors and events, both transient and continuous~\citep{mc07}.
Of course, the dichotomy discussed above is a simplification, and there are many nuances to reasoning that may be useful in the context of open-world AutonoML.
We briefly mention reasoning of the following types: low-granularity induction, causal deduction, symbolic, common-sense, and `exact' combinatorial.

When discussing induction, as in making broad generalisations from specific observations, it is essential to emphasise that knowledge exists in a spectrum of scope and detail.
Unlike other animals, humans have the known ability to understand and communicate high-level information easily~\citep{ha14}.
This quality is primarily due to the capability of abstracting high-level concepts from raw data.
In addition, human intelligence can make inferences at different levels of abstraction to focus only on specific sets of features, disregarding unimportant information~\citep{zu03}.
Now, while the last few sections have stressed the benefit of using deductive reasoning for conceptual knowledge, in apposition to the inductive reasoning applied by ML at the data level, nothing precludes induction from being applied to the higher level of knowledge as well.
Plenty of research exists intending to extend learning algorithms to various granularity levels, e.g.~as proposed by a granular computing framework~\citep{bape16, khch21}, generalising abstract/causal rules via the process of induction.

On that note, causality is a beneficial concept to have a handle on when forging predictions/prescriptions.
A weakness of traditional ML is that it constructs models inductively based on correlation, which means that ML solutions can fail dramatically under data/concept drift if a feature-label association is coincidental and the underlying causal factor remains unknown.
In contrast, once a causal relationship is determined, it can usually be converted easily into a logical consequence, allowing subsequent inferences to be deductively made.
Of course, algorithms still need to extract rules/knowledge from raw data to reason at a higher level with hierarchical structures of causal insights.
One way to do so is with probabilistic programs.
For instance, a recent study used a Bayesian strategy to convert raw data into causal knowledge, handling noise with probabilistic semantics~\citep{lasa15}.

Once an AutonoML system begins to dabble in high-level deduction, this reasoning will usually be expressed in terms of formal logic.
In fact, at a low level, decision trees are an explicit example of formal logic being applied, typically represented in the form of `if-then' rules.
Here, we remind that, while ML constructs models inductively, applying the model to new queries is technically a deductive process, i.e.~ML training converts specific information to general information, while ML inference works oppositely.
Regardless, combining and aggregating logical theorems at a higher level can result in highly complex structures.
Thus, humans usually employ symbols to encode such concepts for both convenient communication and the atomic basis of extended reasoning~\citep{taug18}.
Unsurprisingly, many studies have focussed on embedding symbolic rules and knowledge into traditionally data-driven ML/DL algorithms so that they may infer and capture compositional/causal structures from data~\citep{ga20}.
One proposal considers adopting a three-level causal hierarchy to make ML algorithms more robust and autonomous~\citep{pe19}, i.e.~managing association, intervention, and counterfactuals.
As mentioned above, most current implementations of ML only leverage purely statistical relationships, so they only support the lowest level of association.
However, it is argued that if symbolic rules can be integrated into models such as neural networks, ML will become capable of supporting inference involving intervention and counterfactuals~\citep{ga20}.
In essence, applying symbolic manipulations seems to be very convenient for, if not a prerequisite to, understanding cause-effect connections.
Naturally, given a dichotomous perspective of fast and slow thinking, one possible approach is to allow data-driven learning algorithms to interact with symbolic extraction/reasoning mechanisms in some hybrid manner, e.g.~a neural-symbolic framework~\citep{beku15}.

Finally, the topic and lexicon of cognition are vast, so many other variant terms are associated with reasoning, each with their own nuanced vision of what ML should support, even though aspects overlap with the concepts already discussed.
For instance, any artificial system competitive with human intelligence is arguably required to operate efficiently in common-sense informatic situations~\citep{mc07}.
This notion usually refers to decision-making when information is uncertain or incomplete.
The intention here is that AutonoML systems will need the ability to jump to conclusions based on assumptions, using approximate concepts/theories and, ideally, maintaining a record of the confidence levels for each decision.
Such conclusions may then be updated and adjusted during learning processes when new instances are observed or novel knowledge is extracted~\citep{ga20}.
Practical principles to support common-sense reasoning include Bayesian inference, non-monotonic logic, and fuzzy logic.
Oppositely, when an ML application is rich with knowledge and observations are both complete and certain, `exact' combinatorial reasoning, i.e.~based on counting, may find its own purpose.
Given that combinatorial reasoning could potentially correct conclusions generated by approximate reasoning, it is clear that an AutonoML system would benefit from implementing various forms of logic, working in concert to cover a broad spectrum of possible ML problems/contexts.
Admittedly, high-level AutoML operations are already very resource-expensive in the modern era, so the true challenge of employing cognitive mechanisms may not reveal itself until the engineering/implementation phase. 

\subsubsection{Cognitive Models of Learning}
\label{cognitive_models}

Given that the notion of incorporating high-level cognition in AutonoML is presently speculative, although well reasoned, we do not dwell on matters of implementation and engineering.
However, it is worth noting that, in the broader AI literature, plenty of relevant thought is being paid to artificial cognitive models of learning and how they should function.
For instance, one recurring question is how such systems can simultaneously enrich and leverage high-level knowledge.
Human cognition can again act as an inspiration for this process, which accumulates precise/complete knowledge through continual operation.
Specifically, it obtains, encodes and revises knowledge based on experience or instruction, inferring and deducing new knowledge by the combination of existing concepts and rules~\citep{la12}.
One example of a computational system that simulates this mental procedure is Cascade~\citep{va99}, using rules to represent knowledge and incorporating an impasse-repair-reflect cycle.
During the learning process, Cascade operates normally until it hits an impasse, i.e.~where existing knowledge is insufficient to solve a problem.
This event then triggers a repair phase, where current rules are modified or new ones are constructed from the combination of previous rules, with common-sense knowledge potentially involved.
Finally, the system reflects on the validity of the suggested rules via a self-explanation form of reasoning.

Beyond operating procedure, AutonoML engineers will also need to decide what an open-world system will need to be shipped with.
Indeed, while Section~\ref{reasoning} discussed diverse forms of reasoning with knowledge, earlier sections highlighted arguments that a cognitive model should already be equipped with baseline information about sets of entities and relevant properties, including time and events~\citep{ma20}.
Such proposals suggest that it should not be overly challenging to construct internal cognitive models automatically, based on foundations that include hybrid model architectures, prior knowledge, and reasoning techniques.
Finding effective algorithms to surmise causal relationships from data may also be particularly valuable, providing strong priors for learning/reasoning as well as the generalisation of new knowledge~\citep{la17}.
For the time being, most ML/AutoML research and development have overlooked these matters, focussing on what is perhaps the easier challenge: end-to-end learning with convenient assumptions, e.g.~the availability of large datasets.
This deliberate neglect of building cognitive models within learning systems may stymy progress in the long term.
For example, NLP models based on Transformer architectures~\citep{vash17} can achieve state-of-the-art results, but their reasoning ability is poor.
Without cognitive mechanisms and a conceptual model of the world in which the language is based, such applications lack interpretability and genuine understanding.
The text generation capability of such NLP systems, while impressive, quickly ends up being identified as human mimicry; the lack of context understanding means such approaches will easily struggle with the Turing test.
Thus, it seems inevitable that AutonoML research will have to tread down the path of integrating cognition mechanisms, or equivalent, if it intends to unlock the full power of automation on a maximal spectrum of ML applications.

\subsection{A Brief Debate on the Plausibility of This Future}
\label{plausibility_debate}

Any speculation on the future of ML will always be debated, so, before we consider the full implications of open-world AutonoML on HCI, we first review opinions on whether automated context understanding and the integration of high-level cognitive mechanisms are plausible directions for future research.
At an initial glance, such a pathway does not seem overly attractive to most of the modern data science community, although it is unclear whether that is because of a cost-benefit analysis or simply tunnel vision.
Much cutting-edge DL research focusses on automatically building massive pre-trained data-driven models and then adapting their learned knowledge to specific tasks.
This philosophy has even been recently branded as one of `foundation models', discussed primarily in terms of very large neural NLP networks~\citep{bohu21}.
The movement has been motivated by the initial success of the OpenAI GPT-3 model, consisting of approximately 175 billion parameters~\citep{brma20}, as well as Switch Transformers, consisting of nearly 1.6 trillion parameters~\citep{fezo21}.
Certainly, data-driven learning models have accumulated many significant achievements, being both flexible and capable of discovering complex statistical relationships between features in large amounts of data.

However, as expressed by a growing portion of the community, the rapid gains of the DL era have slowed.
For instance, despite the dramatically successful approach of AlphaGo and its successors for relatively constrained problems, it is claimed that the overseeing institute, DeepMind, has not achieved many equivalent breakthroughs for ML applications involving, for example, cancer or clean energy~\citep{fj20}.
To some extent, this is due to ostensibly unpredictable dynamics in real-world data environments, to which induction-based learning is vulnerable.
Other disadvantages include a reliance on large datasets, a heavy consumption of resources, and a lack of explainability and trustworthiness~\citep{hoca20}.
In essence, substantial progress may require insights and paradigms that are novel or, at the very least, new to the modern era of computational hardware.

A fundamental limitation of ML remains that most correlations are not causal, even if the former may indicate the latter~\citep{fj20}.
Indeed, current data-driven learning models work almost entirely in association mode; see Section~\ref{reasoning}.
It is argued then that, for an autonomous learning system to pass the Turing test and exhibit an intellectual level comparable to human intelligence, such a system must have mastery over causal knowledge~\citep{pema18}.
Others go further and claim that these machines fail the Turing test because they do not inhabit the material/social world that human beings live in, thus having no means to understand it~\citep{fj20}.
This assertion is standard, with similar claims that understanding the environment of a human does not rely on knowing the chemical reactions in the brain or `soul' of a person; one must empathise with their circumstances~\citep{drdr00}.
Of course, some of the discourse veers into topics beyond the scope of this monograph.
An AutonoML framework does not necessarily need to pass a Turing test to solve ML problems proficiently.
However, automated context understanding may ultimately rely on interacting with humans and an open-ended world to both acquire contextual knowledge and stay abreast of dynamic changes.

On that note, the importance of dynamics and adaptation to the future of AutoML cannot be overstated.
These topics were discussed in Section~\ref{roles_inter_cmpl_constrained_env} at the data level, e.g.~in terms of data/concept drift that do not break well-defined problem constraints.
However, at a \textit{higher} level, humans can swiftly respond to changes in a working environment, able to identify rare events and novel concepts after only a few presentations within observed data.
The critical factor that discriminates this human capability from current state-of-the-art AutoML is the ability to reason with limited examples, leveraging rich contextual knowledge.
In particular, deliberative decision-making does not typically rely on fine-grained data, instead generalising conceptual notions for subsequent use in reasoning.
For instance, studies suggest that children utilise high-level abstractions to reason about the distributional and dynamical properties of objects, discovering the underlying causes of observed events and using this knowledge to guide solution search when facing new tasks~\citep{tste15}.
As a result, the acquisition of rich domain knowledge and the compilation of conceptual models, combined with subsequent reasoning, is asserted to be a much better form of learning than the pattern recognition employed by current ML algorithms~\citep{mijo16}.
Moreover, adaptation on a higher level seems to be most successful if there is a ready-to-use knowledge base to leverage, e.g.~by building hierarchical priors to support learning-to-learn~\citep{brme10} or constructing abstract models of the world to enable learning new notions with the aid of previous experience and similar concepts~\citep{kepe07}.

Importantly, while an increasing number of researchers have opined for the integration of high-level cognition into smart autonomous systems, some have urged caution about directly imitating human intelligence~\citep{kova21}.
While certain human capabilities are phenomenologically appealing, there are fundamental distinctions between biological brains and computational systems, e.g.~in terms of organisational architecture, connectivity, speed, scalability, adaptability, and energy consumption.
Furthermore, the development of an AutonoML system is arguably limited only by modern constraints of physics/technology and the ingenuity ceiling of engineers.
In contrast, for all of its merits, biological evolution is slow when adapting to the complex hierarchy of modern-era desires that correlate with biological fitness, e.g.~solving societal problems rather than focussing on immediate survival.
For instance, humans did not evolve to process large-dataset ML applications mentally, and there is evidence that biases and cognitive limitations make humans particularly bad at some forms of reasoning~\citep{jo19}.
Presumably, such cognitive processes are not a biological imperative.
So, while human intelligence remains far advanced in open-ended environments, serving as a helpful inspiration, those that argue caution comment that imitating narrowly-defined human skills via collecting data and building DL models does not contribute to tackling major issues in AI.
Instead, it is argued that computer-appropriate solutions should be sought for bringing meaning/reasoning into autonomous systems, integrating causal representation/inference, and developing computationally tractable representations of uncertainty.

Extreme views are likewise moderated when contemplating how far to go with knowledge-driven learning.
Few modern perspectives espouse the belief that AI research should return to the era of expert systems.
While deductive reasoning and conceptual generalisations have clear benefits, the fallacy remains that rules defined by human experts are constrictive and ultimately do not capture a diversity of real-world situations.
In fact, one study suggests that most of the knowledge humans use in daily life, especially that which is acquired experimentally, is tacit~\citep{po12}.
The study supports this claim with an example of riding a bicycle, stating that it is hard for even an expert to teach this via complete explicit instructions.
For instance, countersteering is an act that everyone is forced to intuitively master in order to keep balance, yet the action remains an unintuitive concept in physics to many; casual bicycle riders do not deliberatively think that turning left first requires steering right.
Accordingly, the study concludes that ``we can know more than we can tell'', asserting that skills are a general prerequisite to articulating knowledge.
Such skills seemingly cannot be acquired by merely following rules in a textbook, instead requiring interactions with operational environments and other agents, applying search and learning, and adjusting from trial and error.
It follows that the process of data-driven ML, exemplified by the success of AlphaGo in a restricted environment, is suggestive of acquiring that tacit knowledge~\citep{fj20}.
Thus, proponents of knowledge-driven learning typically argue for it to complement modern ML/DL, not replace it.

There are unsurprisingly many views on the requirements that a hybrid AI system -- this monograph projects such opinions to AutonoML -- should fulfil.
Some researchers state that statistics and data-driven ML should incorporate reasoning, knowledge, common sense, and human values~\citep{mada21}.
These claims add that AI systems need to maintain a cognitive model of their operating environments, and these must consist of accessible and reusable real-world knowledge.
In effect, such programs need not only to identify entities, as current state-of-the-art learning systems do, but also conduct inference and reasoning regarding the relationships between entities.
Ideally, the systems should also be capable of representing and inferring human values based on accumulated common-sense knowledge.
These are fine principles, but other works have indicated that identifying the fundamental building blocks of an autonomous learning system is difficult enough as it is~\citep{ga20, kemu20}; it will be challenging to compose these blocks and formulate effective techniques to make an AutonoML system adapt and evolve in response to learned knowledge.

Nonetheless, hybrid systems based on data-driven and knowledge-driven learning, acting in concert, have broad appeal.
There are two fundamental aspects of human cognitive behaviour that motivate standard proposals: one, the ability to learn from experience and, two, the ability to reason with learned knowledge~\citep{va03}.
Additionally, strong prior knowledge and inductive biases are commonly suggested for making inferences beyond the data~\citep{teke11}.
One study argues that humans acquire this prior knowledge by learning to learn, transferring knowledge from low-level perception to high-level reasoning, and building compositional representations that capture multi-level causal structures underlying a problem/data environment~\citep{la17}.
Several concepts in current ML theories/practices mirror these notions and may be helpful approaches for their integration: transfer learning, representation learning, and multitask learning.
The study also advocates many of the ideas discussed in Section~\ref{possible_path}.
In summary, it states that intelligent systems should (1) form causal models for understanding/explanation rather than limiting themselves with pattern recognition, (2) leverage theories of physics and psychology to enrich learned knowledge, and (3) collect/generalise knowledge for new contexts based on the ability to build compositional representations and learning-to-learn methods.

Despite the strong favour for this possible path within sections of the data science community, the reality remains that constructing an open-world AutonoML system is not an \textit{immediately} feasible project.
Many requirements must be satisfied.
For instance, such a system must be able to react to its working environment, conduct reasoning based on prior knowledge, generalise knowledge from data, provide explanations for outcomes, and make decisions for diverse situations based on available knowledge and experience.
All of this must be done in a continuous online fashion to make better decisions over time~\citep{lamo18}, as it has already been stated that it is virtually impossible to prepare for an open-ended ML application.
Indeed, these autonomous systems need to regularly revise their behaviours and correct errors/biases to refine their learning processes and accommodate current contexts.
Furthermore, AutonoML also needs to support effective HCI to acquire knowledge from humans; see Section~\ref{human_roles_open}.
The point here is that, although automated learning systems have displayed significant progress in the past few decades, at least on a scattered in-principle basis, only those with narrow focus have been realised in practice, possessing limited functionalities and purposes, e.g.~image identification, label classification, prediction, or NLP~\citep{bofa20}.
Given that the approaches behind such narrow forms of ML depend on the availability of big data and the computational power of modern computers~\citep{ma20}, it is an open question as to whether the inclusion of high-level cognition will be more expensive, e.g.~in terms of a conceptual `mental model', or more efficient, e.g.~in terms of extrapolating knowledge via symbolic logic.

What is clear is that, should the state of AutoML research ever reach the stage of automating context understanding, there is plenty of literature from which to source inspiration.
Indeed, the dual-process model of human thought presented in Section~\ref{human_reasoning} has motivated several approaches.
One study contemplating the design of a hybrid learning/reasoning system~\citep{bofa20} argues that the description of System 1 already aligns very well with data-driven ML, formulating models from sensory data to handle perception-based activities such as reading and pattern recognition.
There are, of course, a few discrepancies, with traditional ML struggling with causality and uncertainty, unlike System 1.
Meanwhile, System 2 seems more connected to conventional approaches used widely in the AI community, e.g.~logical reasoning and planning.
Essentially, System 2 is immediately relevant when utilising explicit knowledge, symbols, and high-level concepts.
Another proposal similarly presents a framework with many layers~\citep{beku15}, aiming to integrate sub-symbolic/neural modules and symbolic/cognitive models within an autonomous system to pursue human-competitive levels of intelligence.
Such a framework can theoretically cope with the challenges of an open-ended world, including the need for real-time bidirectional learning, adaptation to changing environments, and handling the emergence of new entities that were not covered by training, e.g.~objects, events, and tasks.
The authors of the proposal additionally claim that an autonomous learning system needs to operate with various levels of representation corresponding to different layers.

Accordingly, a hierarchical AutonoML architecture could include a neural layer learning on the data at its base, an anchoring layer learning elementary symbolic representations of objects, a reactive layer dealing with critical situations, a DL layer learning on more abstract levels, one symbolic layer conducting reasoning and planning, and an even higher-level symbolic layer managing the essential elements of an ontology.
This multi-layer approach is common in the literature, often organising various forms of complementary data/knowledge processing.
In fact, one study argues that knowledge could itself be considered a multi-layered phenomenon appearing at various abstraction levels, with interactions between those levels also contributing to its manifestation~\citep{beku15b}.
In this state, knowledge can be impacted by interactions with working environments, and it can additionally be linked with actions, perception, and thinking.
Hence, the study contends that a future paradigm for the architecture of smart autonomous systems should involve the accumulation and storage of knowledge in a way that captures system-environment interactions, transfers insights in action patterns to symbolic levels, reasons on abstraction levels, and adapts/repairs initial representations for new contexts.
Whatever the case, the many methods and perspectives of hybridised learning, combining data-driven DL with symbolic reasoning, are reviewed in greater depth elsewhere~\citep{ga20}.

Ultimately, while we do not back any particular forecast for AutonoML, especially given that there may be alternate routes to open-world learning based on unfolding complexity and emergent mechanisms~\citep{kemu20, doke21}, the path described in Section~\ref{possible_path} appears plausible.
Certainly, modern ML approaches are insufficient to support trustworthy inference based on limited observations or in the face of newly encountered concepts/contexts.
So, if an effective bridge is developed between sub-symbolic data and high-level symbolic representations, knowledge-based methods could potentially elevate the adaptability of AutonoML.
This outcome is especially the case if a system is equipped with a conceptual meta-knowledge base comprising symbolically coded expertise in various domains; such a construct could help contextualise novel encounters or, via logic-based inference, flesh out a dataset~\citep{beku15}.
As a result, the combination of data-based sub-symbolic processing and complicated abstract reasoning could potentially compensate for the weaknesses of each other, operating in a continuous cycle of learning, processing, reasoning, updating, and adapting.
One may imagine that this will be sufficient for AutonoML systems to understand -- or autonomously pursue an understanding of -- whatever problem context a stakeholder presents. So, with the installation of this new capability, should one expect the nature of HCI to change?

\subsection{Roles and Modes in Relation to Autonomous Open-world Systems}
\label{human_roles_open}

The distinction between proficient AutonoML in an open-ended environment and proficient AutonoML in a constrained setting simply comes down to whether an associated system can autonomously develop an understanding of problem context without human intervention.
Accordingly, the impact of full technical automation on the roles and modes of human interactions, described in Section~\ref{human_roles_constrained}, remains mostly applicable for the upgraded technology.
However, with the automation of context understanding, the nature of HCI will likely evolve in a couple of ways.
First, where human involvement has shifted focus or intensity due to lessening labour demands, extending the scope of automation should continue these trends.
Second, where novel machinery is integrated into existing systems to enable automated context understanding, e.g.~cognition mechanisms, new roles/modes of interactions will arise to support these operations.
We now describe the expected impacts of open-world AutonoML on HCI more broadly.

\textit{Humans launch ML applications in a more general manner.}
Given that open-world learning requires automated context understanding, it stands to reason that the initial phase of the ML workflow will be most affected by this technological upgrade.
Thus, when AutonoML systems are equipped with the capacity to generalise and transfer higher-level knowledge, possibly with conceptual reasoning mechanisms, humans will no longer need to be as narrow and precise with definitions of an ML problem.
For instance, one can envision a stakeholder asking an AutonoML system through a UI to set up a project by supplying the following terms: prediction, bushfire, and USA.
If the application is `from scratch', a cognitive layer might source terms that are associated online, if not causally related, with `bushfire'.
These may form headers for a template dataset that the system provides back to the stakeholder to fill out.
Perhaps the dataset is immediately populated with publicly available data that the AutonoML system suggests is relevant for training.
Even better, if a knowledge base of past ML applications is available to the service, possibly consisting of an Australia-specific model, solution search may be warm-started with a decent ML pipeline.
This notion of porting to similar settings is considered integral to any framework that would be considered a smart adaptive system~\citep{ga05}.

Of course, we still consider problem formulation a human responsibility, even if it can now be done in loose general terms.
However, at its most advanced, AutonoML may have the capability to designate and pursue sub-problems independently.
For example, consider a human setting up carefully constrained prescriptive problems for stock investments, property purchases, and bank savings.
With sufficient capability for open-world learning, it is conceivable instead that a user simply provides a current multi-faceted portfolio and tells the AutonoML system, ``maximise money in one year''.
The service may then work out on its own what sub-objectives to pursue based on the contents of the portfolio.

\textit{The nature of HCI with AutonoML systems continues shifting from direct instruction to collaboration.}
This evolution is based on the fact that, to the degree that AutonoML can be trusted, automated context understanding pushes the limits of its independent performance.
So, while the progression of technical automation from Section~\ref{role_inter_current_automl} to Section~\ref{roles_inter_cmpl_constrained_env} reflects graduation from micro-management to a more moderate form of direction, stakeholders should be no more than laissez-faire managers for a proficient open-world learner.
Human intervention may even harm performance.
However, many believe that, even in the case of system autonomy, humans and machines may still need to coordinate when performing joint tasks~\citep{en17}.
This assertion is because humans and AutonoML systems have complementary strengths~\citep{wida18}.
Humans have strengths in leadership, teamwork, creativity, and social skills.
Programs are capable in speed, scalability, and quantitative computation.
So, in practice, the relationship between human and machine may even be seen as one of collaborative equals, at least for attacking an ML problem.
In fact, it has been proposed that humans and autonomous systems should form a hybrid decision-making collective by reaching consensus or compromise, with both entities sharing moral values and ethical principles~\citep{grro16}.

Naturally, there is no guarantee that principles will automatically be shared in real-world contexts~\citep{roma19}, so it is an open question of how to embed these into a framework.
Actively learning human values via experience is one possibility; in such a case, human involvement in the data collection phase is considered essential, especially for ethically complex problems~\citep{mant21}.
Additionally, an AutonoML system could potentially produce a set of ML solutions, rather than any single optimum, on which humans would preference-vote.
The idea here is that the system indirectly learns human values from the choices that are made and possibly applies them in the future.
Whatever the approach, it is clear that, if cooperative interactions are to be pursued as part of HCI for AutonoML, then humans must have some way of ensuring principles are aligned.
Biases and discrimination in ML solution outcomes must be corrected.
Alternatively, while engineering an AutonoML system, stakeholders may preemptively identify which tasks and conditions are safe for independent operation and which require human interventions.

So, assuming alignment is possible, what would HCI look like in a collaborative relationship?
One work states that the critical human roles in a collective process would include defining problem context, exploring data/knowledge used by a system, understanding the outcomes of an ML application, providing a correct interpretation of the results, and intervening when necessary via feedback~\citep{mant21}.
Such feedback can help AutonoML systems extend their knowledge base and enhance decision-making efficiency and effectiveness.
To envisage this procedure even more concretely, consider the design process of chairs using a computer-aided drafting system called Dreamcatcher\footnote{https://www.autodesk.com/research/projects/project-dreamcatcher}~\citep{rh16}.
This framework can automatically create multiple industrial designs according to the specific criteria provided by a designer, e.g.~materials, required function, and cost.
Some viable solutions may even surprise the designer, emphasising the power of a computational algorithm.
However, the designer still has collaborative influence, using an interactive UI to provide feedback on user likes and dislikes.
Dreamcatcher then adapts its solutions in real-time to these comments.
Ultimately, the designer has the final call, deciding whether to approve an end product, but the system handles the labour of the drafting process.
In this way, stakeholders can avoid tedious tasks to focus more on uniquely human-centric facets, such as professional judgement and aesthetic sensibilities~\citep{wida18}.

\textit{Novel roles arise to support collaborative interactions.}
After all, while democratisation is an aim of the AutoML endeavour, Section~\ref{stakeholder} indicated that there will always be varying degrees of system proximity among stakeholders, especially when large organisations run ML applications.
Accordingly, some humans may have to work along the interface between a team and an AutonoML system.
For instance, one proposal names two new roles that may arise~\citep{wida18}: explainers and sustainers.
Explainers require a mix of technical skills and domain knowledge, serving to translate the behaviours of smart autonomous systems to other users.
Their activities can be crucial in evidence-based fields, such as medicine and law, where non-experts need to understand how ML solution outputs came about, e.g.~pharmaceutical or legal recommendations.
Importantly, the prospective role supplements the explainability mechanisms discussed in Section~\ref{explanation_type}, assuming they are available.
For example, explanatory mechanisms may tell explainers why a self-driving vehicle took actions resulting in an accident, but the explainers then contextualise this information and communicate it to insurers and law enforcement agencies.
As for sustainers, their role is intended to prevent any harmful outcomes of an autonomous system.
This agenda may involve reviewing risk analyses compiled by explainers, as well as checking/validating data, outcomes and mechanical behaviours to ensure compliance with fairness metrics and ethical norms.
In some instances, sustainers may also assess whether collected data is used responsibly to, for example, maintain user privacy.

\textit{Humans monitor and validate AutonoML systems frequently and in real-time.}
Most opinions on advanced AI believe that, even with the capability for independent action, human supervision is inescapable~\citep{en17}.
Computational systems are not biologically human, so there is always a concern that acquired knowledge and enacted decisions may conflict with human rules and values.
The potential for autonomous systems to induce negative consequences through unpredictable actions~\citep{blgr11, st17} becomes even more significant in open-world settings.
Whereas stakeholders can have some degree of expectation for behavioural risks in a well-constrained problem, all bets are off when the constraints vanish.
There have already been hints of these dangers for modern applications, with the Holocaust-denying Microsoft Tay bot acting as a cautionary tale for open-ended learning on, in this case, Twitter.
Thus, once an AutonoML system extends from extrapolating data to deducing high-level concepts and formulating action plans for complex problems, even more opportunities arise for unintended behaviour.
Certain voices in the data science community additionally state that, not only should autonomous systems be subject to human oversight, trustworthiness requires auditing from a body that is independent of an application development team~\citep{sh20}.

\textit{Emergency procedures are established to override undesirable AutonoML behaviour.}
This outcome is a natural extension of the aforementioned need for oversight.
Autonomous systems interacting with complex environments are unlikely to act optimally at all times.
There are many examples where excessive autonomy resulted in disaster, e.g.~Patriot missiles shooting down two friendly planes~\citep{blgr11} or design flaws leading to Boeing 737 MAX crashes~\citep{niki19}.
Many catastrophic failures actually occur because of operations within an open-ended environment, where the system fails to understand a new encounter or a nuance.
As a result, researchers have considered how best to involve humans in activating emergency protocols, e.g.~with a ``big red button''~\citep{orar16}.
Most of this discussion occurs in the broader literature on AI, so it is difficult to predict how relevant all the considerations are.
For instance, some research is concerned about autonomous systems learning to correlate shutdown with performance reductions in pursuing an objective, particularly if these programs operate persistently on long-running applications.
Within frameworks that are engineered in complex or emergent fashion, perhaps the system might then counter a shutdown by disabling the emergency button or obscuring processes/outcomes, i.e.~lying.
Accordingly, related research considers how to manage HCI in such an adversarial setting, safeguarding emergency protocols.
However, other efforts also contemplate how to minimise disruption from such protocols.
For instance, one study suggests that the ``big red button'' could seamlessly switch an autonomous system into an illusory environment, insulating the real world from adverse impacts~\citep{ri16}.
Indeed, for an open-world AutonoML system with well-developed ML pipelines and substantial meta-knowledge bases, a complete reboot can be costly, so any procedure that avoids this is valuable.

\textit{Humans are forced to engineer more rigorous tests for AutonoML.}
Certainly, if an emergency button serves as a reactionary countermeasure to open-world AutonoML going off the rails, then a powerful testing suite serves as a proactive approach to avoid disaster.
Furthermore, unlike emergency protocols, the alternative of continual self-evaluation and error diagnosis~\citep{arsc18} does not need a human to trigger it.
This fact potentially makes a system-testing process a more effective and responsive safeguard, independent of human reflexes.
Thus, both measures for disaster avoidance are likely to be employed in practice.
That said, a prescribed testing suite is only as good as the adverse events it can predict.
Admittedly, the virtually limitless instances of potential error in open-world ML may be compressible into a finite set of error types.
High-level reasoning may not just automate context understanding; it might make it easier to comprehend error.
Nevertheless, the point remains that great effort needs to be directed into making open-world AutonoML robust, and some of that human labour may be available once no longer invested into carefully defining ML problems.
Hopefully, the resources saved by automated context understanding outweigh those required to maintain it.

\textit{Interactions are designed to instill human/social values into AutonoML systems.}
This facet expands on the requirements that collaborative HCI requires.
After all, with AutonoML systems being granted greater agency on account of their presumed superior performance at conducting open-world ML, a key issue is how best to ensure that automated behaviours/actions are for the sake of humans~\citep{mant21}, e.g.~ethically compliant.
Many commentaries in the literature have debated this angle, warning that, to be helpful to humanity, autonomous systems must grapple with potentially conflicting factors, e.g.~in terms of legal requirements, ethics, and safety~\citep{vada18}.
Some additionally stress that socioethical values vary for stakeholders in different multicultural contexts~\citep{di18}; this makes it even more complicated to support the diversity of requirements and explain outcomes/actions to all.
The takeaway here is that, while it is not currently clear how to embed human values into an AutonoML system, HCI will need to support such processes.
This requirement becomes even more relevant when considering that AI applications are moving into the sphere of interpersonal relationships, e.g.~working to adopt complex/subtle human characteristics such as sympathy.
One example is an AI assistant developed by the Koko start-up~\citep{po18}, which analyses the emotions of a user in real-time and then provides advice appropriate for their current sentiment.
In essence, once certain stakeholders begin to interact with ML solutions embedded in open-world AutonoML systems on an emotional level, psychological damage becomes a real risk that must be managed with intelligent approaches.

\textit{Society begins to debate what responsibility and trust mean concerning AutonoML.}
The projected evolution of the field means that such concepts are no longer as clear-cut as in Section~\ref{human_roles_constrained}.
Indeed, while equipping AutonoML systems with automated context understanding has a functional purpose, enabling a way to effectively/efficiently solve general ML problems with minimal human effort, the development of high-level cognition inevitably raises profound philosophical and jurisprudential issues.
In purely functional terms, an open-world AutonoML framework can be compared to a human problem solver that is (1) extremely proficient at applying inductive reasoning to observations and (2) has no individual agency, i.e.~they do not come up with their own core problems to solve.
Then again, perhaps even this assertion is controversial?
Ultimately, we do not take sides in such broader debates, but it is clear that the notion of responsibility becomes critical to the topic of autonomous systems~\citep{di17}.
Admittedly, much discussion in this area relates more to modern systems, i.e.~non-autonomous tools, and the risks of irresponsible usage when left to non-experts.
Some argue then that experts should take more responsibility for the development/deployment of smart autonomous systems, ensuring consistency with fundamental human principles and values~\citep{di18}; these works urge that responsibility be foremost in mind rather than an afterthought when designing such frameworks.
However, when decision-making arises purely algorithmically, potentially via human-adjacent cognitive processes, society will have to debate where to attribute legal/moral culpability for adverse outcomes.
Does the fault lie with the theoretician that designed the algorithm?
The developer that implemented it?
The user that applied it?
The machine?
Nobody, i.e.~an ``act of God''?

Closely related to this issue is the matter of trust.
As Section~\ref{outcome_trust} indicates, there have already been many proposals on how to define this concept.
For instance, one proposal claims that trust in automation is a confluence of several factors~\citep{en17}, i.e.~system-related, human-related, and situational.
System factors include validity, reliability, robustness, interpretability, predictability, and the number of recent failures.
Human factors capture the ability of a user to understand outcomes/processes, their willingness to trust, and other personal characteristics; these include culture, gender, and age.
Situational factors cover time/resource constraints, required effort, workload, and availability.
Even so, once open-world AutonoML enters the public domain, it will not be sufficient to simply quantify trust.
Society will also have to decide the threshold for trust, per ML application, beyond which, if things go wrong, stakeholders are prepared to ask the hard questions on responsibility.

\textit{Regulatory and governance mechanisms are established for AutonoML.}
This outcome is predicted to be a social consequence of the open-world ML technology entering the mainstream.
Indeed, it stands to reason that, given all the previously discussed risks of AutonoML operating in open-ended environments, legislation will eventually cover such services.
One study contemplates this topic, discussing governance and regulatory mechanisms for autonomous systems~\citep{ra18}.
It argues that the learning algorithms in these systems have to be transparent, fair, accountable, and concordant with values shared by all stakeholders.
The study proposes a framework that combines a human-in-the-loop paradigm with automated mechanisms, seeking to identify a consensus set of values for all stakeholders and then ensure that the system complies with these established values.
Of course, where human intervention is potentially limited, explanatory mechanisms become the primary tool for regulatory oversight.
Such processes are also helpful for sufficiently supporting humans in deciding whether to trust and use an autonomous system~\citep{en17}.

In conclusion, it may seem that such a focus in this section on the negative aspects of open-world AutonoML may suggest that the technology should be avoided.
However, counterintuitively, this simply highlights the prospective power of automation, in that human resources are freed up from so many processes that they are, by and large, foreseeably redirected to mitigating risks.
This eventuality ideally ensures that open-world AutonoML is as robust and safe to use as possible, with most of the HCI involved serving as a conduit to deliver beneficial ML solutions/outcomes to stakeholders.

\subsection{Optimising Interactions with Autonomous Open-world Systems}
\label{optimisation_interaction}

In operational terms, it seems likely that most stakeholders interacting with open-world AutonoML, for the functionality of solving ML problems, will do so in a collaborative manner.
This eventuality would be a shift from HCI practices with modern AutoML, as discussed in Section~\ref{role_inter_current_automl}, so it is worth diving deeper into how such collaborative interactions may be optimised.
First, it is worth acknowledging that biological brains and computer programs, to date, have different strengths and weaknesses.
While machines excel at swift computation, humans are better at tasks requiring strong cognitive and social skills, also being capable of quick adaptation to a wide range of unforeseen situations and events~\citep{kova21}.
Therefore, to optimise collaborative performance when solving a set of ML problems, it may be worth acknowledging individual strengths and using them to devise effective protocols/frameworks for distributing sub-tasks between stakeholders and the autonomous system.
Finding the right balance for this is immediately challenging because it can be difficult to determine the appropriate amount of autonomy that establishes a good `teammate' for a human stakeholder~\citep{ac21}.
Too little autonomy goes against the core point of the automation endeavour, burdening humans with system management and control.
Too much autonomy risks the rise of issues related to trust, safety, and explainability.

Contemplating further on how human-computer collaboration may be designed, one study delves further into the distinct qualities that each entity possesses~\citep{kova21}.
It notes that AI systems are much better than humans when selecting and handling vast amounts of data quickly, correctly, reliably, and consistently.
The nature of electronic circuitry also ensures that computer programs do not fatigue or suffer from emotional volatility and stress.
Such systems are also arguably better at memorising insights and recalling skills.
The study thus advises that tasks suiting these capabilities should be delegated to an AI system, despite noting that it is still crucial for humans to somewhat understand and master these processes so that sufficient action can be taken when an autonomous system errs.
On the other hand, compared with standard ML/AutoML, humans are more responsive to unpredictable events, creating innovative solutions for open-ended problems that are ill-defined.
These abilities are arguably based on human qualities, primarily high-level cognition, although there is a possibility that psychosocial skills contribute too.
So, by delegating sub-tasks appropriately, the study mentioned above suggests a hybrid decision-making process could leverage the power of the human mind where beneficial while also addressing cognitive constraints and decision biases.
Granted, a core idea behind automating context understanding and moving towards open-world AutonoML is to gradually shift the high-level cognitive load from human to machine.
However, as with the full technical automation projected for closed-world AutonoML, we neither claim that mechanised operations will be perfect nor that they should never be complemented by humans, especially where stakeholders have access to knowledge that is not immediately available to the system.
Thus, compromise designs that balance automation with meaningful levels of human control may arguably form the safest and most effective frameworks for autonomous AI~\citep{sh20b}.
Compromise designs do not need to be inflexible either, with mechanisms generally functioning automatically whenever they are understandable, predictable, and accountable.
The ideal is that a UI simply allows human override when contexts produce issues, e.g.~high levels of uncertainty.

Many more opinions exist on the topic of human-computer collaboration and its requirements.
During one panel discussion, researchers argued that this higher form of interaction requires mutual goal understanding, predefined task co-management, and shared progress tracking~\citep{wach20}.
They suggested that collaborative HCI should pursue an approach towards what they termed Computer-Supported Cooperative Work (CSCW).
The CSCW perspective attempts to systematise key facets of cooperative work, i.e.~awareness, articulation, and appropriation.
High awareness indicates that both stakeholder and AutonoML system are cognisant of the operations that the other entity partakes in.
Good articulation indicates that work can be split into logical units and distributed between stakeholders and the AutonoML system, subsequently integrated.
Effective appropriation indicates that a stakeholder can adapt and apply an AutonoML system to their specific situation.

Awareness in particular, as in the exchange of information between cooperating parties, has been deemed particularly important to collaborative pursuits.
One study asserts that, beyond following a common protocol, humans and an autonomous system must have a shared comprehension concerning the objectives of a problem and the aspects that are relevant to them~\citep{koen12}.
In the terminology defined within this monograph, there must be adequate HCI support for ensuring that problem formulation is sufficiently comprehensive and its requirements are propagated into an AutonoML system, especially given that this initial sub-phase of an ML workflow still relies on humans.
That said, the study continues to suggest that the comprehension of goals and relevant processes to conquer these goals can abstractly be considered a mental model, serving as the basis for both humans and systems to plan goal-driven approaches for arbitrary situations, reacting capably to dynamic changes in working environments.
Consequently, an ontology-based method is proposed to maintain situational awareness, represent/infer knowledge about operating contexts, and communicate individual perceptions between humans and the system.
Indeed, a shared level of situational awareness is commonly thought to be a fundamental requirement for successful coordination between collaborating parties~\citep{en17}.

\begin{figure}[!ht]
	\centering
	\includegraphics[width=0.95\textwidth]{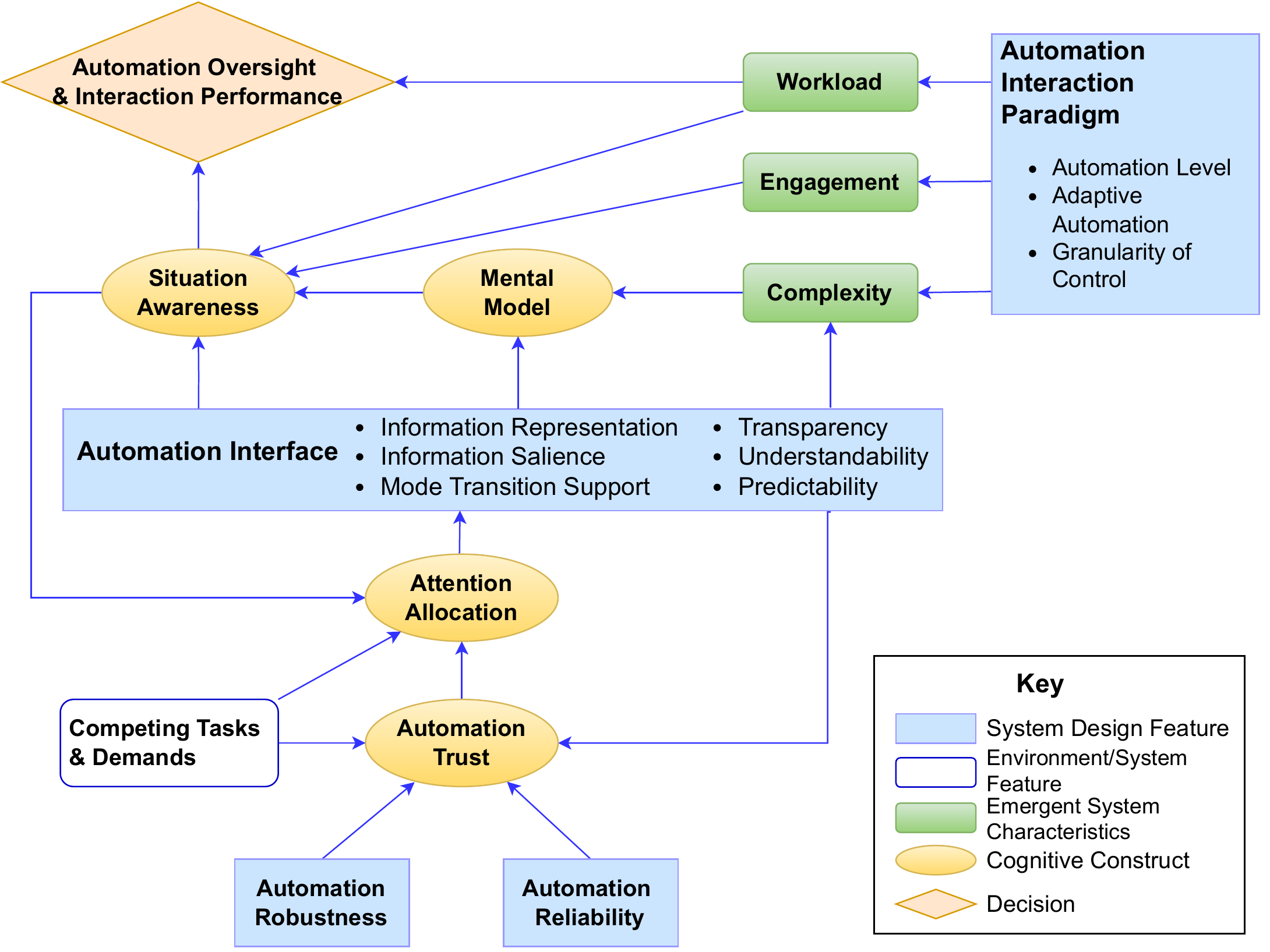}
	\caption{A human-autonomy system oversight (HASO) model~\citep{en17}. The paradigm presents major system design characteristics impacting human cognitive processes during engagement with an autonomous system, e.g.~inspection, intervention, and general interaction.}
	\label{fig_human_autonomy_interface}
\end{figure}

Now, once again, we restate that an ideal open-world AutonoML system should be proficient at solving general ML problems independently, so, while joint operations for the pursuit of \textit{learning} should be supported, they may not be required in all cases.
Thus, most modern discussion about HCI concerning advanced AI keeps returning to matters of \textit{oversight}.
Herein lies a conundrum: the more automation a system has and the more robust/reliable it becomes, the less aware human stakeholders are of its situation and the more challenging it is for them to take manual control when needed.
In response to this, one research effort proposes a human-autonomy system oversight (HASO) model~\citep{en17}.
Schematised in Fig.~\ref{fig_human_autonomy_interface}, HASO describes important system-design characteristics that potentially influence human performance when working with autonomous and semi-autonomous systems.
The framework consists of features relating to system robustness/reliability, operator interfaces, and three paradigms for interaction.
In general, situational awareness and workload are vital to determining the quality of oversight and interaction outcomes, dictating where humans allocate their attention within an autonomous system.
Situational awareness can be heavily impacted by UI design and the manner in which HCI is performed.
Of course, increasing system reliability and robustness can increase the trust level of operators, thus somewhat relaxing the need for stakeholder attention.

Based on existing studies of UI characteristics and automation paradigms described in the HASO model~\citep{en17}, guidelines can be proposed for the design of trustworthy autonomous systems, supporting an optionally high level of human oversight and situational awareness.
This guidance advocates for three key principles, i.e.~supporting human perception, minimising system complexity, and improving situational awareness for operators.
The proposal also cautions that increasing the degree of autonomy for a system must be accompanied by the development of UIs that provide a high level of transparency, understandability, and predictability regarding its behaviour.
Simultaneously, these UIs must also provide operators with critical features to comprehend pivotal states and mode transitions occurring within an autonomous system.
Admittedly, this discussion may seem abstract and unnecessary overengineering for modern AutoML services, which primarily focus on the model development phase of an ML workflow.
However, as described in Section~\ref{roles_inter_cmpl_constrained_env}, AutonoML systems are already expected to become quite complex when automation extends along the other technical phases of the workflow~\citep{kemu20}.
Mechanisms are forecast to become wired up in an even more complicated manner once higher-level cognition enters the picture.
Accordingly, HASO and the like can become useful distillations of HCI requirements to pursue when complexity threatens to further obscure operating processes.

Ultimately, the pursuit of open-world AutonoML will require updated perspectives on supporting operation and oversight with solid HCI strategies and mechanisms.
Traditional thinking in this space tends to be one-dimensional, in that the price of pursuing automation is a necessary reduction in human controls~\citep{shve78, pash00}.
However, modern viewpoints argue that this perspective is incorrect, especially with the surge in concerns around reliability, safety, and trust.
An example of one such counterpoint proposes a two-dimensional Human-Centered Artificial Intelligence (HCAI) framework that treats automation and human control on 
orthogonal axes~\citep{sh20}, emphasising that humans should have the \textit{option} of intervention alongside the \textit{option} of letting the machine do all the work.
According to the HCAI framework, a good design embraces the advantages of automation, including sophisticated algorithms, large-sized databases, powerful computational performance, and advanced sensors.
A good design also supports human capability via smart UIs, augmenting, improving and empowering stakeholders so that they may use, oversee, master and refine autonomous systems.
All this requires well-engineered services that are transparent, so that machine states and operational processes are understandable, indicating when humans should be involved and when they may sit back and relax.
Admittedly, the devil is always in the details, and fully automated AutonoML systems are yet to be implemented for well-defined problems, let alone open-ended environments.
Engineering issues may yet unveil unexpected challenges for how UIs connect with internal mechanisms.
Nonetheless, the ongoing conversation about HCI design in the broader AI literature is almost directly applicable to ML.
Hashing out principled approaches in the present can only accelerate technological progress once the world finally opens up for AutonoML.

\section{Critical Discussion and Future Directions}
\label{open_issues}

As of the early 2020s, there exists an abundance of review articles on the \textit{mechanics} of AutoML, even though few are as broadly scoped as those that precede this monograph~\citep{kemu20, doke21}.
However, deeper discussions on AutoML ergonomics, studying how humans and these systems can most effectively interact, are sparse.
In fact, while recent work has surveyed how modern-day AutoML technology has organically evolved to suit human engagement in real-world applications~\citep{scke21}, comprehensive considerations of HCI, present and prospective, are currently novel in the field.
Thus, having already reviewed perspectives on HCI in detail, mainly how it is likely to evolve alongside expected advances in high-level ML automation, we present an overarching discussion of critical themes in Section~\ref{discussion}.
We then conclude in Section~\ref{future} with an examination of potential future directions for HCI research in the field of AutoML.

\subsection{Critical Discussion}
\label{discussion}

Although a cautious perspective is always appropriate, the mainstream emergence of AutoML theory/practice has achieved many remarkable results over the last decade.
Initially propelled by the optimisation community as an approach to tackle the long-standing problem of algorithm selection, AutoML has since begun to expand to the other phases of an ML workflow, e.g.~data engineering, deployment, and monitoring and maintenance.
As progress continues, many researchers have developed their own views on what the `final goal' of the endeavour may be.
Some are motivated by pushing the mechanical limits of what might be achievable, e.g.~whether the learning and problem-solving capabilities of an independent AutonoML system may ever rival and surpass those of a biological brain.
Others appear invested in the field for the possibility of more human-centric benefits, e.g.~making ML models more transparent, fair, reliable, and accessible to non-experts~\citep{xiwu21}.
Unsurprisingly, these agendas need not necessarily be exclusive, even though limitations in computer resources will always put practical constraints on which system features should be prioritised.
Ideally, hardware willing, AutoML will advance to a state where a framework is capable of autonomously searching, learning, reasoning and adapting within arbitrary problem/data contexts, while still supporting trustworthy collaborations with humans and other mechanised systems.

Whatever the state of AutonoML becomes in the long-term future, it seems clear that, at one level or another, humans need to remain in the loop for AI systems~\citep{br20}.
This sentiment is echoed by ML pioneer Michael Jordan~\citep{pr21}, who claims that ``we will need well-thought-out interactions of humans and computers to solve our most pressing problems''.
His view additionally suggests that ML research should not prioritise a mimicry of human thinking in the pursuit of AGI.
The power of ML is in discovering hidden insights in multiple large datasets, mining obscure patterns, and recommending novel courses of action.
In essence, the advocated ideal is that learning systems continue to act in applied service to humans, augmenting biological intelligence and offering new services to users in varied domains, e.g.~commerce, manufacturing, health care, and transportation.
Of course, in purely practical terms, autonomous systems will continue to need good HCI solutions for the foreseeable future, possibly for decades~\citep{en17}, so that human controllers can collaborate or, at the very least, oversee system performance and safety.
Certainly, on the operational side, the involvement of technical stakeholders is expected to remain valuable simply due to their domain knowledge as well as the use of intuition/experience to cover the blind spots of an AutoML mechanism~\citep{xiwu21}.
However, ensuring AutoML usage remains socially responsible is also a strong motivation for efficient and productive human-computer communications.
Accordingly, the uptake of AutoML technology may be better served by proactively researching and planning strong HCI guidelines/protocols at every stage of its technical evolution, rather than waiting for UI paradigms to arise reactively as an afterthought~\citep{scke21}.

The historical neglect of debating human roles in the context of AutoML means that many human-centric questions, both important and challenging, have mostly been overlooked.
For instance, in which situations and to what extent can AutoML-supplied predictions/prescriptions be believed?
Are there thresholds that demarcate when independent operation is appropriate?
Whose interests are AutoML systems serving?
Are the outcomes fair for everyone?
At which point of an ML workflow does a discovered bias enter the problem-solving process?
How can humans and AutoML systems best compensate for the limitations of each other?
Is there an optimal way of translating outcome into impact and human benefit?
The answers to many of these questions have broader implications, tying into what society expects for the roles of humans concerning the evolution of industry.

Notably, in recent years, certain terminology has been popularised for the notion of industrial revolution.
Put simply, the following associations have been made: Industry 1.0 with steam/water power, Industry 2.0 with broad electrification, and Industry 3.0 with digital computer technology.
It has been argued that society is currently undergoing a fourth revolution based on the rising importance of data networks to modern applications.
Specifically, Industry 4.0 can be characterised as contributing to an interconnected world, where physical items such as sensors, devices, computers and enterprise assets are linked, communicating with each other and the Internet~\citep{sial16}.
Commentary in the literature suggests that Industry 4.0 appears to be the perfect setting for the construction of smart autonomous systems fuelled by big data and ML~\citep{vaam18}.
This claim is because the development of large-scale machine-to-machine communication, such as with the `Internet of Things', arguably supports the engineering of complex systems that are better able to work, self-monitor and diagnose problems without human intervention~\citep{mo18}.
In effect, the focus of Industry 4.0 appears to be an increasing level of automation and digitalisation, mainly manifested via AI-driven technologies.
Unfortunately, as is asserted by some sectors of the data science community, human roles and social considerations, e.g.~fairness, seem to take a back seat in the Industry 4.0 paradigm, at least with respect to pure technical capability.

Just as this monograph attempts to advance the discussion around AutoML from technical operations to HCI factors, so too has there been a movement to shift the paradigm of progress into what is being called Industry 5.0.
Indeed, this concept of industrial revolution appears to be a logical progression of the existing Industry 4.0 paradigm, simply bringing human, social and environmental factors back into the bigger picture~\citep{mu20}.
In effect, the driver for industrial progress is no longer automated task proficiency, which remains important, but is instead a collection of values including human-centricity, sustainability, resilience, and ecological/social benefits~\citep{brde21}.
A broader range of human demands and interests become core to the development process of technologies.
Consequently, the adoption of technological processes is dependent on an ethical rationalisation of how human values are served by a system, rather than relying on a purely technical viewpoint~\citep{mu20}.
For instance, many DL approaches that are considered state-of-the-art under Industry 4.0 may be less appealing under the Industry 5.0 banner due to energy usage and their impact on climate change~\citep{doke21}.
Additionally, the notion of \textit{co-working} becomes very important to Industry 5.0, rebounding from the pure automation focus of the previous revolution.
The idea is that merging human and machine abilities will complement disparate strengths and cover individual weaknesses.
This paradigm supports humans using their creativity in problem-solving, formulating new roles specific to problem context, and improving skills and knowledge in collaboration with autonomous systems.
Clearly, Industry 5.0 puts a greater weight than Industry 4.0 on intelligent human-machine interfaces, the integration of human cognitive abilities into AI systems, and collaborative interactions.
These characteristics are pursued in the hope that products and services will satisfy human needs, reduce environmental impacts, and use energy resources efficiently~\citep{mu20}.

With all the above context, this review contributes to fleshing out the overarching picture of human interactions, i.e.~their roles and modes, in the context of the evolving AutoML field.
Section~\ref{role_inter_current_automl} anchored this perspective by first establishing the stakeholders and phases of an ML workflow that are currently involved in modern-day AutoML interactions.
Given that AutoML can be assessed as having far further to go, simply in terms of technical operations, Section~\ref{roles_inter_cmpl_constrained_env} proceeded to step forward in time and envisage full automation for every process following problem formulation and context understanding.
This advance already has significant implications for how humans engage with these newly autonomous problem-solvers.
However, with learning in open-ended environments as a holy grail for the field of AI in general, the subsequent Section~\ref{roles_inter_open_env} considered how AutonoML could approach such an ideal.

Open-world learning may not be entirely blue-sky, as knowledge transfer between similar and/or different domains has been investigated intensively as part of the meta-learning research agenda, e.g.~reusing knowledge and search strategies from prior tasks to speed up ML pipeline construction for similar problems~\citep{woho18,albu20}.
The greatest obstacle remains identifying when and which meta-knowledge can be leveraged once incoming data and related concepts, perhaps even ML tasks, are no longer stable.
It seems unavoidable then that the pursuit of context understanding must be automated.
Perhaps the most prevalent opinion on this matter, especially from the traditional AI community, is that this cannot be achieved without installing AutonoML with the ability to reason, transfer and generalise knowledge on a higher level than data.
Whether this is the right approach is debatable, but it is at least clear what the capabilities of an ideal open-world AutonoML system are, meaning that well-reasoned forecasts for HCI are available.
There are, of course, grander implications when AutonoML systems become able to autonomously learn, self-monitor, correct errors, reason, and generalise knowledge.
Some have speculated that a machine with these human-level qualities may even mark its own revolution: Industry 6.0.
Perhaps a follow-up to this monograph will eventually be necessary, contemplating the roles and modes of human interactions for systems with general intelligence and self-cognition.
For now, we limit speculation.

For now, present-day thought in the data science community cannot foresee a future where human involvement is completely extricated from AutonoML operations.
A recent study specific to modern AutoML systems affirms the prevalence of an opinion that human developers need to provide inputs/feedback simply because a computer, in the absence of expert knowledge, cannot attain a human-level understanding of context beyond the datasets provided~\citep{wawe19}.
To many, it seems essential that technical and business stakeholders guide data cleaning/transformation phases so that AutoML algorithms can better satisfy application requirements.
As a result of these views, the study advocates that an AutoML system should act as a collaborative tool that augments human capability rather than fully automating human operations.

In fairness, even were automated context understanding to be achieved, humans who formulate a problem are typically the ones who ultimately evaluate whether an ML solution satisfies their requirements.
There is no standard procedure for developing trust, any more than there is a universal metric for fairness, and different cultural backgrounds will equip stakeholders with different expectations.
Thus, even if an AutonoML system can solve a problem `perfectly', human involvement is indispensable for ensuring that form of perfection accords with the definition a stakeholder chooses.
Unsurprisingly, given all these motivations, several recent studies favour the development of human-guided AutoML to balance stakeholder control with automation~\citep{lema19, giho19, wawe19}.
Various proposals further suggest different requirements for the design of effective HCI~\citep{giho19, amwe19}.
In effect, human-guided AutoML enshrines one mode of collaboration, where human users define their domain knowledge and high-level expectations of a system; the AutoML framework then performs its model search autonomously to create high-quality ML solutions for a given problem~\citep{giho19, lema19}.

Currently, it is not surprising that computer programs fuelled by sophisticated state-of-the-art ML algorithms struggle with adapting to new contexts without human innovation and rigid problem definitions.
Existing AutoML mechanisms tend to automate only one aspect of an ML workflow, and often only in principle, e.g.~feature engineering, model selection and HPO, deployment, or monitoring and model maintenance.
Moreover, most applications still require heavy human involvement across the entire ML workflow to supervise/guide ML model construction, complement the lack of domain knowledge, verify results, and intervene when outcomes are unexpected.
Admittedly, as new learning paradigms and supporting procedures are developed, so that automated high-level operations can be described as `proficient', the need for human intervention is likely to relax.
Section~\ref{search_strategy} mentioned AI systems capable of surpassing humans in Chess and Go without relying on expert knowledge as examples of this expected progression in AutoML.
Even then, the research-and-development journey to fully autonomous open-world learning is a long one, beset by many challenges described in Section~\ref{open_ended_learning}.

Accordingly, although human intelligence may not be the perfect standard for problem-solving, it can still inspire research in the way humans learn, infer, and efficiently address diverse and unpredictable situations within open-ended environments despite the absence of massive datasets.
The challenge has long been known that an autonomous open-world system does not only have to collect relevant data regarding a new working environment; it also has to comprehend the \textit{meaning} of that information and quickly model environmental dynamics to engage in a proactive decision-making process~\citep{en95}.
Likewise, another study argues that responding quickly to environmental changes requires capturing information from various sources, making decisions based on both obtained information and existing knowledge, and launching actions following these decisions~\citep{ulmo14}.
Additionally, the system must have feedback about the impact of its actions on an environment so as to update its knowledge base and pursue better decisions in the future.
This continuous cycle of learning may seem overly involved for most present-day AutoML systems that solve problems under a one-and-done paradigm, but the AutonoML frameworks of the future will certainly need to fulfil these requirements.
Moreover, if collaborative operations become the optimal form of HCI for open-world AutonoML, careful design of advanced interfaces will be required to imbue a team of stakeholders and computers with a shared amount of situational awareness~\citep{en17, koen12}.

Ultimately, the most significant motivating factor for further research into HCI may simply be what is the greatest weakness of modern AutoML offerings.
Simply put, in pursuing automation, many current AutoML tools have obscured their internals, acting as black-box systems~\citep{xiwu21}.
This situation immediately restricts those that wish to have flexibility in managing the ML process.
The insertion of domain knowledge is likewise hindered without a clear understanding of what is happening internally~\citep{liwa17}.
Indeed, many have hinted at the dangers of an autonomous system that cannot show stakeholders what they are doing, why they are doing it, whether an action can be averted, and how a human may avert such an action~\citep{brho13}.
Arguably, if humans are not allowed to provide inputs to an AutoML system, the bare minimum of interactivity should be the provision of outputs, i.e.~explanatory mechanisms that allow stakeholders to understand system behaviours and the rationale behind decision-making processes.
This support would allow users to at least determine whether it is worth accepting an ML solution, although being able to direct a learning process and integrate domain knowledge is obviously preferable.
In fact, experimental observations and analyses of case studies have shown that ML model development can benefit from human involvement via the notion of `soft information and knowledge'~\citep{tako16}.
This concept covers additional knowledge that may be ill-defined or otherwise unavailable for pure machine-centric methods~\citep{chgo15}; it includes theories, intuition, belief, value judgements, and so on.
Regardless of details, the principle remains that people tend not to use results they cannot understand or explain.
Therefore, explainability is critical to support understanding and build confidence, e.g.~allowing for the diagnosis of odd behaviour.
Section~\ref{improve_outcome} indicated that these matters of trust in AI are highly topical in the early 2020s, and, without a serious effort from the AutoML community at grappling with HCI and AutoML ergonomics, uptake of related technologies may stall.

In summary, this monograph has reviewed several aspects regarding the roles and modes of human interactions with AutoML systems.

\textbf{What are the types of stakeholders that may interact with AutoML systems?}
Many humans may directly or indirectly engage in an ML application, particularly as part of a business organisation.
We categorise them as the following: technical, business, regulatory, and end-user.
Technicians are typically responsible for developing, testing, deploying, monitoring, and maintaining ML models.
Business stakeholders tend to translate business criteria into functional requirements for an ML solution, with some assessment of whether an application is feasible.
Stakeholders in the regulatory group are in charge of independently assessing and verifying the safety and reliability of the systems, ensuring that they comply with legal, ethical and social norms.
End-users are the ones that feel the impacts of an ML solution; if this is done knowingly via direct interaction, they can evaluate its usefulness, potentially provide feedback, and determine whether the solution is accepted.
Due to differences in these roles, interaction modes and the flow of information between stakeholders and systems can vary.
Therefore, this monograph clarified how different types of users currently interact with AutoML systems in Section~\ref{stakeholder}.

\textbf{What are the current interactions involved when developing, deploying, and maintaining ML solutions?}
To answer this question, we reviewed how stakeholders, primarily technical, can provide inputs into an ML application at every stage of an ML workflow.
However, we considered this in the context of modern AutoML and also attempted to gauge the current balance of human and machine involvement in these activities.
Discussions in Section~\ref{roles_dev} included how stakeholders guide the development process of ML algorithms, how they deploy and monitor ML solutions, and how such solutions are maintained and updated in response to changes in a data environment.
We also reviewed the current and prospective communication modalities with which stakeholders can interact with an AutoML system or, more frequently, the ML solution it houses.
These modalities, detailed in Section~\ref{multimodality}, covered UIs based on touch screens, voice, gestures, VR/AR, and even brain signals.

\textbf{How do socioethical factors influence interactions with AutoML systems?}
As supported by various studies~\citep{drwe20}, data scientists only trust ML solutions provided by AutoML tools if mechanisms establish a high degree of transparency and understandability regarding system processes and outcomes.
Similarly, end-users only accept ML predictions/prescriptions if the learning system can show the reasons behind these decisions~\citep{wani21}.
For example, if an automated decision-making system in a self-driving car can explain how it identifies/avoids obstacles and provides warnings for various road situations within a simulated environment, users will have more trust in the autopilot system.
Therefore, explainability is a critical factor that is likely to influence public uptake of AutoML technology.
However, different forms of explanation have different impacts on human comprehension, so this monograph delved into the nuances of the topic, considering how interfaces affect explainability and how, in turn, explainability affects human interactions.
Additionally, we linked trust in AutoML outcomes with other factors such as cognitive biases, fairness, and ethical norms.
The importance of these facets, as well as ways to manage them with strong human-computer collaborative practices, was explored in Section~\ref{improve_outcome}.

\textbf{Are there optimal design principles and requirements for AutoML UIs?}
Certainly, to foster effective interactions between stakeholders and AutoML systems, intelligent UIs are required~\citep{giho19}, particularly those that allow a technician to specify ML problem settings, explore the characteristics of data, and constrain ML solution search spaces.
A good UI essentially allows users to efficiently inject their domain knowledge, optionally assisting AutoML mechanisms in finding optimal ML pipelines.
Such domain knowledge may be unidentifiable from merely inspecting the data, sourced instead from contexts where the data is generated, previous results in the literature, or simply conventional wisdom for applying ML models~\citep{giho19}.
Furthermore, UIs must allow technicians to easily explore hyperparameter space themselves so that they may compare various generated solutions and potentially even reject them in favour of a manually selected ML pipeline.
These are just some of the requirements to support model development; there are many other general guidelines for designing a UI that supports practical operations during deployment, continuous learning/adaptation of an ML solution, and other significant interactions.
Together, these principles and their proposals were reviewed in Section~\ref{user_interface}.
Of course, once AutoML is capable of problem-solving in open-ended environments, adapting and operating with heightened autonomy, the nature of HCI potentially evolves into a new form of collaboration.
Thus, Section~\ref{optimisation_interaction} returned to the question of UI design, detailing many proposals in the literature covering oversight paradigms, task division methods, and shared situational awareness.

\textbf{How do the roles of humans change as AutoML technology evolves?}
The original purpose of the AutoML endeavour was to aid data science workers, and ideally non-experts, in finding optimal ML models for arbitrary ML problems with a minimal amount of human intervention.
Even after a decade of concerted efforts, the field remains nascent, despite its great promise.
However, it is expected that advances in technical problem-solving paradigms will eventually bring automation to virtually every phase of an ML workflow.
Advanced strategies for search, meta-learning and adaptation will produce continuously operating AutonoML systems that are proficient, not necessarily perfect, at finding good ML solutions to constrained ML problems.
In time, AutonoML research is then likely to open up the problem contexts in which a system operates, although the capacity for efficient adaptation to novel tasks and concepts may require revolutionary progress.
Throughout this technological evolution, the roles and modes of human interactions are also projected to change.
In current practice, described by Section~\ref{roles_dev}, humans are usually involved directly in the development and deployment of ML solutions, selecting suitable training data and features, establishing configuration ranges for hyperparameters, assessing resulting pipelines, and monitoring the operation of solutions when deployed.
Once full technical automation is enabled by strategic advances, discussed in Section~\ref{search_strategy}, the direct involvement of stakeholders is no longer required.
Instead, humans are more likely to focus on the innovative phases of problem formulation and context understanding, setting up constraints and working environments for AutoML systems, as well as supervising, auditing, and understanding the outcomes of AutoML systems.
Once context understanding or the pursuit thereof is itself automated, the key roles of humans evolve even further; one responsibility is to maintain the novel mechanisms that handle knowledge generalisation and transfer, e.g.~by fine-tuning a `mental model' of the real world.
Oversight of autonomous operations is another task, which is especially important in an unconstrained environment.
However, ultimately, an open-world AutonoML system may start to be seen as a teammate, hence HCI being reframed as a conduit for collaborative problem-solving.
Humans may become responsible for finding the most productive way to cooperate with an AutonoML entity, assessing socioethical factors in system behaviour, monitoring/validating outcomes in real-time, and seizing control in the case of erroneous operations.
In short, we predict that human roles will likely prioritise tasks requiring high-level cognitive abilities, leaving AutonoML for the most part to do its job and proficiently solve ML problems.

\subsection{Potential Research Directions}
\label{future}

It is a common opinion that one of the most challenging issues of the coming years will be how to leverage the full power of autonomous machines while simultaneously supporting seamless coordination with humans, generating decisions that align with sophisticated values and needs~\citep{liaj21}.
The topic of HCI in AutoML is presently in its infancy, only recently emerging as the technological rise of AutoML motivates questions about how humans fit in.
This transition in thinking, from a purely technical focus to one that ponders human-centric solutions, mirrors gravitation in the broader data science literature towards the concept of Industry 5.0.
Therefore, given the seemingly inevitable attention that HCI will need to be paid, we now identify potential research directions for this topic.

\begin{itemize}
	
	\item \textbf{Generating human-centred design principles for the selection of optimal ML solutions.}
	Crucially, while ML theory has been researched for many decades, it can be argued that ML technologies have only existed in the mainstream for a couple of those decades, and genuine industrial use has only taken off within the last few years.
	Consequently, there has been a significant historical focus on maximising the technical quality of an ML approximation, whereas satisfying human needs and desires is a relatively novel notion.
	The two aims do not necessarily have to compete, but, as Industry 5.0 attempts to recognise, technical performance is a narrow view of good performance.
	It is thus likely that human-centred design (HCD) principles~\citep{mant21} will become increasingly important in the coming years.
	The main activities of HCD for AI systems include the specification and comprehension of contexts and requirements, the latter being held by business stakeholders and end-users.
	Managing these processes well requires developing a solid understanding of who the relevant stakeholders are, what they care about, how they are expected to interact with an AutoML system, and what the working environments and constraints for the framework are.
	Hashing out ergonomic HCD principles in the near future should make the next generation of AutoML services much more amenable to collaborative interaction, thus enhancing functionality, operational efficiency, and UX.
	
	\item \textbf{Promoting transparency and explainability in AutoML systems.}
	This monograph stressed these socioethical concepts several times as being of surging importance in the current AI climate.
	Alongside related concepts such as reproducibility and predictability, these factors support stakeholders supervising the actions and behaviours of an AutoML system, giving them the ability to intervene when errors and other adverse outcomes occur.
	Certainly, humans are less likely to use and trust AutoML if they do not feel in control or are unable to explain system processes~\citep{sh20}.
	Thus, in lieu of loftier aims such as open-world learning, moving towards transparency with explanatory UIs and white-box system design may be an important and achievable short-term goal within the field of AutoML. 
	
	\item \textbf{Integrating human values in terms of fairness and ethics into AutoML systems.}
	Once stakeholders can understand an autonomous system, the next step is to ensure its processes and outcomes adhere to human requirements.
	However, it is currently unclear how to do that because current definitions, rules and guidelines related to concepts such as fairness are generic and abstract~\citep{mant21}.
	It is not clear how to define every manifestation of bias and discrimination within the context of AI in a simple fashion~\citep{cant21}.
	Unsurprisingly, it is even more challenging to devise realistic solutions that address the fallacies.
	In contrast, there is a high bar of expectation for ethical and fair AI systems, which need to possess good normative models that easily integrate their behaviours into human normative institutions~\citep{liaj21}.
	Trustworthy AI must not only explain how an outcome was achieved but also provide normative justification for how and why these results are consistent with human values.
	In short, much more research is needed to make ML and AutoML `performant' in the broadest socioethical sense of the word.
	
	\item \textbf{Establishing concepts of operation for HCI in AutoML.}
	Concepts of operation cover the determination of tasks, the responsibilities of stakeholders, the degree of interrelation between operations that humans and AutoML systems separately partake in, and how interactions between humans and AutoML systems should be governed~\citep{en17}.
	Much research into HCI for automated systems relies on a concept of operation that has stakeholders play the roles of supervisors, i.e.~responsible for guiding and overseeing system performance.
	However, when machines become able to work more autonomously, a popular alternative for such a concept of operation is with humans and the system acting as partners or teammates, i.e.~cooperating and coordinating on the tasks required by a problem~\citep{tare95}.
	Therefore, working out how to support collaborative interactions with UX design and other protocols is a future path for AutoML research.
	Already, it is known that well-designed UIs must provide humans with essential supporting functions that reduce workload, increase performance, and guarantee both safety and reliability~\citep{sh20}.
	Many more requirements have been proposed for concepts of operation that facilitate human-system cooperation~\citep{klwo04}.
	These include establishing joint goals, communicating/understanding the roles of each other, having adequate mental models of team members, being predictable to each other, sharing and comprehending the status and intentions of other team members, being able to adapt and negotiate goals in new situations, and replanning collaborative problem-solving strategies if needed.
	Shared situational awareness likewise remains an important principle~\citep{en17}, and developing smart interfaces with sound UX design principles that accommodate this is appealing~\citep{koen12}.
	Essentially, it is up to the AutoML field to decide the ideal way that stakeholders should interact with automated services in the coming years.
	If the benefits of collaborative interactions are to be embraced, researchers and developers will need to determine appropriate concepts of operation for the future.
	
	\item \textbf{Building common protocols and design principles for open-world ML.}
	This monograph has forecasted, with supporting evidence, two major phases of progress to expect in the future of AutoML. 
	Although innovation and ingenuity are required for any significant advances, fully automating most of an ML workflow -- the first phase of progress -- will unlikely require any substantially unconventional thought.
	As exemplified by AI systems dominating Chess and Go, continually improving search strategies may imbue automated mechanisms with sufficient proficiency for solving well-defined and constrained ML problems.
	On the other hand, the initial stage of an ML workflow, i.e.~problem formulation and context understanding, still relies heavily on human creativity and versatility.
	If -- the second phase of progress -- AutonoML is ever to grapple capably with open-ended environments, developing a standard and comprehensive ontology for how humans and machines are expected to operate in this space may be necessary.
	For instance, researchers and developers may need to contemplate the varied ways in which an ML problem and objectives can be defined, decide what knowledge base should support an automated process of context understanding, establish whether this procedure should be guided or restricted, and generally determine how to make open-ended environments manageable.
	Also, so that this endeavour remains feasible and practicable, it is worth contemplating how best to incrementally open up the constraints of ML problems, ensuring theory and practice can handle those tasks before advancing even further.
	For instance, a gradual relaxation may proceed as follows: (1) ensure AutonoML can adapt to changes in data distribution, (2) deal with changes in concept such as new features/labels without discarding already learned knowledge, (3) efficiently attack novel ML problems related by concept to previous encounters by transferring models and knowledge as appropriate, and so on.
	
	\item \textbf{Embedding situational and context awareness into AutoML.}
	The abilities to perceive situations and understand contexts are factors that imbue humans with the versatility to address diverse problems in complex and dynamic environments.
	Thus, to progress AutoML beyond restricted AutonoML into open-world learning, mechanisms will first need to be engineered to identify concepts/contexts that do not immediately fit with prior knowledge accumulations.
	Such a mechanism for situational awareness arguably needs to satisfy several characteristics~\citep{en11}.
	First of all, an autonomous system must construct and maintain a model of itself and its operational environment, at least for relevant factors of interest.
	It must then be able to facilitate dynamic information flow and integrate low-level data into this model.
	Next, the situational awareness model must be able to actively learn from new insights captured by the system.
	Additionally, the model needs an effective well-structured representation of existing situations for quick recall, in real-time, whenever a similar situation is encountered.
	These principles are well-reasoned, but, given that a major focus of ML research is pushing the technical capability of deep neural networks, many proposals for high-level ML frameworks have not yet been translated into practical application.
	There is thus much scope for future research impact by merely verifying if proposed principles supporting prerequisite phenomena to open-world learning, such as situational awareness, are robust, complete, and useful.
	
	\item \textbf{Fusing data-driven search methods with higher-level reasoning.}
	Although advanced strategies for search and learning can bring success to the Alpha Zero and MuZero systems, modern-day ML is constrained to problems and data environments where contexts and relevant rules are well defined.
	In contrast, humans can transfer knowledge and insight among tasks, adapt to changes in a dynamic environment, reason causally and with abstraction, and logically generalise concepts rather than simply recognise patterns and detect associations~\citep{bohu21}.
	Therefore, advancing the capabilities of AutonoML beyond situational awareness may require finding ways of integrating knowledge-driven learning and reasoning approaches with data-driven search algorithms.
	As the most speculative future research direction here, it is unclear whether this is the best path towards open-world learning.
	However, it is \textit{a} path with plenty of expert backing and warrants investigation at some point; any successes here will definitively push the versatility and general applicability of AutonoML.
	
\end{itemize}

\section{Conclusions}\label{conclusion}

This monograph reviewed the roles and modes of human interactions with AutoML systems.
It detailed current practices with modern-day packages and services, then extrapolated to an expected future where frameworks can proficiently run ML applications autonomously.
This forecasted advance was broken into two sequential phases.
First, AutoML reaches a stage of full technical automation at which, now capable of continuous independent operation, AutonoML can deftly handle constrained and well-defined problems.
Second, AutonoML attains automated context understanding, at which point associated systems become capable of open-world learning.

To support the discussion of HCI and its evolution alongside this projected technical progression, different types of stakeholders were identified, as were their expected interactions at various stages of an ML workflow.
These stages covered problem formulation, context understanding, data engineering, model development, deployment, monitoring, and maintenance.
Additionally, the way users engage with an ML solution housed by an AutoML system was also considered part of stakeholder interactions.
With this conceptual framework established, existing HCI literature was then reviewed in the field of AutoML and beyond.
Accordingly, research opinions on UI requirements and approaches were discussed.
This monograph also highlighted the topical issue of trust in AI, recontextualised for AutoML, detailing numerous factors that influence stakeholder engagement, such as explainability and human cognitive biases.
Various perspectives on strengthening HCI were likewise surveyed, examining the problem of fairness and ethical compliance.

Having discussed the present state of human interactions in AutoML, this monograph began its forecast of HCI in earnest, identifying changes to expect as ML problem-solvers become steadily more autonomous.
By and large, humans are likely to shift further and further into hands-off supervisorial roles, with less need to be involved in advanced search and learning processes.
The freed labour may instead be redistributed to more creative and innovative aspects of ML applications, such as formulating problems better, enriching the understanding of their contexts, and interpreting ML solution outcomes.
However, with more freedom and independence comes more risk and responsibility.
Prevalent opinions maintain that mechanisms for HCI will need to support effective human oversight; this becomes even more important as problem constraints relax.

The end-point of AutoML evolution, at least within the scope of this review, is open-world learning.
This monograph considered one possible pathway for the technology to attain this capability, reviewing research and perspectives on hybridising data-driven ML with both knowledge-driven learning and high-level reasoning.
The consequence of elevating AutonoML with these cognitive capabilities is that systems will likely be seen more as partners or teammates of human operators than mere tools.
Subsequently, opinions on how to optimise these collaborative interactions were surveyed.
However, doing so in the best manner remains but one of many open questions in the field.
Acknowledging this, the monograph ended by identifying and listing several promising research directions.

Ultimately, this review concludes that, while considerations of HCI for AutoML have begun in earnest, there is much further to go.
It remains a frontier topic in an emergent field, nascent but both essential and promising.

\bibliographystyle{ACM-Reference-Format}
\bibliography{main}

\end{document}